\pdfoutput=1

\documentclass[11pt]{article}

\usepackage{acl}

\usepackage{times}
\usepackage{latexsym}

\usepackage[T1]{fontenc}

\usepackage[utf8]{inputenc}

\usepackage{microtype}

\usepackage{inconsolata}

\usepackage{url}
\usepackage{bbding}
\usepackage{pifont}
\definecolor{green}{HTML}{009B55}
\usepackage{microtype}
\usepackage{graphicx}
\usepackage{subfigure}
\usepackage{booktabs} 
\usepackage{xspace}
\usepackage[ruled,vlined]{algorithm2e}
\usepackage{macros}
\usepackage{wrapfig}
\usepackage{lipsum}
\usepackage{amsfonts}       
\usepackage{nicefrac}       
\usepackage{xcolor}         
\usepackage{hyperref}
\usepackage{multicol}
\usepackage{multirow}
\usepackage{listings}
\usepackage{pythonhighlight}
\usepackage{xcolor, colortbl}  
\definecolor{nblue}{cmyk}{0.95,0.0,0.2,0.2}

\usepackage{makecell}

\newcommand{\method}{\texttt{MedAdapter}\xspace}

\usepackage{listings}
\usepackage{pythonhighlight}
\usepackage{fancyvrb}
\RecustomVerbatimCommand{\VerbatimInput}{VerbatimInput}{fontsize=\footnotesize,
 frame=single,  
 framesep=0.5em, 
 labelposition=topline,
}
\usepackage{tcolorbox}
\tcbuselibrary{listings}
\title{\method: Efficient Test-Time Adaptation of Large\\ Language Models Towards Medical Reasoning}

\author{Wenqi Shi$^{\spadesuit}$\thanks{~Equal contribution.}, Ran Xu$^\heartsuit$\footnotemark[1], Yuchen Zhuang$^{\spadesuit}$, Yue Yu$^\spadesuit$, Haotian Sun$^\spadesuit$, \\ \bf Hang Wu$^{\spadesuit}$,  Carl Yang$^{\heartsuit}$, May D. Wang$^{\spadesuit}$\\\\
$^{\spadesuit}$ Georgia Tech~~~~$^{\heartsuit}$ Emory University\\
\texttt{\{wqshi,yczhuang,yueyu,haotian.sun,hangwu,maywang\}@gatech.edu}\\
\texttt{\{ran.xu,j.carlyang\}@emory.edu}
}
\begin{document}
\maketitle

\begin{abstract}
Despite their improved capabilities in generation and reasoning, adapting large language models (LLMs) to the biomedical domain remains challenging due to their immense size and privacy concerns. 
In this study, we propose \method\footnote{Our implementation of \method is available at \url{https://github.com/wshi83/MedAdapter}.}, a unified post-hoc adapter for test-time adaptation of LLMs towards biomedical applications.
Instead of fine-tuning the entire LLM, \method effectively adapts the original model by fine-tuning only a small BERT-sized adapter to rank candidate solutions generated by LLMs.
Experiments on four biomedical tasks across eight datasets demonstrate that \method effectively adapts both white-box and black-box LLMs in biomedical reasoning, achieving average performance improvements of 18.24\% and 10.96\%, respectively, without requiring extensive computational resources or sharing data with third parties.
\method also yields enhanced performance when combined with train-time adaptation, highlighting a flexible and complementary solution to existing adaptation methods.
Faced with the challenges of balancing model performance, computational resources, and data privacy, \method provides an efficient, privacy-preserving, cost-effective, and transparent solution for adapting LLMs to the biomedical domain.
\end{abstract}

\section{Introduction}

\begin{figure}[t]
    \centering    
    \includegraphics[width=0.99\linewidth]{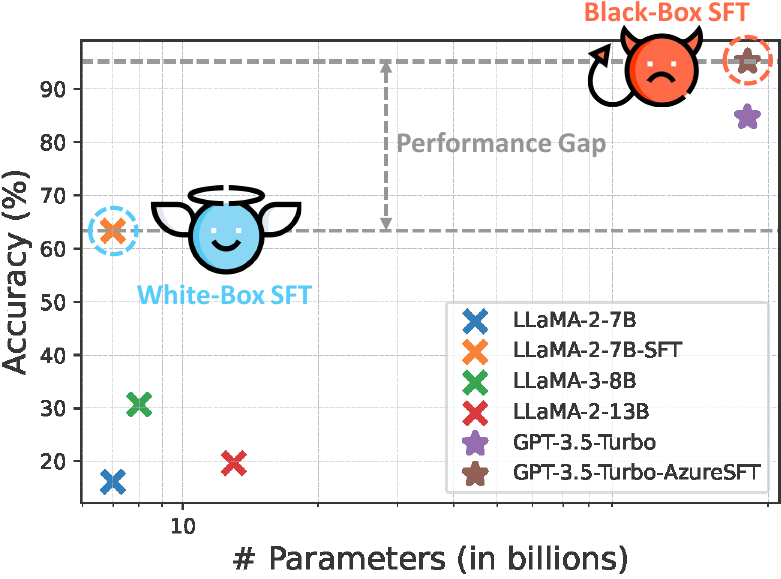}
    \caption{Evaluation results on BioASQ. X-axis in log scale. Moderately-sized white-box LLMs consistently \emph{underperform} larger black-box LLMs, regardless of fine-tuning on biomedical corpora. However, fine-tuning black-box LLMs through APIs can pose potential \emph{data privacy risks} and incur \emph{substantial costs}.}
    \label{fig:tease}
\end{figure}

Large language models (LLMs)~\cite{chatgpt,achiam2023gpt,team2023gemini} have demonstrated superior generation and reasoning capabilities compared to traditional BERT-sized language models, primarily due to the massive number of parameters and extensive pre-training on vast textual corpora.
In the biomedical domain, researchers have developed LLMs that are either pre-trained~\citep{chen2023meditron,bolton2024biomedlm} or fine-tuned~\citep{singhal2023large,han2023medalpaca} on large-scale domain-specific corpora to enhance performance on biomedical natural language processing (NLP) tasks. 
However, tuning biomedical domain-specific LLMs triggers additional considerations due to their \emph{immense size} and \emph{corporate privacy}, especially given (1) the \textbf{resource constraints} in academic institutions and medical centers and (2) the \textbf{sensitive nature} of clinical data.

Although fine-tuning LLMs accelerates biomedical discovery and improves patient care~\citep{han2023medalpaca,zhang2023alpacare,wang2024drg}, it usually necessitates complete access to internal parameters, which is currently limited to white-box LLMs like LLaMA-series models~\citep{touvron2023llama,llama3}.
However, a significant \emph{performance discrepancy} still exists between larger black-box LLMs (\eg, GPT-3.5-Turbo) and smaller white-box LLMs (\eg, LLaMA-2)~\citep{labrak2024biomistral,singhal2023large,chen2023meditron}, even when the latter are fine-tuned on biomedical-specific corpora (Figure~\ref{fig:tease}). 
Moreover, fine-tuning even a moderately-sized LLM with 7B parameters requires \emph{substantial computational resources}~\citep{bolton2024biomedlm}, often exceeding the capabilities of many academic and medical centers.

Such intrinsic limitations of white-box LLMs intuitively motivate the exploration of adapting black-box LLMs to the biomedical domain. 
While it is possible to fine-tune black-box LLMs like GPT-3.5~\citep{chatgpt} via third-party APIs~\citep{SFTgpt} without direct access to internal parameters, this approach presents several unique challenges within the field of biomedicine:
(1) Uploading patient data via APIs poses significant risks of \emph{privacy leakage} and potential conflicts with Health Insurance Portability and Accountability Act (HIPAA) compliance, including unauthorized third-party access to personally identifiable information (PII)~\citep{lukas2023analyzing,marks2023ai,wang2023ethical};
(2) Fine-tuning API services could incur prohibitively \emph{high financial and environmental costs}~\citep{luccioni2023estimating}, exceeding typical academic or clinical budgets;
(3) The opaque fine-tuning process, limited to very few adjustable hyperparameters within a specific range, often results in \emph{suboptimal performance} in downstream tasks~\citep{sun2024bbox}, whereas medical applications often demand precise outcomes.

In this study, we rethink the trade-off between model performance concerns in white-box LLMs and data privacy issues in black-box LLMs for biomedical tasks from a new perspective. We introduce \method, a unified test-time adapter that fine-tunes \textbf{a lightweight BERT-sized language model (110M)} to facilitate the adaptation of both white-box and black-box LLMs for medical reasoning. 
Instead of updating the parameter for the entire LLM, \method fine-tunes a small outcome-supervised adapter that ranks candidate solutions generated by LLMs, effectively and efficiently adapting the original LLM to the target domain.
In addition, it also eliminates the need to (1) access the large-scale internal model parameters or (2) share any private patient information with third parties through fine-tuning APIs.

Extensive experiments on \textbf{four biomedical reasoning tasks across eight datasets} demonstrate that \method effectively adapts both white-box and black-box LLMs for medical reasoning, achieving average performance improvements of 18.24\% and 10.96\%, respectively.
\emph{For white-box LLMs}, \method reaches 99.35\% of supervised fine-tuning performance using only 14.75\% of the GPU memory on BioASQ.
\emph{For black-box LLMs}, it achieves comparable performance or even surpasses fine-tuning APIs at only 15.59\% of the budget, while also eliminating the risks associated with private data sharing. 
We summarize our contributions as follows:
\begin{itemize}
    \item We introduce \texttt{\method}, a \textbf{unified} post-hoc adapter designed to facilitate the efficient test-time adaptation of both white-box and black-box LLMs for medical reasoning.
    \item Compared to supervised fine-tuning of white-box LLMs, \method achieves \textbf{effective} domain adaptation using a \textbf{BERT-sized} language model with only 110M parameters.
    \item Compared to supervised fine-tuning of black-box LLMs via APIs, \method offers a more \textbf{privacy-preserving}, \textbf{cost-efficient}, and \textbf{transparent} alternative, eliminating the need for access to any model parameters.
    \item When combined with train-time adaptation, \method outperforms either train-time or test-time adaptation alone, underscoring its utility as a \textbf{flexible} and \textbf{complementary} solution to existing adaptation methods. 
\end{itemize}

\section{\method: Adapting LLMs to Medical Reasoning}
\subsection{Preliminaries}
\noindent\textbf{Problem Formulation.}
Test-time adaptation\footnote{We adopt a slightly different definition of test-time adaptation than several existing studies~\citep{zancato2023train,karmanov2024efficient}; we only require target domain label information to remain invisible to the original LLM and stay accessible to the adapter.} refers to the process of customizing models to test data that may exhibit distributional deviations from the original training data.
Given a pre-trained LLM $G_\phi$ and a training dataset from the target domain $\mathcal{D}=\{(\mathbf{x}_i,\mathbf{y}_i)\}_{i=1}^{|\mathcal{D}|}$, where $\mathbf{x}_i$ typically describes the task input and $\mathbf{y}_i$ represents the ground-truth answer for the $i$-th example.
The goal is to adapt the outputs of the LLM $\mathbf{\hat{y}}_i^s\in\mathcal{Y}^S$ from the general source domain to a specific target domain $\mathbf{y}^t\in\mathcal{Y}^T$ for each input instance $\mathbf{x}_i$.
Such adaptation can be crucial for enhancing the capability of an LLM to exhibit biomedical domain-specific reasoning, which may be underdeveloped in its original outputs.
According to the accessibility of model parameters, existing approaches can be categorized into two main groups: (1) \emph{white-box} LLM adaptation, which allows full access to model parameters, and (2) \emph{black-box} LLM adaptation, which permits no such access.

\begin{figure*}[t]
    \centering    
    \includegraphics[width=0.99\linewidth]{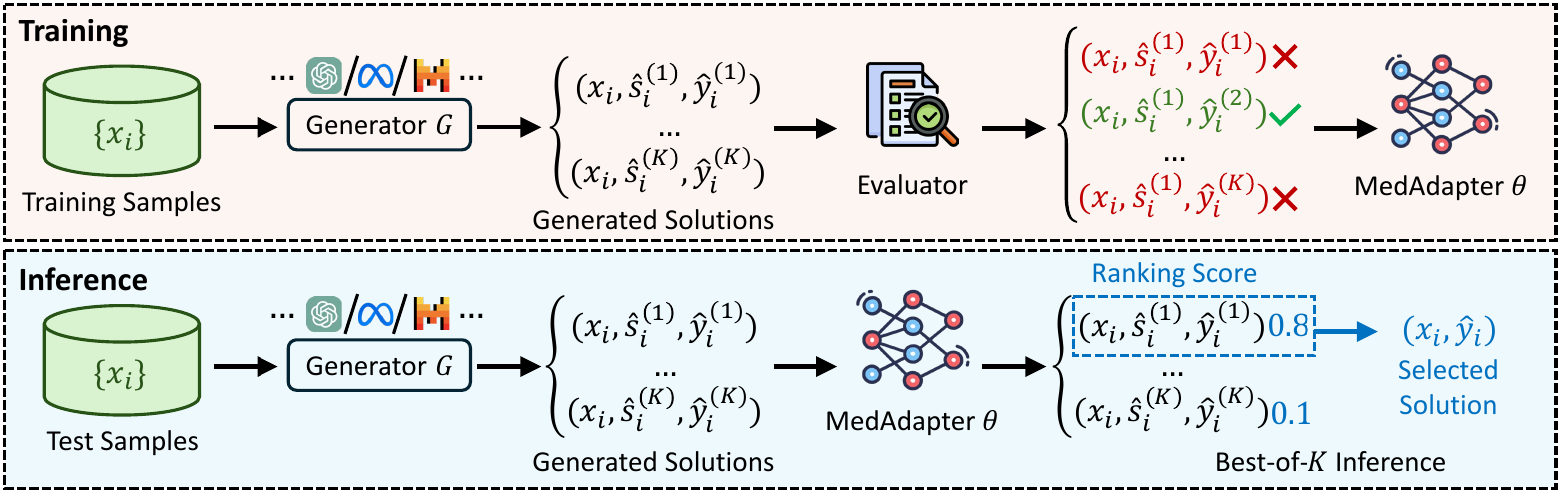}
    \caption{Overview of \method for efficient test-time LLM adaptation towards medical reasoning. We fine-tune a small adapter, \method, to rank candidate solutions generated by LLMs, thereby effectively establishing a distinction between the source and target domains for efficient domain adaptation.}
    \label{fig:overview}
\end{figure*}

\noindent\textbf{White-box LLM Adaptation.}
With model parameters available in white-box LLMs, the most direct approach for domain adaptation is supervised fine-tuning~\citep{wei2021finetuned, chung2024scaling} with the negative log-likelihood learning objective on the training data:
\begin{equation}
    \begin{aligned}
        \mathcal{L}_{\text{SFT}}(\phi)=-\mathbb{E}_{(\mathbf{x},\mathbf{y})\sim \mathcal{D}}\sum_{t=1}^T\log G_{\phi}(\mathbf{y}_t|\mathbf{y}_{<t},\mathbf{x}).
    \end{aligned}
\end{equation}
In practice, for efficient adaptation of large pre-trained models to various downstream applications, parameter-efficient fine-tuning (PEFT) methods~\cite{houlsby2019parameter, hu2021lora} have been proposed.  These methods involve fine-tuning only a small subset of (additional) model parameters, significantly reducing both computational and storage costs.
Although PEFT-based methods provide a practical solution with limited computational resources, they compromise model performance for efficiency.

\noindent\textbf{Black-box LLM Adaptation.}
State-of-the-art LLMs, including GPT-4~\cite{achiam2023gpt}, Claude~\cite{claude3}, and Gemini~\cite{team2023gemini}, adhere to a trend of non-disclosure of model parameters to the public.
Consequently, fine-tuning these black-box LLMs relies solely on fine-tuning web service APIs, such as the OpenAI GPT-3.5-turbo fine-tuning API~\cite{SFTgpt}, which lacks transparency and incurs high costs.
In response, recent black-box adaptation methods~\cite{liu2024tuning, ormazabal2023comblm, huang2023k} have explored the adjustment of logit biases 
for increasing the frequency of tokens from the target domain appearing in the output while penalizing those from the source domain. 
However, such black-box adaptation methods remain inapplicable to the latest cutting-edge black-box LLMs, such as GPT-3.5-turbo~\citep{chatgpt}, due to the \emph{unavailability of token probabilities}.
Although few recent studies~\cite{xu2023small, sun2024bbox} bypass the need for full parameter access, they are \emph{limited to specific tasks}: they only support classification tasks that rely on label predictions with confidence~\cite{xu2023small}, or multi-step reasoning tasks that require process-level supervision and beam search~\cite{sun2024bbox}. These constraints significantly limit the applicability of such methods to diverse biomedical reasoning applications.

\subsection{Overview of \method}
The rapid increase in the size of LLMs exacerbates the existing disparity between resource-abundant and resource-scarce biomedical institutions~\cite{gema2023parameter}, especially given the high privacy of patient information.
To address this, we propose \method, a unified post-hoc adapter that facilitates test-time adaptation without the need for significant computational resources or access to model parameters (Figure~\ref{fig:overview}).
Benefiting from the strong generation capabilities of recent LLMs, we first leverage LLMs to generate candidate reasoning solutions (Section~\ref{sec:generation}).
We then fine-tune a BERT-sized language model, \method, to rank all candidate solutions, thereby establishing the distinction between the source and target domains (Section~\ref{sec:score}).
Finally, 
\method adapts LLMs by sampling the candidate solution with the highest adaptation score (Section~\ref{sec:best-of-k}).

\subsection{Candidate Solutions Generation}\label{subsec:data-gen}
\label{sec:generation}
For each problem $\mathbf{x}_i$ in the training dataset $\mathcal{D}$, we generate $k$ intermediate candidate reasoning solutions $\{\hat{\mathbf{s}}_{i,j}\}_{j=1}^k$ (\eg, chain-of-thought rationales or multi-step reasonings) and the corresponding answer $\{\hat{\mathbf{y}}_{i,j}\}_{j=1}^k$ using greedy decoding with the language model generator $G$.
With access to ground-truth answers $\mathbf{y}_i$, we can verify the correctness of each generated solution $\hat{\mathbf{y}}_{i,j}$ and assign a corresponding binary correctness label $z_i$ as:
\begin{equation}
    \begin{aligned}
        z_i=\mathbbm{1}(\hat{\mathbf{y}}_{i,j}=\mathbf{y}_i),\ z_i\in\{0,1\}.
    \end{aligned}
\end{equation}
With the generated solutions, we formulate a new dataset for the adapter training, denoted as: 
\begin{equation}
\scalebox{0.95}{
    $\mathcal{D}_{\text{ada}}=\{(\mathbf{h}_{i,j},z_i) \mid 1\leq i \leq |\mathcal{D}|, 1 \leq j \leq k\}$,
}
\end{equation}
where $\mathbf{h}_{i,j}=[\mathbf{x}_i||\hat{\mathbf{s}}_{i,j}||\hat{\mathbf{y}}_{i,j}]$, represents the concatenation of the medical question and the entire candidate generation, and $z_i$ is a binary label indicating whether $\hat{\mathbf{y}}_{i,j}$ is a correct or incorrect solution.

\subsection{Outcome-Supervised Adapter}
\label{sec:score}
To enable the distinction between source and target domain, we train an outcome-supervised adapter (\ie, verifier) that assesses the probability of correctness for a candidate solution relative to a given problem. 
During inference, the language model $G$ generates a set of candidate solutions, and the one ranked highest by the verifier is selected as the final answer, aligning closely with the target domain.
More specifically, given a medical reasoning problem $\mathbf{x}$ and its corresponding candidate solutions $\hat{\mathbf{y}}$, the outcome verifier ($V:\mathcal{X}\times\mathcal{Y}\to\mathbb{R}$) assigns a normalized adaptation score, ranging from 0 to 1, to each solution to indicate the correctness.

In \method, we fine-tune a BERT-sized language model $\theta$ ($\sim 110$M parameters), to function as an outcome-supervised adapter on $\mathcal{D}_{\text{ada}}$. Following the empirical study on the effect of different objective functions in Section~\ref{sec:loss}, we employ a combination of language modeling and binary classification as the objective function:
\begin{equation}
    \begin{aligned}
        \mathcal{L}_{\text{ada}}=z\log V_\theta(\mathbf{h})+(1-z)\log (1-V_\theta(\mathbf{h})),
    \end{aligned}
\end{equation}
where $z$ is the binary label verified against the ground-truth answer provided in $\mathcal{D}_{\text{ada}}$, and $V_\theta(\mathbf{h})$ is the sigmoid adaptation score of the question-solution pair $\mathbf{h}$ assigned by the adapter model.

\subsection{Best-of-$K$ Inference}
\label{sec:best-of-k}

During the inference stage, for each test question $\mathbf{x}_i$, we adopt the best-of-$K$ inference, often referred to as rejection sampling, to select the best solution from multiple candidates.
We initially sample $K$ candidate solutions $\{\hat{\mathbf{s}}_{i,j},\hat{\mathbf{y}}_{i,j}\}_{j=1}^K$ from the generator $G$. The solution with the highest adaptation score is then selected:
\begin{equation}
    \begin{aligned}
        \hat{\mathbf{y}}_i=\arg\max_{j=1,\cdots,K} r_\theta([\mathbf{x}_i||\hat{\mathbf{s}}_{i,j}||\hat{\mathbf{y}}_{i,j}]).
    \end{aligned}
\end{equation}

\paragraph{Remark.} We note that, in contrast to prior verification-guided in-context learning methods~\citep{li-etal-2023-making,khalifa-etal-2023-grace} that depend on large-scale intermediate reasoning annotations, \method utilizes candidate solutions generated by LLMs to form positive and negative examples, thus removing the need for human-annotated intermediate reasoning steps. Additionally, the lightweight design of the adapter $\theta$ results in a minimal increase in memory usage and inference time. The efficiency study is presented in Section~\ref{sec:pefficiency}.


\section{Experiments}

\begin{table*}[t]
\centering
\fontsize{8}{10}\selectfont\setlength{\tabcolsep}{0.4em}
\resizebox{\linewidth}{!}{
\begin{tabular}{@{}lcccccccccc cc cc cc@{}}
\toprule
\textbf{Dataset ($\rightarrow$)} & \multicolumn{2}{c}{\textbf{MedMCQA}} & \multicolumn{2}{c}{\textbf{MedQA}} &  \multicolumn{2}{c}{\textbf{MMLU-Med}} & \multicolumn{2}{c}{\textbf{PubMedQA}} & \multicolumn{2}{c}{\textbf{BioASQ}}
& \multicolumn{2}{c}{\textbf{MedNLI}}
& \multicolumn{2}{c}{\textbf{MediQA-RQE}}
& \multicolumn{2}{c}{\textbf{PubHealth}}
\\
\cmidrule(lr){2-3} \cmidrule(lr){4-5} \cmidrule(lr){6-7} \cmidrule(lr){8-9} \cmidrule(lr){10-11}  \cmidrule(lr){12-13}  \cmidrule(lr){14-15}  \cmidrule(lr){16-17}
\textbf{Method ($\downarrow$)}$\slash$\textbf{Metrics ($\rightarrow$)} & Acc. (\%) & $\Delta$ (\%) & Acc. (\%) & $\Delta$ (\%) &  Acc. (\%) & $\Delta$ (\%) & Acc. (\%) & $\Delta$ (\%)  & Acc. (\%) & $\Delta$ (\%) & Acc. (\%) & $\Delta$ (\%) & Acc. (\%) & $\Delta$ (\%) & Acc. (\%) & $\Delta$ (\%) \\\midrule
\texttt{LLaMA-2-7B}~\shortcite{touvron2023llama} & 16.00	& -- & 16.42	& -- & 20.13	& -- & 17.00	& -- & 16.13 & -- & 17.80 & -- & 23.91 & -- & 16.89 & -- \\
\quad +Self-Consistency~\shortcite{wang2023selfconsistency} & 21.20 & +5.20	& 22.39	& + 5.97 & 23.27 & +3.14 & 28.00 & +11.00 &	17.74	& +1.61 & 27.87 & +10.07 & 	25.22  & +1.31 & 	17.79 & +0.90 \\
\rowcolor{teal!10} \quad +\method & 32.00 & +16.00	& 32.52 & +16.10 & 27.67 & +7.54	& 58.00 & +41.00 & 62.90 & +46.77 & 30.46	& +12.66 & 27.39 & +3.48 &	19.25  & +2.36 \\ 
\rowcolor{gray!16} \quad +SFT$^\dagger$ & 42.86	& +26.86 & 33.39	& +16.97 & 28.22	& +8.09 & 60.80	& +43.80 & 63.31 & +47.18 & 65.52 & +47.72  &	35.42	& +11.51 & 22.00 & +5.11\\ \midrule
\texttt{BioMistral-7B}~\shortcite{labrak2024biomistral} & 28.95 & -- & 29.77 & --  & 33.33	& -- & 26.20	& -- & 28.53 & -- &22.03 &	-- &42.37	& -- & 25.73& -- \\
\quad +Self-Consistency~\shortcite{wang2023selfconsistency} & 29.18 & +0.23 &	32.68	& +2.91 & 39.62 & +6.29 &	30.60 & +4.40 &	31.45 & +2.92 &  31.46 & +9.43 & 	44.68& +2.31 & 	28.90 & +2.17 \\
\rowcolor{teal!10} \quad +\method & 30.31 & +1.36 & 34.88 & +5.11  & 46.54 & +13.21 & 33.20 & +7.00 & 33.06 & +4.53 & 35.96	& +13.93 & 45.53	& +3.18 & 30.84 & +5.11\\ \midrule
\texttt{LLaMA-3-8B}~\shortcite{llama3} & 20.44 & -- & 27.81 & --  & 25.16	& -- & 19.00	& -- & 30.65 & -- & 21.96	& -- & 49.13	& -- & 27.70 & -- \\
\quad +Self-Consistency~\shortcite{wang2023selfconsistency} & 26.87 & +6.43 & 	31.50 & +3.69 & 	31.45 & +6.29 &	37.00 & +18.00 &	33.06 & +2.41 & 	30.12 & +8.16 & 	50.87 & +1.74 &	35.01 & +7.31\\
\rowcolor{teal!10} \quad +\method & 32.08 & +11.64 & 32.44 & +4.63  & 35.22 & +10.06 & 55.00 & +36.00 & 64.52 & +31.46 & 32.09 & +10.13 & 51.74	& +2.61 & 36.07 & +8.37\\ \midrule
\texttt{LLaMA-2-13B}~\shortcite{touvron2023llama} & 19.66	& -- & 28.04	& -- & 22.01	& -- & 47.40	& -- & 19.66 & -- & 21.75	& -- & 30.44 & -- &	19.33 & -- \\
\quad +Self-Consistency~\shortcite{wang2023selfconsistency} & 28.40	& + 8.74 & 31.03 & +2.99 & 	28.30	& + 6.29 & 56.80	& +9.40 & 51.61	& +31.95 & 24.21	& +2.46 & 43.04	& +12.60 & 24.70	& +5.37 \\
\rowcolor{teal!10} \quad +\method & 32.00 & +12.34	& 37.47	& +9.43 & 33.96	& +11.95 & 63.60 & +16.20 &	65.32 & +45.66 & 26.88	& +5.13 & 44.78 & +14.34 & 	27.46 &  +8.13\\\midrule
\texttt{gpt-3.5-turbo}~\shortcite{chatgpt} & 49.74 & -- & 61.51 & -- & 59.75 & -- &  56.00 & -- &  84.68 & --  & 66.64 & -- & 50.00	& -- & 23.38 & --\\
\quad +Self-Consistency~\shortcite{wang2023selfconsistency} &  56.20 & +6.46 & 67.71 & +6.20 & 69.81  & +10.06 &  71.60 & +15.60 & 87.90  & +3.22  & 69.18 & +2.54 & 	51.30 & +1.30 & 	25.41 & +2.03\\
\quad +MedRAG~\shortcite{xiong2024benchmarking} & 51.80 & +2.06 & 64.36 & +2.85 & 68.85 & +9.10 &  50.00 & -6.00 &  87.55 & +2.87  & --& --& -- & --& -- &-- \\
\rowcolor{violet!10} \quad +\method & 59.02 & +9.28 & 68.66 & +7.15 & 73.58 & +13.83 &  73.40 & +17.40 &  93.55 & +8.87  & 75.09 & +8.45 &	52.61 & +2.61 & 33.43 & +10.05 \\ 
\rowcolor{gray!16} \quad +Azure-SFT$^\dagger$~\shortcite{SFTgpt} & 61.82 & +12.08 & 63.32 & +1.81 & 70.55 & +10.80 & 71.40 & +15.40 & 95.16 & +10.48 & 91.27 & +24.63 & 	58.08 & +8.08 & 36.56 & +13.18\\ \midrule
\texttt{gpt-4}~\shortcite{achiam2023gpt} & 69.48 & -- & 83.90 & -- & 85.53 & -- &  69.20 & -- &  92.74 & -- & 86.77 & -- & 	51.30 & -- & 	38.52 & -- \\
\quad +Self-Consistency~\shortcite{wang2023selfconsistency} & 70.08	 & +0.60 & 84.05	& +0.15 & 86.79	& +1.26 & 72.20	& +3.00 & 93.54	& +0.8 & 87.26	& +0.49 & 51.74	& +0.44 & 43.35 & +4.83 \\
\quad +MedRAG~\shortcite{xiong2024benchmarking} & 66.65 & -2.83  & 82.80 & -1.10 & 87.24 & +1.71 &  70.60 & -1.40 &  92.56 & -0.18  & --& --& -- & --& -- &-- \\

\rowcolor{violet!10} \quad +\method & 72.09 & +2.61 & 84.13 & +0.23 &  87.42 & +1.89 & 77.40 & +8.20 & 95.97 & +3.23 & 87.68	& +0.91 & 53.04	& +1.74 & 46.34 & +7.82 \\\bottomrule
\end{tabular}
}
\caption{Main results (accuracy) of adapting \colorbox{teal!10}{white-box} and \colorbox{violet!10}{black-box} LLMs to biomedical tasks. $\dagger$ denotes the upper bound in theory using \colorbox{gray!16}{supervised fine-tuning (SFT)}. Specifically, we perform \colorbox{gray!16}{Azure-SFT} for black-box LLMs via Microsoft Azure OpenAI fine-tuning API services to ensure compliance with HIPAA regulations. Notations are consistent across tables. The results of MedRAG on smaller LLMs are not reported in their paper.}
\label{tab:main-white-box}
\end{table*}

\subsection{Experimental Setups}
\noindent \textbf{Tasks and Datasets.} For a comprehensive evaluation, we examine \method mainly on five datasets for biomedical QA task: (1) \textbf{MedMCQA}~\citep{pal2022medmcqa}, (2) \textbf{MedQA}~\citep{jin2021disease}, (3) \textbf{MMLU}~\citep{hendrycks2020measuring}, (4) \textbf{PubMedQA}~\citep{jin2019pubmedqa}, (5) \textbf{BioASQ}~\citep{tsatsaronis2015overview};
and three additional biomedical NLP tasks, including   
(6) \textbf{MedNLI}~\citep{mednli} for natural language inference (NLI),
(7) \textbf{MediQA-RQE}~\citep{mediqa-rqe} for recognizing question entailment (RQE), and
(8) \textbf{PubHealth}~\citep{PUBHEALTH} for health fact-checking.
For detailed information, please refer to Appendix~\ref{app:dataset}.

\noindent \textbf{Baselines.}
We conduct our main experiments using both white-box and black-box backbone LLMs.
We employ the \emph{Chain-of-Thoughts} (CoT) results~\citep{wei2022chain} as the baseline performance for all backbone LLMs without adaptation.

\noindent $\diamond$ For \underline{white-box} LLM adaptation, we primarily compare \method against \emph{supervised fine-tuning}, which updates all of the model parameters and serves as the upper-performance benchmark. We adapt widely used open-source LLaMA models~\citep{touvron2023llama,llama3} across various versions and scales, as well as medical domain-specific LLMs like BioMistral-7B~\cite{labrak2024biomistral} for a comprehensive evaluation.

\noindent $\diamond$ For \underline{black-box} LLM adaptation,
we focus on the comparison between \method and \emph{supervised fine-tuning} using the Microsoft Azure OpenAI fine-tuning API service~\citep{SFTgpt}. In addition, we compare \method with other privacy-preserving solutions, including \emph{self-consistency}~\citep{wang2023selfconsistency} and medical domain-specific \emph{retrieval-augmented generation} (RAG)~\citep{xiong2024benchmarking}, which do not require uploading training data to third parties\footnote{
We incorporate in-context learning baselines in biomedical applications from privacy-preserving perspectives. Note that due to context length limits, in-context learning can only rely on a limited number of supervised examples; the model performance is only for reference.
}.

\noindent \textbf{Evaluation Metric.}
Following \citet{bolton2024biomedlm}, we adopt accuracy as the main evaluation metric for all biomedical tasks.

\noindent \textbf{Implementation Details.} 
In this work, we employ \texttt{LongFormer-Base} (110M)~\citep{beltagy2020longformer} as the base language model for \method. 
We set $k=8$ for all generations of intermediate candidate reasoning solutions using \method.
Additional implementation details, including prompt templates, are available in Appendix~\ref{app:prompt}.

\begin{table*}[t]
\centering
\fontsize{8}{10}\selectfont\setlength{\tabcolsep}{0.4em}
\resizebox{\linewidth}{!}{
\begin{tabular}{@{}lcccccccccc cc cc cc@{}}
\toprule
\textbf{Dataset ($\rightarrow$)} & \multicolumn{2}{c}{\textbf{MedMCQA}} & \multicolumn{2}{c}{\textbf{MedQA}} &  \multicolumn{2}{c}{\textbf{MMLU-Med}} & \multicolumn{2}{c}{\textbf{PubMedQA}} & \multicolumn{2}{c}{\textbf{BioASQ}}
& \multicolumn{2}{c}{\textbf{MedNLI}}
& \multicolumn{2}{c}{\textbf{MediQA-RQE}}
& \multicolumn{2}{c}{\textbf{PubHealth}}
\\
\cmidrule(lr){2-3} \cmidrule(lr){4-5} \cmidrule(lr){6-7} \cmidrule(lr){8-9} \cmidrule(lr){10-11}  \cmidrule(lr){12-13}  \cmidrule(lr){14-15}  \cmidrule(lr){16-17}
\textbf{Method ($\downarrow$)}$\slash$\textbf{Metrics ($\rightarrow$)} & Acc. (\%) & $\Delta$ (\%) & Acc. (\%) & $\Delta$ (\%) &  Acc. (\%) & $\Delta$ (\%) & Acc. (\%) & $\Delta$ (\%)  & Acc. (\%) & $\Delta$ (\%) & Acc. (\%) & $\Delta$ (\%) & Acc. (\%) & $\Delta$ (\%) & Acc. (\%) & $\Delta$ (\%) \\\midrule
\texttt{LLaMA-2-7B}~\shortcite{touvron2023llama} & 16.00	& -- & 16.42	& -- & 20.13	& -- & 17.00	& -- & 16.13 & -- & 17.80	& -- & 23.91 & -- & 	16.89 & -- \\
\rowcolor{teal!10} \quad +\method & 32.00 & +16.00	& 32.52 & +16.10 & 27.67 & +7.54	& 58.00 & +41.00 & 62.90 & +46.77 & 27.87	& +10.07 & 25.22 & +1.31 & 	17.79 & +0.90 \\ 
\quad +SFT & 42.86	& -- & 33.39	& -- & 28.22	& -- & 60.80	& -- & 63.31 & -- & 65.52 & -- & 	35.42 & -- & 	22.00 & -- \\
\rowcolor{teal!10}\qquad +\method & \textbf{44.85} & +1.99 & \textbf{40.61} & +7.22 & \textbf{35.85}	& +7.63 & \textbf{68.00}	& +7.20 & \textbf{66.94} & +3.63  & \textbf{74.95}	& +9.43 & \textbf{46.54}	& +11.12 & \textbf{33.36} & +11.36\\
\quad +SFT-LoRA~\shortcite{hu2021lora} & 28.95 & -- & 23.89	& -- & 24.54	& -- & 55.00	& -- & 50.00 & -- & 25.97 & -- &	32.17	& -- & 20.47 & -- \\
\rowcolor{teal!10}\qquad +\method & 35.69 & +6.74 & 28.04 & +4.15 & 31.90	& +7.36 & 62.90	& +7.90 & 60.48 & +10.48 & 35.96 & +9.99 & 	39.57 & +7.40 &	25.26 & +4.79\\ \midrule
\texttt{gpt-3.5-turbo}~\shortcite{chatgpt} & 49.74 & -- & 61.51 & -- & 59.75 & -- &  56.00 & -- &  84.68 & -- & 66.64 & -- & 	50.00 & -- &	23.38 & -- \\
\rowcolor{violet!10} \quad +\method & 59.02 & +9.28 & 68.66 & +7.15 & 73.58 & +13.83 &  73.40 & +17.40 &  93.55 & +8.87 & 75.09 & +8.45 &	52.61 & +2.61 & 	33.43 & +10.05 \\ 
\quad +Azure-SFT~\shortcite{SFTgpt} & 61.82 & -- & 63.32 & -- & 70.55 & -- & 71.40 & -- & 95.16 & -- & 91.27	& -- & 58.08 & -- & 	36.56 & --\\
\rowcolor{violet!10} \quad \quad +\method & \textbf{65.50} & +3.68  & \textbf{68.89} & +5.57 & \textbf{76.73} & +6.18 &  \textbf{77.00} & +5.60 &\textbf{ 95.97}  & +0.81 & \textbf{91.42}	& +0.15 & \textbf{59.56} & +1.48 & 	\textbf{42.49} & +5.93  \\ 
\quad +MedRAG~\shortcite{xiong2024benchmarking} & 51.80 & -- & 64.36 & -- & 68.85 & -- &  50.00 & -- &  87.55 & -- & -- & --  & -- & -- & -- & --\\
\rowcolor{violet!10} \quad \quad +\method & 56.20 & +4.40 & 67.16 & +2.80 & 74.86 & +6.01 &  63.00 & +13.00 &  94.42 & +6.87 & -- & -- & -- & -- & -- & --   \\ 
\bottomrule
\end{tabular}
}
\caption{Complementary analysis results (accuracy) of combining training- and test-time adaptation for both white- and black-box LLMs on biomedical tasks. \textbf{Bold} indicates the best performance within white/black-box LLMs.}\label{tab:complementary}
\end{table*}

\subsection{Main Results}
In Table~\ref{tab:main-white-box}, we summarize the experimental results of adapting both white-box and black-box LLMs for four biomedical tasks across eight datasets. 

\paragraph{White-box LLM Adaptation.}
\noindent $\diamond$ \underline{Effectiveness}:
Across all downstream biomedical applications, \method consistently outperforms the original white-box LLM, \texttt{LLaMA-2-7B}~\citep{touvron2023llama}, with an average performance improvement of 25.48\% for QA task, 12.66\% for NLI, 3.48\% for RQE, and 2.36\% for fact-checking, respectively, demonstrating the adaptability of \method towards diverse biomedical domain-specific applications. 
\noindent $\diamond$ \underline{Efficiency}:
Notably, \method demonstrates its efficiency by achieving 87.50\% of the performance level of the fully supervised fine-tuning model while only updating an adapter comprising 110M parameters, which constitutes merely 1.57\% of the parameters (7B) of the original model.
\noindent $\diamond$ \underline{Robustness}:
It also demonstrates an average improvement of 13.34\% over another lightweight test-time adaptation solution, self-consistency~\cite{wang2023selfconsistency}, with more robust adaptation across all tasks. 
\noindent $\diamond$ \underline{Generalization}:
Additionally, \method further improves the performance of domain-specific LLMs like \texttt{BioMistral-7B}~\citep{labrak2024biomistral} and general-domain LLMs at different scales, such as \texttt{LLaMA-3-8B} and \texttt{LLaMA-2-13B}~\citep{touvron2023llama}, demonstrating a generalizable solution for white-box LLM biomedical domain adaptation.

\paragraph{Black-box LLM Adaptation.}
As expected, black-box LLMs, with their extensive model parameters and large pre-training corpora, significantly outperform white-box LLMs (Table~\ref{tab:main-white-box}) across all biomedical applications.
\noindent $\diamond$ \underline{Effectiveness}:
We observe that \texttt{\method} successfully adapts \texttt{gpt-3.5-turbo}~\citep{chatgpt} across all tasks, achieving an average performance improvement of 11.31\% for QA, 8.45\% for NLI, 2.61\% for	RQE, and 20.05\% for health fact-checking.  
\noindent $\diamond$ \underline{Privacy-Preserving}:
Notably, \method achieves competitive or even superior performance compared to supervised fine-tuning via Microsoft Azure APIs, without necessitating the sharing of local training samples with third parties. This may be due to the opacity of the fine-tuning service, which only allows access to a very limited number of adjustable parameters within a prescribed range\footnote{In the Microsoft OpenAI fine-tuning service, users are permitted to modify only four hyperparameters within a limited range: (1) the number of epochs, (2) the batch size, (3) the learning rate multiplier, and (4) the random seed. Details for parameter studies of supervised fine-tuning via Microsoft Azure APIs are available in Appendix~\ref{app:parameter}.}, leading to suboptimal fine-tuning performance. 
\noindent $\diamond$ \underline{Generalization}:
We could also extend \method for more advanced LLMs such as \texttt{gpt-4}~\citep{achiam2023gpt}, demonstrating a flexible and generalizable solution for adapting black-box LLMs in medical reasoning.
\noindent $\diamond$ \underline{Robustness}:
\method provides more effective adaptation compared to other privacy-preserving methods, such as self-consistency~\citep{wang2023selfconsistency} and MedRAG~\citep{xiong2024benchmarking}. 
Specifically, we observe only a slight improvement or even a decrease in performance when adapting RAG-based methods compared to direct adaptations of backbone black-box LLMs. This can be attributed to the conditional generation nature of RAG, which typically results in less diverse candidate solutions.

\subsection{\method Complements Other Adaptation Techniques}
In Table~\ref{tab:complementary}, we perform a complementary analysis to demonstrate the flexibility of \method by integrating both train-time and test-time adaptation.
For example, in the biomedical QA tasks, \method yields an additional performance improvement of 5.53\% and 4.37\% for white-box and black-box LLMs, respectively, over train-time adaptation (\ie, supervised fine-tuning).
When combined with train-time adaptation, \method outperforms either train-time or test-time adaptation alone, demonstrating its utility as a flexible solution that complements existing train-time adaptation methods (\eg, LoRA)~\citep{hu2021lora} and even test-time adaptation (\eg, MedRAG)~\cite{xiong2024benchmarking} to further boost model performance.

\subsection{Cost Estimation}
\begin{table*}[t]
\centering
\fontsize{8}{10}\selectfont\setlength{\tabcolsep}{0.3em}
\begin{tabular}{@{}lcccccccccc@{}}
\toprule
\textbf{Dataset ($\rightarrow$)} & \multicolumn{2}{c}{\textbf{MedMCQA}} & \multicolumn{2}{c}{\textbf{MedQA}} &  \multicolumn{2}{c}{\textbf{MMLU-Med}} & \multicolumn{2}{c}{\textbf{PubMedQA}} & \multicolumn{2}{c}{\textbf{BioASQ}}\\
\cmidrule(lr){2-3} \cmidrule(lr){4-5} \cmidrule(lr){6-7} \cmidrule(lr){8-9} \cmidrule(lr){10-11}
\textbf{Method ($\downarrow$)} $\slash$ \textbf{Costs ($\$$)} & Training & Inference & Training & Inference & Training & Inference & Training & Inference & Training & Inference \\\midrule
\texttt{gpt-3.5-turbo}~\cite{chatgpt} & -- & 1.37 & -- & 0.67 & -- & 0.06	 & -- & 0.16	& -- & 0.03\\
\rowcolor{violet!10} \quad +\method & 7.67 & 10.40 & 42.57 & 5.37 & 3.49 & 0.44 & 0.92 & 1.14 & 1.41 & 0.35   \\ 
\rowcolor{gray!16} \quad +Azure-SFT~\cite{SFTgpt} & 71.18 & 10.88 & 172.85 & 6.83 & 38.93 & 3.18 & 38.17 & 3.76 & 38.48 & 3.24 \\
\quad +OpenAI-SFT$^*$ & 23.07 & 32.87 & 195.45 &  16.10  & 4.01 & 3.12 & 15.76 & 1.34 & 6.77 & 1.05\\\bottomrule
\end{tabular}
\caption{Cost ($\$$) estimations of adapting black-box LLMs to biomedical QA tasks based on \texttt{gpt-35-turbo-1106}. $*$ denotes an estimated cost, as the OpenAI-SFT is not compliant with HIPAA regulations.}\label{tab:cost}
\end{table*}
Table~\ref{tab:cost} compares the cost estimations of different black-box LLM adaptation methods in the main biomedical QA tasks. 
Compared to the Microsoft OpenAI service, which achieves an average improvement of 10.11\% over the backbone LLM, \method obtains an improvement of 11.31\% at only 15.59\% of the cost during the fine-tuning stage. 
This is because \method relies on inference APIs (\$1 per 1M token) to generate candidate solutions, which is significantly less expensive than using fine-tuning APIs (\$8 per 1M token).
Moreover, customized models accessed through APIs incur 1.58$\times$ higher costs during the inference stage than \method due to the increased prices for input (\$3 per 1M tokens) and output (\$6 per 1M tokens) usage compared to the original models (\$1 per 1M tokens for input usage and \$2 per 1M tokens for output usage).

In addition, we also report an estimated cost through the OpenAI supervised fine-tuning API~\footnote{\url{https://openai.com/pricing}} without implementation due to the conflict with HIPAA compliance, which is significantly higher than \method in both the fine-tuning and inference stages.
Notably, there are differences between the Microsoft Azure OpenAI fine-tuning API service and the OpenAI fine-tuning API: 
(1) Microsoft Azure service charges based on training hours, including an additional hosting cost for model deployment, and (2) OpenAI fine-tuning API incurs a higher cost per token for both training and inference but does not include additional hosting fees.

\subsection{Parameter Efficiency}
\label{sec:pefficiency}
\begin{table}[t]
\centering
\fontsize{8}{10}\selectfont\setlength{\tabcolsep}{0.3em}
\begin{tabular}{@{}lccc@{}}
\toprule
\textbf{Dataset ($\rightarrow$)} & \multicolumn{3}{c}{\textbf{BioASQ}} \\
\cmidrule(lr){2-4} 
\textbf{Method ($\downarrow$)} $\slash$ \textbf{Memory (GiB)} & Training & Inference & Acc. ($\%$) \\\midrule
\texttt{LLaMA-2-7B}~\cite{touvron2023llama} & --	& 25.42 & 16.13\\
\rowcolor{teal!10} \quad +\method & 11.60 & 33.00 & 62.90  \\ 
 \quad +SFT-LoRA~\cite{hu2021lora} & 54.76 & 34.65 & 50.00 \\
\rowcolor{gray!16} \quad +SFT & 78.65 & 25.42 & 63.31 \\\bottomrule
\end{tabular}
\caption{GPU memory (GiB) usage estimations of adapting white-box LLMs to biomedical QA tasks. }\label{tab:mem}
\end{table}
Table~\ref{tab:mem} evaluates the GPU memory (GiB) usage of different white-box LLMs adaptation methods, including PEFT methods.
Compared to supervised fine-tuning of a \texttt{LLaMA-2-7B}~\citep{touvron2023llama}, \method achieves competitive performance while only fine-tuning a 110M-parameter model, using 14.75\% of the GPU memory.  
Compared to other parameter-efficient adaptation methods, such as LoRA~\citep{hu2021lora}, which updates approximately 170M parameters, \method demonstrates a 12.90\% improvement in model performance while utilizing only 21.18\% of the GPU memory.
We also observe \method requires a slightly higher GPU memory usage during the inference stage, as it requires loading the original model. However, this usage remains lower than that required for supervised fine-tuning or LoRA.


\subsection{Scale-up Analysis}
\begin{figure}[t]
	\centering
	\subfigure[General LMs.]{
		\includegraphics[width=0.46\linewidth]{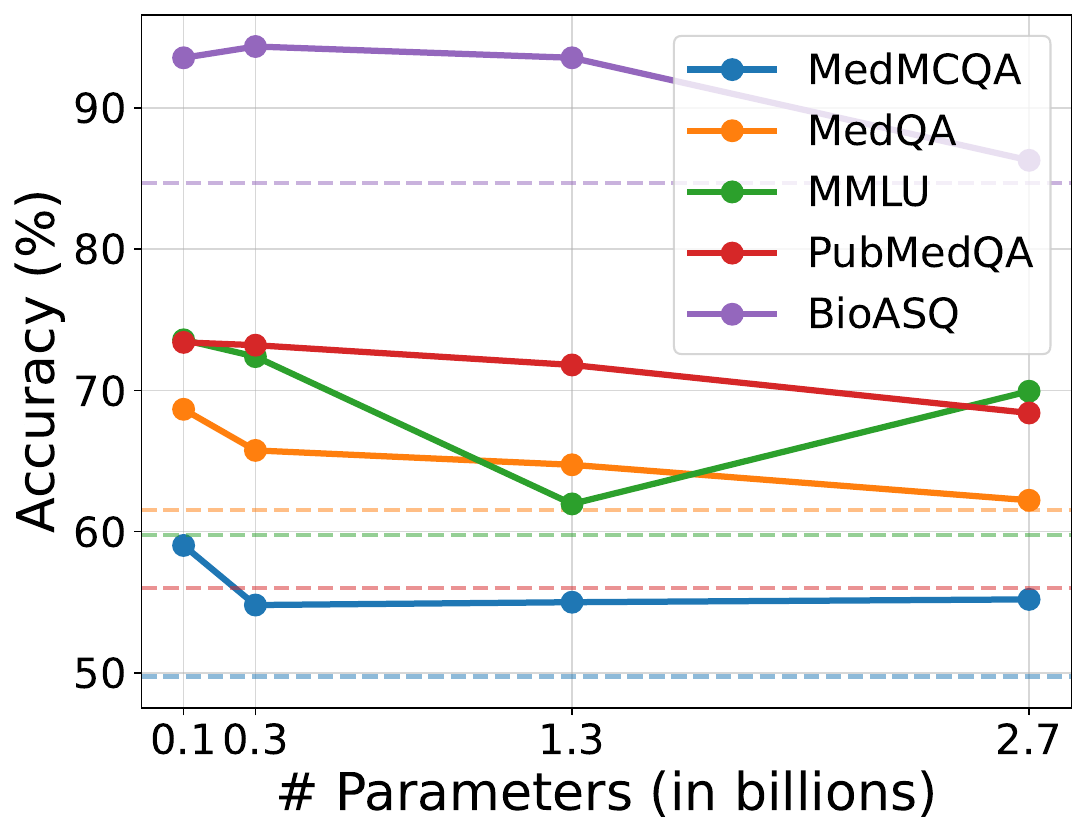}
		\label{fig:scale-general}
	} 
     \subfigure[Biomedical LMs.]{
		\includegraphics[width=0.46\linewidth]{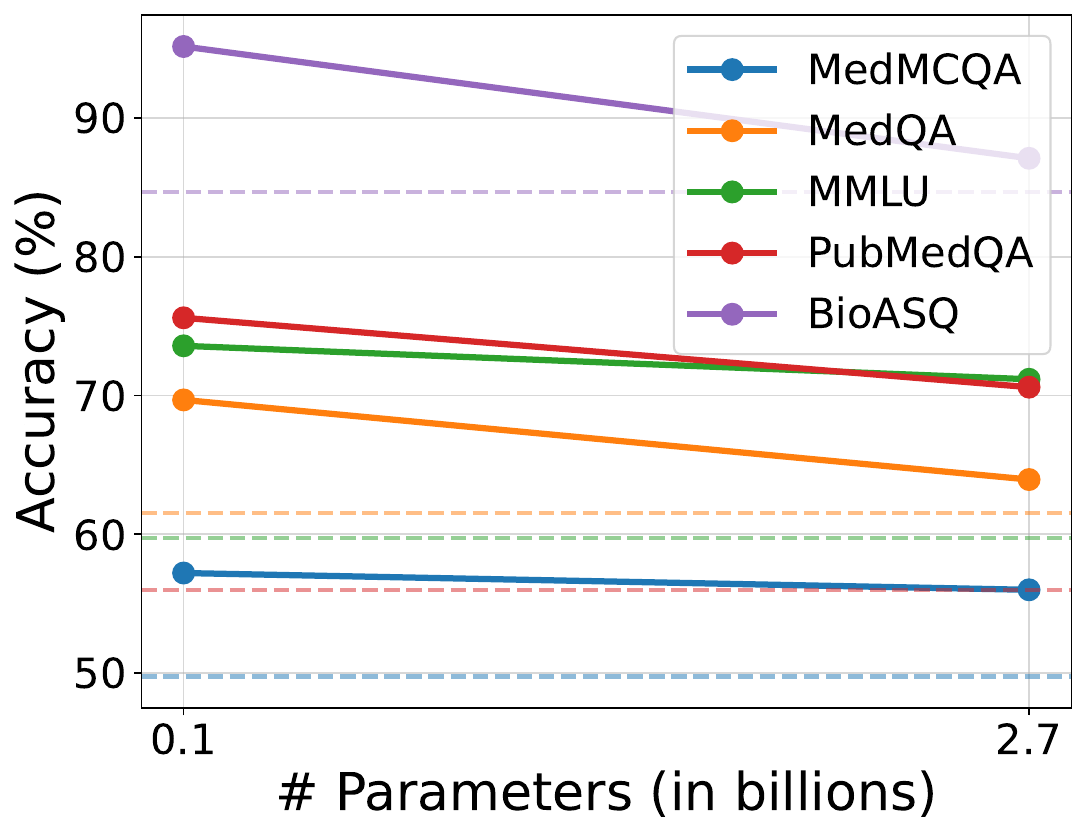}
		\label{fig:scale-bio}
	}
    \vspace{-2ex}
	\caption{Scale-up performance on multiple general and biomedical domain-specific language models (LMs) as the base LM of \method. 
 The dashed line denotes the performance of the base model, \texttt{gpt-3.5-turbo}. 
 }
\label{fig:scale}
\end{figure}

In Figure~\ref{fig:scale}, we explore the impact of scaling up the base model of \method from 110M to 2.7B parameters, utilizing both general-domain and biomedical domain-specific language models. 
Additional model details for the scale-up analysis are available in Appendix~\ref{app:scale}.
Interestingly, we observe very limited or no improvement with the increase in model size, potentially due to the following reasons: (1) \method serves as a scoring function that heavily relies on language comprehension rather than generative capabilities, which is a natural fit to encoder-only model; and (2) the limited fine-tuning data available may allow smaller models to more effectively capture underlying patterns within the candidate solutions. Additionally, domain-specific language models exhibit slightly superior performance, likely due to the integration of more targeted knowledge during their pre-training phase.

\subsection{Effect of Learning Objectives}
\label{sec:loss}
We compare the cross-entropy loss (classification) utilized in \method with the InfoNCE loss~\cite{oord2018representation} and pairwise loss~\cite{stiennon2020learning} in Table~\ref{tab:loss} to empirically study the effect of different learning objectives. 
The pairwise loss demonstrates inferior performance compared to the classification loss, especially when the base model performs well. This is due to the limited availability of negative samples, which makes it challenging to construct positive-negative pairs. Conversely, for those with limited base performances, it is relatively easier to sample such pairs during the generation process.
In addition, the InfoNCE loss imposes even more demanding prerequisites than the pairwise loss and classification loss. It necessitates the inclusion of one positive sample and multiple negative samples within a single batch. 
We include additional loss function details in Appendix~\ref{app:loss}.
\begin{table}[t]
\centering
\fontsize{8}{10}\selectfont\setlength{\tabcolsep}{0.3em}
\resizebox{\linewidth}{!}{
\begin{tabular}{@{}lccc@{}}
\toprule
\textbf{Loss ($\downarrow$) $\slash$ Dataset ($\rightarrow$)} & \textbf{BioASQ} & \textbf{MMLU} & \textbf{MedMCQA} \\\midrule
InfoNCE~\cite{oord2018representation} & 87.90	& 69.18	& 57.43\\
Pairwise~\cite{stiennon2020learning} & 92.74	& 72.33	& \textbf{59.83}  \\ 
Cross-entropy (\textbf{Ours}) & \textbf{93.55}	& \textbf{73.58}	& 59.02 \\\bottomrule
\end{tabular}
}
\caption{Comparison of different learning objectives with \texttt{gpt-3.5-turbo} as the backbone LLM. }\label{tab:loss}
\vspace{-1ex}
\end{table}

\subsection{Effect of Training Samples}
Figure~\ref{fig:plot_data_curve} presents the effect of training samples regarding the performance gain. 
We find that \method is label-efficient, achieving noticeable performance improvements with only 40\% to 60\% of the training examples (e.g., input-label pairs). Additionally, \method reduces the dependency on costly high-quality reasoning step annotations, particularly valuable in the context of low-resource medical reasoning tasks.

\subsection{\method on Generation Tasks}
\begin{table}[t]
\centering
\fontsize{8}{10}\selectfont\setlength{\tabcolsep}{0.4em}
\begin{tabular}{@{}lcccc@{}}
\toprule
\textbf{Dataset ($\downarrow$)} & \textbf{Method ($\downarrow$)} & \textbf{BLEU} & \textbf{Rouge-1} & \textbf{Rouge-L} \\\midrule
\multirow{2}{*}{MediQA} & \texttt{gpt-3.5-turbo} & 2.697 & 0.2370 & 0.1571\\
&  \cellcolor{teal!10}\quad +\method & \cellcolor{teal!10}\textbf{3.096} & \cellcolor{teal!10}\textbf{0.2464} & \cellcolor{teal!10}\textbf{0.1591}  \\ \midrule
\multirow{2}{*}{CORD19} & \texttt{gpt-3.5-turbo} & 1.420 & 0.1672 & 0.1312 \\
& \cellcolor{teal!10}\quad +\method & \cellcolor{teal!10}\textbf{1.739} & \cellcolor{teal!10}\textbf{0.1816} & \cellcolor{teal!10}\textbf{0.1559}\\\bottomrule
\end{tabular}
\caption{Generalization of \method into medical generative tasks, including open-ended medical QA (MediQA) and clinical text summarization (CORD19). }\label{tab:gen}
\end{table}
To demonstrate the effectiveness of adapting LLMs for generative tasks, we conduct additional experiments on two medical generative tasks (see Table~\ref{tab:gen}), including open-ended question answering using MediQA~\cite{savery2020question} and text summarization with Medical$\_$CORD19~\cite{wang2020cord}.
Experimental results demonstrate that \method successfully improves the black-box LLM \texttt{GPT-3.5-turbo} for both tasks, demonstrating its generalizability and effectiveness in domain adaptation for medical generative applications.

\subsection{Human Study on Adaptation Score}
\begin{figure}[t]
    \centering    
    \begin{minipage}{0.235\textwidth}
    \centering
    \includegraphics[width=0.99\textwidth]{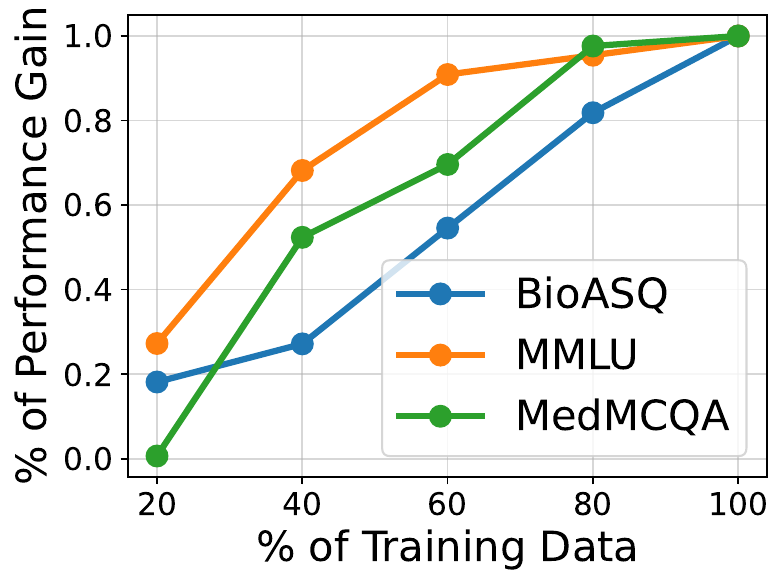}
    \caption{Label Efficiency.}
    \label{fig:plot_data_curve}
    \end{minipage}
    \hfill
    \begin{minipage}{0.235\textwidth}
    \centering
    \includegraphics[width=0.99\textwidth]{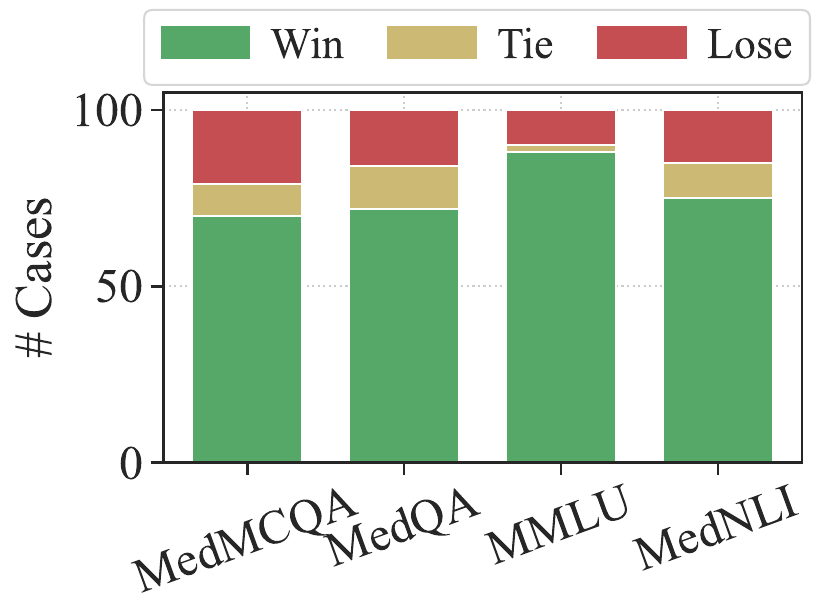}
    \caption{Human Study.}
    \label{fig:human_study}
    \end{minipage}
\end{figure}

Following the guideline in Appendix~\ref{app:human}, we conduct human studies to measure the alignment between adaptation scores generated by \method and human preferences. We randomly select 100 instances from two distinct tasks (QA and NLI) in four datasets (MedMCQA, MedQA, MMLU, and MedNLI) for a thorough evaluation.
From Figure~\ref{fig:human_study}, we observe that \method achieves a relatively high win rate across multiple datasets, indicating a meaningful adaptation score that aligns with human preferences. We present more case studies with adaptation scores in Appendix~\ref{app:case}.

\section{Related Works}
\paragraph{Train-Time Adaptation of LLMs for Biomedical Domains.}
To enhance the biomedical capabilities of LLMs, prior research has employed large-scale domain-specific corpora to customize white-box LLMs for medical reasoning, including: 
(1) \emph{Pre-Training}, such as BioGPT~\citep{luo2022biogpt}, Meditron~\citep{chen2023meditron}, Biomistral~\citep{labrak2024biomistral} and BioMedLM~\cite{bolton2024biomedlm};
(2) \emph{Fine-Tuning}, such as MedAlpaca~\citep{han2023medalpaca}, ChatDoctor~\citep{yunxiang2023chatdoctor}, PMC-LLaMA~\citep{wu2024pmc}; and
(3) \emph{Parameter-Efficient Fine-Tuning (PEFT)}, such as Clinical LLaMA-LoRA~\citep{gema2023parameter}.
Pre-training or fine-tuning LLMs necessitates substantial computational resources, particularly as model sizes continue to increase, which may not be readily accessible to academic or medical researchers~\citep{bolton2024assessing}. 
For example, Biomistral~\citep{labrak2024biomistral} requires approximately 5K computing hours of A100 80GB GPU.
While PEFT-based adaptation methods~\citep{gema2023parameter} are more efficient as they only update a small subset of parameters, they might yield suboptimal performance.
Alternatively, \method offers a different test-time solution by leveraging the emerging generative capabilities of LLMs, avoiding exclusive training on large-scale domain-specific data while utilizing significantly fewer model parameters.



\paragraph{Test-Time Adaptation of LLMs.}
Test-time adaptation involves customizing models to test data that may differ in distribution from the original training data~\citep{liang2023comprehensive,ye2023effective}. 
Existing methods for test-time adaptation of LLMs towards medical reasoning include:
(1) \emph{Prompting}-based methods, such as Med-PaLM~\citep{nori2023capabilities}; and 
(2) \emph{Retrieval-Augmented Generation (RAG)}-based methods, such as MedRAG~\citep{xiong2024benchmarking} and Self-BioRAG~\citep{jeong2024improving}.
\method introduces a third option for test-time adaptation of LLMs in medical reasoning by training a small adapter to score the candidate solutions generated by large models, thereby eliminating the need for fine-tuning the original LLM while still effectively facilitating target domain adaptation.

\section{Conclusion}

In this study, we propose \method to address a unique challenge in adopting LLMs in real-world clinical scenarios with limited computational resources and strict privacy requirements.
\texttt{\method} strikes a balance between effective model adaptation and reasonable computational costs by employing a BERT-sized language model as an adapter to select candidate solutions generated by larger LLMs, thereby obviating the need to fine-tune the entire LLMs.
\method may offer a unified and generalizable practical solution for effectively, privacy-preservingly, cost-effectively, and transparently adapting LLMs to real-world biomedical research and practice.

\section*{Limitations}
In this work, we propose \method for test-time adaptation of LLMs in medical reasoning applications. However, we have identified several limitations of \method:
(1) \textbf{Access to Label Information}: \method still requires access to task-specific labeled data to fine-tune a small adapter. This may not be feasible in some real-world scenarios where label information is restricted or unavailable.
(2) \textbf{On-Device Inference}: In the adaptation of black-box LLMs, the fine-tuning process does not share any data with third parties through APIs; however, it cannot handle queries involving sensitive or patient-identifiable information during the inference stage. Furthermore, the extensive parameters of black-box LLMs post challenges for on-device inference.
(3) \textbf{Resource Limitations}: Due to restricted access to fine-tuning API services and budget constraints, our experiments with black-box fine-tuning are limited to GPT-3.5-Turbo via the Microsoft Azure fine-tuning API service.

\section*{Ethics Statements} 
In strict adherence to the PhysioNet Credentialed Health Data Use Agreement 1.5.0\footnote{\url{https://physionet.org/about/licenses/physionet-credentialed-health-data-license-150/}}, we expressly forbid the dissemination of confidential patient information to any third party, including via online services such as APIs. To guarantee the responsible utilization of Azure OpenAI Service in accordance with the guideline\footnote{\url{https://physionet.org/news/post/gpt-responsible-use}}, we have deliberately withdrawn from the human review process by submitting the Azure OpenAI Additional Use Case Form\footnote{\url{https://aka.ms/oai/additionalusecase}}.
It effectively precludes third parties from accessing and processing protected health information (PHI) for any purpose. We maintain a rigorous monitoring process to ensure our compliance with these guidelines and pertinent privacy legislation, thereby upholding the highest ethical standards in the use of data throughout our research.

\section*{Acknowledgments}
We thank the anonymous reviewers and area chairs for their valuable feedback. 
This research was partially supported by Accelerate Foundation Models Academic Research Initiative from Microsoft Research.
This research was also partially supported by the National Science Foundation under Award Number 2319449 and Award Number 2312502, the National Institute Of Diabetes And Digestive And Kidney Diseases of the National Institutes of Health under Award Number K25DK135913, the Emory Global Diabetes Center of the Woodruff Sciences Center, Emory University.
\bibliography{anthology, custom}

\begin{thebibliography}{62}
\expandafter\ifx\csname natexlab\endcsname\relax\def\natexlab#1{#1}\fi

\bibitem[{Anthropic(2024)}]{claude3}
Anthropic. 2024.
\newblock \href {https://www-cdn.anthropic.com/de8ba9b01c9ab7cbabf5c33b80b7bbc618857627/Model_Card_Claude_3.pdf} {The claude 3 model family: Opus, sonnet, haiku}.
\newblock \emph{Claude-3 Model Card}.

\bibitem[{Beltagy et~al.(2020)Beltagy, Peters, and Cohan}]{beltagy2020longformer}
Iz~Beltagy, Matthew~E Peters, and Arman Cohan. 2020.
\newblock \href {https://arxiv.org/abs/2004.05150} {Longformer: The long-document transformer}.
\newblock \emph{ArXiv preprint}, abs/2004.05150.

\bibitem[{Ben~Abacha et~al.(2019)Ben~Abacha, Shivade, and Demner-Fushman}]{mediqa-rqe}
Asma Ben~Abacha, Chaitanya Shivade, and Dina Demner-Fushman. 2019.
\newblock \href {https://doi.org/10.18653/v1/W19-5039} {Overview of the {MEDIQA} 2019 shared task on textual inference, question entailment and question answering}.
\newblock In \emph{Proceedings of the 18th BioNLP Workshop and Shared Task}, pages 370--379, Florence, Italy. Association for Computational Linguistics.

\bibitem[{Bolton et~al.(2024{\natexlab{a}})Bolton, Venigalla, Yasunaga, Hall, Xiong, Lee, Daneshjou, Frankle, Liang, Carbin et~al.}]{bolton2024biomedlm}
Elliot Bolton, Abhinav Venigalla, Michihiro Yasunaga, David Hall, Betty Xiong, Tony Lee, Roxana Daneshjou, Jonathan Frankle, Percy Liang, Michael Carbin, et~al. 2024{\natexlab{a}}.
\newblock \href {https://arxiv.org/abs/2403.18421} {Biomedlm: A 2.7 b parameter language model trained on biomedical text}.
\newblock \emph{ArXiv preprint}, abs/2403.18421.

\bibitem[{Bolton et~al.(2024{\natexlab{b}})Bolton, Xiong, Muralidharan, Schamroth, Muralidharan, Manning, and Daneshjou}]{bolton2024assessing}
Elliot Bolton, Betty Xiong, Vijaytha Muralidharan, Joel Schamroth, Vivek Muralidharan, Christopher~D Manning, and Roxana Daneshjou. 2024{\natexlab{b}}.
\newblock \href {https://arxiv.org/abs/2404.15894} {Assessing the potential of mid-sized language models for clinical qa}.
\newblock \emph{ArXiv preprint}, abs/2404.15894.

\bibitem[{Chen et~al.(2023{\natexlab{a}})Chen, Du, Hu, Keloth, Peng, Raja, Zhang, Lu, and Xu}]{chen2023large}
Qingyu Chen, Jingcheng Du, Yan Hu, Vipina~Kuttichi Keloth, Xueqing Peng, Kalpana Raja, Rui Zhang, Zhiyong Lu, and Hua Xu. 2023{\natexlab{a}}.
\newblock \href {https://arxiv.org/abs/2305.16326} {Large language models in biomedical natural language processing: benchmarks, baselines, and recommendations}.
\newblock \emph{ArXiv preprint}, abs/2305.16326.

\bibitem[{Chen et~al.(2023{\natexlab{b}})Chen, Cano, Romanou, Bonnet, Matoba, Salvi, Pagliardini, Fan, K{\"o}pf, Mohtashami et~al.}]{chen2023meditron}
Zeming Chen, Alejandro~Hern{\'a}ndez Cano, Angelika Romanou, Antoine Bonnet, Kyle Matoba, Francesco Salvi, Matteo Pagliardini, Simin Fan, Andreas K{\"o}pf, Amirkeivan Mohtashami, et~al. 2023{\natexlab{b}}.
\newblock \href {https://arxiv.org/abs/2311.16079} {Meditron-70b: Scaling medical pretraining for large language models}.
\newblock \emph{ArXiv preprint}, abs/2311.16079.

\bibitem[{Chung et~al.(2024)Chung, Hou, Longpre, Zoph, Tay, Fedus, Li, Wang, Dehghani, Brahma et~al.}]{chung2024scaling}
Hyung~Won Chung, Le~Hou, Shayne Longpre, Barret Zoph, Yi~Tay, William Fedus, Yunxuan Li, Xuezhi Wang, Mostafa Dehghani, Siddhartha Brahma, et~al. 2024.
\newblock Scaling instruction-finetuned language models.
\newblock \emph{Journal of Machine Learning Research}, 25(70):1--53.

\bibitem[{Gema et~al.(2023)Gema, Daines, Minervini, and Alex}]{gema2023parameter}
Aryo Gema, Luke Daines, Pasquale Minervini, and Beatrice Alex. 2023.
\newblock \href {https://arxiv.org/abs/2307.03042} {Parameter-efficient fine-tuning of llama for the clinical domain}.
\newblock \emph{ArXiv preprint}, abs/2307.03042.

\bibitem[{Han et~al.(2023)Han, Adams, Papaioannou, Grundmann, Oberhauser, L{\"o}ser, Truhn, and Bressem}]{han2023medalpaca}
Tianyu Han, Lisa~C Adams, Jens-Michalis Papaioannou, Paul Grundmann, Tom Oberhauser, Alexander L{\"o}ser, Daniel Truhn, and Keno~K Bressem. 2023.
\newblock \href {https://arxiv.org/abs/2304.08247} {Medalpaca--an open-source collection of medical conversational ai models and training data}.
\newblock \emph{ArXiv preprint}, abs/2304.08247.

\bibitem[{Hendrycks et~al.(2021)Hendrycks, Burns, Basart, Zou, Mazeika, Song, and Steinhardt}]{hendrycks2020measuring}
Dan Hendrycks, Collin Burns, Steven Basart, Andy Zou, Mantas Mazeika, Dawn Song, and Jacob Steinhardt. 2021.
\newblock \href {https://openreview.net/forum?id=d7KBjmI3GmQ} {Measuring massive multitask language understanding}.
\newblock In \emph{9th International Conference on Learning Representations}.

\bibitem[{Houlsby et~al.(2019)Houlsby, Giurgiu, Jastrzebski, Morrone, de~Laroussilhe, Gesmundo, Attariyan, and Gelly}]{houlsby2019parameter}
Neil Houlsby, Andrei Giurgiu, Stanislaw Jastrzebski, Bruna Morrone, Quentin de~Laroussilhe, Andrea Gesmundo, Mona Attariyan, and Sylvain Gelly. 2019.
\newblock \href {http://proceedings.mlr.press/v97/houlsby19a.html} {Parameter-efficient transfer learning for {NLP}}.
\newblock In \emph{Proceedings of the 36th International Conference on Machine Learning}, pages 2790--2799. {PMLR}.

\bibitem[{Hu et~al.(2022)Hu, Shen, Wallis, Allen{-}Zhu, Li, Wang, Wang, and Chen}]{hu2021lora}
Edward~J. Hu, Yelong Shen, Phillip Wallis, Zeyuan Allen{-}Zhu, Yuanzhi Li, Shean Wang, Lu~Wang, and Weizhu Chen. 2022.
\newblock \href {https://openreview.net/forum?id=nZeVKeeFYf9} {Lora: Low-rank adaptation of large language models}.
\newblock In \emph{The Tenth International Conference on Learning Representations}.

\bibitem[{Huang et~al.(2023)Huang, Liu, Zhong, Shi, and Lee}]{huang2023k}
Yangsibo Huang, Daogao Liu, Zexuan Zhong, Weijia Shi, and Yin~Tat Lee. 2023.
\newblock \href {https://arxiv.org/abs/2302.10879} {$ k $ nn-adapter: Efficient domain adaptation for black-box language models}.
\newblock \emph{ArXiv preprint}, abs/2302.10879.

\bibitem[{Jeong et~al.(2024)Jeong, Sohn, Sung, and Kang}]{jeong2024improving}
Minbyul Jeong, Jiwoong Sohn, Mujeen Sung, and Jaewoo Kang. 2024.
\newblock \href {https://arxiv.org/abs/2401.15269} {Improving medical reasoning through retrieval and self-reflection with retrieval-augmented large language models}.
\newblock \emph{ArXiv preprint}, abs/2401.15269.

\bibitem[{Jin et~al.(2021)Jin, Pan, Oufattole, Weng, Fang, and Szolovits}]{jin2021disease}
Di~Jin, Eileen Pan, Nassim Oufattole, Wei-Hung Weng, Hanyi Fang, and Peter Szolovits. 2021.
\newblock What disease does this patient have? a large-scale open domain question answering dataset from medical exams.
\newblock \emph{Applied Sciences}, 11(14):6421.

\bibitem[{Jin et~al.(2019)Jin, Dhingra, Liu, Cohen, and Lu}]{jin2019pubmedqa}
Qiao Jin, Bhuwan Dhingra, Zhengping Liu, William Cohen, and Xinghua Lu. 2019.
\newblock \href {https://doi.org/10.18653/v1/D19-1259} {{P}ub{M}ed{QA}: A dataset for biomedical research question answering}.
\newblock In \emph{Proceedings of the 2019 Conference on Empirical Methods in Natural Language Processing and the 9th International Joint Conference on Natural Language Processing (EMNLP-IJCNLP)}, pages 2567--2577, Hong Kong, China. Association for Computational Linguistics.

\bibitem[{Karmanov et~al.(2024)Karmanov, Guan, Lu, Saddik, and Xing}]{karmanov2024efficient}
Adilbek Karmanov, Dayan Guan, Shijian Lu, Abdulmotaleb~El Saddik, and Eric Xing. 2024.
\newblock \href {https://arxiv.org/abs/2403.18293} {Efficient test-time adaptation of vision-language models}.
\newblock \emph{ArXiv preprint}, abs/2403.18293.

\bibitem[{Khalifa et~al.(2023)Khalifa, Logeswaran, Lee, Lee, and Wang}]{khalifa-etal-2023-grace}
Muhammad Khalifa, Lajanugen Logeswaran, Moontae Lee, Honglak Lee, and Lu~Wang. 2023.
\newblock \href {https://doi.org/10.18653/v1/2023.findings-emnlp.1022} {{GRACE}: Discriminator-guided chain-of-thought reasoning}.
\newblock In \emph{Findings of the Association for Computational Linguistics: EMNLP 2023}, pages 15299--15328, Singapore. Association for Computational Linguistics.

\bibitem[{Kotonya and Toni(2020)}]{PUBHEALTH}
Neema Kotonya and Francesca Toni. 2020.
\newblock \href {https://doi.org/10.18653/v1/2020.emnlp-main.623} {Explainable automated fact-checking for public health claims}.
\newblock In \emph{Proceedings of the 2020 Conference on Empirical Methods in Natural Language Processing (EMNLP)}, pages 7740--7754, Online. Association for Computational Linguistics.

\bibitem[{Labrak et~al.(2024)Labrak, Bazoge, Morin, Gourraud, Rouvier, and Dufour}]{labrak2024biomistral}
Yanis Labrak, Adrien Bazoge, Emmanuel Morin, Pierre-Antoine Gourraud, Mickael Rouvier, and Richard Dufour. 2024.
\newblock \href {https://arxiv.org/abs/2402.10373} {Biomistral: A collection of open-source pretrained large language models for medical domains}.
\newblock \emph{ArXiv preprint}, abs/2402.10373.

\bibitem[{Li et~al.(2023{\natexlab{a}})Li, Lin, Zhang, Fu, Chen, Lou, and Chen}]{li-etal-2023-making}
Yifei Li, Zeqi Lin, Shizhuo Zhang, Qiang Fu, Bei Chen, Jian-Guang Lou, and Weizhu Chen. 2023{\natexlab{a}}.
\newblock \href {https://doi.org/10.18653/v1/2023.acl-long.291} {Making language models better reasoners with step-aware verifier}.
\newblock In \emph{Proceedings of the 61st Annual Meeting of the Association for Computational Linguistics (Volume 1: Long Papers)}, pages 5315--5333, Toronto, Canada. Association for Computational Linguistics.

\bibitem[{Li et~al.(2022)Li, Wehbe, Ahmad, Wang, and Luo}]{li2022clinical}
Yikuan Li, Ramsey~M Wehbe, Faraz~S Ahmad, Hanyin Wang, and Yuan Luo. 2022.
\newblock \href {https://arxiv.org/abs/2201.11838} {Clinical-longformer and clinical-bigbird: Transformers for long clinical sequences}.
\newblock \emph{ArXiv preprint}, abs/2201.11838.

\bibitem[{Li et~al.(2023{\natexlab{b}})Li, Bubeck, Eldan, Del~Giorno, Gunasekar, and Lee}]{li2023textbooks}
Yuanzhi Li, S{\'e}bastien Bubeck, Ronen Eldan, Allie Del~Giorno, Suriya Gunasekar, and Yin~Tat Lee. 2023{\natexlab{b}}.
\newblock \href {https://arxiv.org/abs/2309.05463} {Textbooks are all you need ii: phi-1.5 technical report}.
\newblock \emph{ArXiv preprint}, abs/2309.05463.

\bibitem[{Liang et~al.(2023)Liang, He, and Tan}]{liang2023comprehensive}
Jian Liang, Ran He, and Tieniu Tan. 2023.
\newblock \href {https://arxiv.org/abs/2303.15361} {A comprehensive survey on test-time adaptation under distribution shifts}.
\newblock \emph{ArXiv preprint}, abs/2303.15361.

\bibitem[{Liu et~al.(2024)Liu, Han, Wang, Tsvetkov, Choi, and Smith}]{liu2024tuning}
Alisa Liu, Xiaochuang Han, Yizhong Wang, Yulia Tsvetkov, Yejin Choi, and Noah~A Smith. 2024.
\newblock \href {https://arxiv.org/abs/2401.08565} {Tuning language models by proxy}.
\newblock \emph{ArXiv preprint}, abs/2401.08565.

\bibitem[{Luccioni et~al.(2023)Luccioni, Viguier, and Ligozat}]{luccioni2023estimating}
Alexandra~Sasha Luccioni, Sylvain Viguier, and Anne-Laure Ligozat. 2023.
\newblock Estimating the carbon footprint of bloom, a 176b parameter language model.
\newblock \emph{Journal of Machine Learning Research}, 24(253):1--15.

\bibitem[{Lukas et~al.(2023)Lukas, Salem, Sim, Tople, Wutschitz, and Zanella-B{\'e}guelin}]{lukas2023analyzing}
Nils Lukas, Ahmed Salem, Robert Sim, Shruti Tople, Lukas Wutschitz, and Santiago Zanella-B{\'e}guelin. 2023.
\newblock Analyzing leakage of personally identifiable information in language models.
\newblock In \emph{IEEE Symposium on Security and Privacy}, pages 346--363. IEEE.

\bibitem[{Luo et~al.(2022)Luo, Sun, Xia, Qin, Zhang, Poon, and Liu}]{luo2022biogpt}
Renqian Luo, Liai Sun, Yingce Xia, Tao Qin, Sheng Zhang, Hoifung Poon, and Tie-Yan Liu. 2022.
\newblock Biogpt: generative pre-trained transformer for biomedical text generation and mining.
\newblock \emph{Briefings in bioinformatics}, 23(6):bbac409.

\bibitem[{Luo et~al.(2023)Luo, Zhang, Fan, Yang, Wu, Qiao, and Nie}]{luo2023biomedgpt}
Yizhen Luo, Jiahuan Zhang, Siqi Fan, Kai Yang, Yushuai Wu, Mu~Qiao, and Zaiqing Nie. 2023.
\newblock \href {https://arxiv.org/abs/2308.09442} {Biomedgpt: Open multimodal generative pre-trained transformer for biomedicine}.
\newblock \emph{ArXiv preprint}, abs/2308.09442.

\bibitem[{Marks and Haupt(2023)}]{marks2023ai}
Mason Marks and Claudia~E Haupt. 2023.
\newblock Ai chatbots, health privacy, and challenges to hipaa compliance.
\newblock \emph{Jama}.

\bibitem[{Meta-AI(2024)}]{llama3}
Meta-AI. 2024.
\newblock \href {https://github.com/meta-llama/llama3/blob/main/MODEL_CARD.md} {Llama 3 model card}.

\bibitem[{Nori et~al.(2023)Nori, King, McKinney, Carignan, and Horvitz}]{nori2023capabilities}
Harsha Nori, Nicholas King, Scott~Mayer McKinney, Dean Carignan, and Eric Horvitz. 2023.
\newblock \href {https://arxiv.org/abs/2303.13375} {Capabilities of gpt-4 on medical challenge problems}.
\newblock \emph{ArXiv preprint}, abs/2303.13375.

\bibitem[{Oord et~al.(2018)Oord, Li, and Vinyals}]{oord2018representation}
Aaron van~den Oord, Yazhe Li, and Oriol Vinyals. 2018.
\newblock Representation learning with contrastive predictive coding.
\newblock \emph{arXiv preprint arXiv:1807.03748}.

\bibitem[{OpenAI(2022)}]{chatgpt}
OpenAI. 2022.
\newblock \href {https://openai.com/blog/chatgpt} {Introducing chatgpt}.
\newblock \emph{OpenAI Blog}.

\bibitem[{OpenAI(2023)}]{achiam2023gpt}
OpenAI. 2023.
\newblock \href {https://arxiv.org/abs/2303.08774} {Gpt-4 technical report}.
\newblock \emph{ArXiv preprint}, abs/2303.08774.

\bibitem[{Ormazabal et~al.(2023)Ormazabal, Artetxe, and Agirre}]{ormazabal2023comblm}
Aitor Ormazabal, Mikel Artetxe, and Eneko Agirre. 2023.
\newblock Comblm: Adapting black-box language models through small fine-tuned models.
\newblock In \emph{Proceedings of the 2023 Conference on Empirical Methods in Natural Language Processing}, pages 2961--2974.

\bibitem[{Pal et~al.(2022)Pal, Umapathi, and Sankarasubbu}]{pal2022medmcqa}
Ankit Pal, Logesh~Kumar Umapathi, and Malaikannan Sankarasubbu. 2022.
\newblock Medmcqa: A large-scale multi-subject multi-choice dataset for medical domain question answering.
\newblock In \emph{Conference on health, inference, and learning}, pages 248--260. PMLR.

\bibitem[{Peng et~al.(2023)Peng, Wu, Allard, Kilpatrick, and Heidel}]{SFTgpt}
Andrew Peng, Machael Wu, John Allard, Logan Kilpatrick, and Steven Heidel. 2023.
\newblock \href {https://openai.com/blog/gpt-3-5-turbo-fine-tuning-and-api-updates} {Gpt-3.5 turbo fine-tuning and api updates}.
\newblock \emph{OpenAI Blog}.

\bibitem[{Rasley et~al.(2020)Rasley, Rajbhandari, Ruwase, and He}]{rasley2020deepspeed}
Jeff Rasley, Samyam Rajbhandari, Olatunji Ruwase, and Yuxiong He. 2020.
\newblock \href {https://dl.acm.org/doi/10.1145/3394486.3406703} {Deepspeed: System optimizations enable training deep learning models with over 100 billion parameters}.
\newblock In \emph{The 26th {ACM} {SIGKDD} Conference on Knowledge Discovery and Data Mining}, pages 3505--3506. {ACM}.

\bibitem[{Savery et~al.(2020)Savery, Abacha, Gayen, and Demner-Fushman}]{savery2020question}
Max Savery, Asma~Ben Abacha, Soumya Gayen, and Dina Demner-Fushman. 2020.
\newblock Question-driven summarization of answers to consumer health questions.
\newblock \emph{Scientific Data}, 7(1):322.

\bibitem[{Shi et~al.(2024)Shi, Ajith, Xia, Huang, Liu, Blevins, Chen, and Zettlemoyer}]{shi2023detecting}
Weijia Shi, Anirudh Ajith, Mengzhou Xia, Yangsibo Huang, Daogao Liu, Terra Blevins, Danqi Chen, and Luke Zettlemoyer. 2024.
\newblock \href {https://openreview.net/forum?id=zWqr3MQuNs} {Detecting pretraining data from large language models}.
\newblock In \emph{The Twelfth International Conference on Learning Representations}.

\bibitem[{Shivade(2017)}]{mednli}
Chaitanya Shivade. 2017.
\newblock \href {https://doi.org/10.13026/C2RS98} {Mednli — a natural language inference dataset for the clinical domain}.

\bibitem[{Singhal et~al.(2023)Singhal, Azizi, Tu, Mahdavi, Wei, Chung, Scales, Tanwani, Cole-Lewis, Pfohl et~al.}]{singhal2023large}
Karan Singhal, Shekoofeh Azizi, Tao Tu, S~Sara Mahdavi, Jason Wei, Hyung~Won Chung, Nathan Scales, Ajay Tanwani, Heather Cole-Lewis, Stephen Pfohl, et~al. 2023.
\newblock Large language models encode clinical knowledge.
\newblock \emph{Nature}, 620(7972):172--180.

\bibitem[{Stiennon et~al.(2020)Stiennon, Ouyang, Wu, Ziegler, Lowe, Voss, Radford, Amodei, and Christiano}]{stiennon2020learning}
Nisan Stiennon, Long Ouyang, Jeffrey Wu, Daniel Ziegler, Ryan Lowe, Chelsea Voss, Alec Radford, Dario Amodei, and Paul~F Christiano. 2020.
\newblock Learning to summarize with human feedback.
\newblock \emph{Advances in Neural Information Processing Systems}, 33:3008--3021.

\bibitem[{Sun et~al.(2024)Sun, Zhuang, Wei, Zhang, and Dai}]{sun2024bbox}
Haotian Sun, Yuchen Zhuang, Wei Wei, Chao Zhang, and Bo~Dai. 2024.
\newblock \href {https://openreview.net/forum?id=jdRIaUu3xY} {{BB}ox-adapter: Lightweight adapting for black-box large language models}.
\newblock In \emph{Forty-first International Conference on Machine Learning}.

\bibitem[{Team et~al.(2023)Team, Anil, Borgeaud, Wu, Alayrac, Yu, Soricut, Schalkwyk, Dai, Hauth et~al.}]{team2023gemini}
Gemini Team, Rohan Anil, Sebastian Borgeaud, Yonghui Wu, Jean-Baptiste Alayrac, Jiahui Yu, Radu Soricut, Johan Schalkwyk, Andrew~M Dai, Anja Hauth, et~al. 2023.
\newblock \href {https://arxiv.org/abs/2312.11805} {Gemini: a family of highly capable multimodal models}.
\newblock \emph{ArXiv preprint}, abs/2312.11805.

\bibitem[{Touvron et~al.(2023)Touvron, Martin, Stone, Albert, Almahairi, Babaei, Bashlykov, Batra, Bhargava, Bhosale et~al.}]{touvron2023llama}
Hugo Touvron, Louis Martin, Kevin Stone, Peter Albert, Amjad Almahairi, Yasmine Babaei, Nikolay Bashlykov, Soumya Batra, Prajjwal Bhargava, Shruti Bhosale, et~al. 2023.
\newblock \href {https://arxiv.org/abs/2307.09288} {Llama 2: Open foundation and fine-tuned chat models}.
\newblock \emph{ArXiv preprint}, abs/2307.09288.

\bibitem[{Tsatsaronis et~al.(2015)Tsatsaronis, Balikas, Malakasiotis, Partalas, Zschunke, Alvers, Weissenborn, Krithara, Petridis, Polychronopoulos et~al.}]{tsatsaronis2015overview}
George Tsatsaronis, Georgios Balikas, Prodromos Malakasiotis, Ioannis Partalas, Matthias Zschunke, Michael~R Alvers, Dirk Weissenborn, Anastasia Krithara, Sergios Petridis, Dimitris Polychronopoulos, et~al. 2015.
\newblock An overview of the bioasq large-scale biomedical semantic indexing and question answering competition.
\newblock \emph{BMC bioinformatics}, 16:1--28.

\bibitem[{Wang et~al.(2023{\natexlab{a}})Wang, Liu, Yang, Guo, Wu, and Liu}]{wang2023ethical}
Changyu Wang, Siru Liu, Hao Yang, Jiulin Guo, Yuxuan Wu, and Jialin Liu. 2023{\natexlab{a}}.
\newblock Ethical considerations of using chatgpt in health care.
\newblock \emph{Journal of Medical Internet Research}, 25:e48009.

\bibitem[{Wang et~al.(2024)Wang, Gao, Dantona, Hull, and Sun}]{wang2024drg}
Hanyin Wang, Chufan Gao, Christopher Dantona, Bryan Hull, and Jimeng Sun. 2024.
\newblock Drg-llama: tuning llama model to predict diagnosis-related group for hospitalized patients.
\newblock \emph{npj Digital Medicine}, 7(1):16.

\bibitem[{Wang et~al.(2020)Wang, Lo, Chandrasekhar, Reas, Yang, Burdick, Eide, Funk, Katsis, Kinney et~al.}]{wang2020cord}
Lucy~Lu Wang, Kyle Lo, Yoganand Chandrasekhar, Russell Reas, Jiangjiang Yang, Doug Burdick, Darrin Eide, Kathryn Funk, Yannis Katsis, Rodney~Michael Kinney, et~al. 2020.
\newblock Cord-19: The covid-19 open research dataset.
\newblock In \emph{Proceedings of the 1st Workshop on NLP for COVID-19 at ACL 2020}.

\bibitem[{Wang et~al.(2023{\natexlab{b}})Wang, Wei, Schuurmans, Le, Chi, Narang, Chowdhery, and Zhou}]{wang2023selfconsistency}
Xuezhi Wang, Jason Wei, Dale Schuurmans, Quoc~V Le, Ed~H. Chi, Sharan Narang, Aakanksha Chowdhery, and Denny Zhou. 2023{\natexlab{b}}.
\newblock \href {https://openreview.net/forum?id=1PL1NIMMrw} {Self-consistency improves chain of thought reasoning in language models}.
\newblock In \emph{The Eleventh International Conference on Learning Representations}.

\bibitem[{Wei et~al.(2022{\natexlab{a}})Wei, Bosma, Zhao, Guu, Yu, Lester, Du, Dai, and Le}]{wei2021finetuned}
Jason Wei, Maarten Bosma, Vincent~Y. Zhao, Kelvin Guu, Adams~Wei Yu, Brian Lester, Nan Du, Andrew~M. Dai, and Quoc~V. Le. 2022{\natexlab{a}}.
\newblock \href {https://openreview.net/forum?id=gEZrGCozdqR} {Finetuned language models are zero-shot learners}.
\newblock In \emph{The Tenth International Conference on Learning Representations}. OpenReview.net.

\bibitem[{Wei et~al.(2022{\natexlab{b}})Wei, Wang, Schuurmans, Bosma, Xia, Chi, Le, Zhou et~al.}]{wei2022chain}
Jason Wei, Xuezhi Wang, Dale Schuurmans, Maarten Bosma, Fei Xia, Ed~Chi, Quoc~V Le, Denny Zhou, et~al. 2022{\natexlab{b}}.
\newblock Chain-of-thought prompting elicits reasoning in large language models.
\newblock \emph{Advances in neural information processing systems}, 35:24824--24837.

\bibitem[{Wu et~al.(2024)Wu, Lin, Zhang, Zhang, Xie, and Wang}]{wu2024pmc}
Chaoyi Wu, Weixiong Lin, Xiaoman Zhang, Ya~Zhang, Weidi Xie, and Yanfeng Wang. 2024.
\newblock Pmc-llama: toward building open-source language models for medicine.
\newblock \emph{Journal of the American Medical Informatics Association}, page ocae045.

\bibitem[{Xiong et~al.(2024)Xiong, Jin, Lu, and Zhang}]{xiong2024benchmarking}
Guangzhi Xiong, Qiao Jin, Zhiyong Lu, and Aidong Zhang. 2024.
\newblock \href {https://arxiv.org/abs/2402.13178} {Benchmarking retrieval-augmented generation for medicine}.
\newblock \emph{ArXiv preprint}, abs/2402.13178.

\bibitem[{Xu et~al.(2023)Xu, Xu, Wang, Liu, Zhu, and McAuley}]{xu2023small}
Canwen Xu, Yichong Xu, Shuohang Wang, Yang Liu, Chenguang Zhu, and Julian McAuley. 2023.
\newblock Small models are valuable plug-ins for large language models.
\newblock \emph{arXiv preprint arXiv:2305.08848}.

\bibitem[{Ye et~al.(2023)Ye, Sun, Arik, and Pfister}]{ye2023effective}
Xi~Ye, Ruoxi Sun, Sercan~{\"O} Arik, and Tomas Pfister. 2023.
\newblock \href {https://arxiv.org/abs/2311.09533} {Effective large language model adaptation for improved grounding}.
\newblock \emph{ArXiv preprint}, abs/2311.09533.

\bibitem[{Yunxiang et~al.(2023)Yunxiang, Zihan, Kai, Ruilong, and You}]{yunxiang2023chatdoctor}
Li~Yunxiang, Li~Zihan, Zhang Kai, Dan Ruilong, and Zhang You. 2023.
\newblock \href {https://arxiv.org/abs/2303.14070} {Chatdoctor: A medical chat model fine-tuned on llama model using medical domain knowledge}.
\newblock \emph{ArXiv preprint}, abs/2303.14070.

\bibitem[{Zancato et~al.(2023)Zancato, Achille, Liu, Trager, Perera, and Soatto}]{zancato2023train}
Luca Zancato, Alessandro Achille, Tian~Yu Liu, Matthew Trager, Pramuditha Perera, and Stefano Soatto. 2023.
\newblock Train/test-time adaptation with retrieval.
\newblock In \emph{Proceedings of the IEEE/CVF Conference on Computer Vision and Pattern Recognition}, pages 15911--15921.

\bibitem[{Zhang et~al.(2023)Zhang, Tian, Yang, Chen, Li, and Petzold}]{zhang2023alpacare}
Xinlu Zhang, Chenxin Tian, Xianjun Yang, Lichang Chen, Zekun Li, and Linda~Ruth Petzold. 2023.
\newblock \href {https://arxiv.org/abs/2310.14558} {Alpacare: Instruction-tuned large language models for medical application}.
\newblock \emph{ArXiv preprint}, abs/2310.14558.

\end{thebibliography}

\appendix

\section{Dataset Details}
\label{app:dataset}

We evaluate the domain adaptation capabilities of both white-box and black-box LLMs in medical reasoning tasks using five biomedical QA and three additional biomedical NLP datasets.
We have selected these datasets due to their extensive utilization in assessing the language comprehension and reasoning capabilities of LLMs in the medical domain
~\citep{bolton2024biomedlm,xiong2024benchmarking,luo2023biomedgpt,jeong2024improving}.
Dataset statistics are available in Table~\ref{tab:data}. 

\begin{table}[ht]
\centering
\resizebox{\linewidth}{!}{
\begin{tabular}{@{}lccc@{}}
\toprule
\textbf{Dataset}        & \textbf{\# Train}     & \textbf{\# Test}     & \textbf{Source}      \\ \midrule
\textbf{MedMCQA}~\citep{pal2022medmcqa}          & 3000          &   4183  & Exam                            \\ 
\textbf{MedQA}~\citep{jin2021disease}          & 10178             &   1273   & Exam                             \\ 
\textbf{MMLU}~\citep{hendrycks2020measuring}          & 1299           & 163  & Exam  \\
\textbf{PubMedQA}~\citep{jin2019pubmedqa}      & 450        &     500     & Literature                           \\
\textbf{BioASQ}~\citep{tsatsaronis2015overview}          & 494              & 124   &  Literature \\ 
\midrule
\textbf{MedNLI}~\citep{mednli} & 11232 & 1422 & Patient Query\\
\textbf{MediQA-RQE}~\citep{mediqa-rqe} & 8588 & 302 & Patient Query\\
\textbf{PubHealth}~\citep{PUBHEALTH} & 9804 & 1231 & Literature \\
\bottomrule
\end{tabular}
}
\caption{Dataset statistics.}
\label{tab:data}
\end{table}

\subsection{Biomedical QA Dataset Details}

\paragraph{MedMCQA.} 
MedMCQA\footnote{\url{https://medmcqa.github.io}}~\citep{pal2022medmcqa} is a large-scale and comprehensive dataset for multi-choice (four-option) medical question answering. It is derived from real-world medical entrance exam questions (Indian AIIMS and NEET-PG) and consists of over 194,000 high-quality medical questions. These questions cover 2,400 healthcare topics and 21 medical subjects, exhibiting a wide range of topical diversity. The average token length is 12.77. 

\paragraph{MedQA.}
MedQA\footnote{\url{https://github.com/jind11/MedQA}}~\citep{jin2021disease} is a multi-choice question-answering dataset collected from the professional medical board exam, the United States Medical License Exams (USMLE). It comprises 12,723 questions sourced from a comprehensive collection of 18 English medical textbooks that have been extensively utilized by medical students and USMLE candidates.
Questions in MedQA cover a wide range of topics in clinical medicine, necessitating responses with professional expertise and complex multi-hop reasoning across multiple pieces of evidence. The average question and option length is 116.6 and 3.5, respectively. 

\paragraph{MMLU-Med.}
MMLU\footnote{\url{https://github.com/hendrycks/test}}~\citep{hendrycks2020measuring} is a comprehensive multi-task language understanding test dataset that encompasses 57 tasks across various domains such as mathematics, history, computer science, law, and \etc In our experiments, we specifically focus on a subset of seven medical reasoning-related tasks~\citep{singhal2023large}, including clinical knowledge, college biology, college medicine, high school biology, medical genetics, professional medicine, and virology. 

\paragraph{PubMedQA.}
PubMedQA\footnote{\url{https://pubmedqa.github.io}}~\citep{jin2019pubmedqa} is a biomedical question and answering dataset derived from PubMed abstracts. It contains 1k expert-annotated multi-choice question-and-answer samples based on 211.3k PubMed articles. The task of PubMedQA is to provide answers to research questions with yes/no/maybe responses based on the corresponding abstracts. The average question and context length is 14.4 and 238.9, respectively.

\paragraph{BioASQ.}
BioASQ\footnote{\url{https://github.com/AKSW/BioASQ-AT}}~\citep{tsatsaronis2015overview} is a large-scale biomedical semantic indexing and question-answering dataset. It includes tasks related to information retrieval (Task A) and machine reading comprehension (Task B).
Similar to PubMedQA~\citep{jin2019pubmedqa}, BioASQ leverages biomedical scientific articles, providing text fragments that serve as the ground truth for machine reading comprehension.
Following~\citet{xiong2024benchmarking}, we focus on 618 machine reading comprehension questions (Task B) with binary (yes/no) answers from the most recent five years (from 2019 to 2023). The average token length of each question is 17.

\subsection{Additional Biomedical Dataset Details}

\paragraph{MedNLI.}
MedNLI\footnote{\url{https://jgc128.github.io/mednli/}}~\citep{mednli} is a collection of natural language inference tasks for ascertaining whether a hypothesis can be deduced from a given premise. It is derived from MIMIC-III and annotated by medical professionals. It comprises 14,049 distinct sentence pairs grounded in the medical history of patients.

\paragraph{MediQA-RQE.}
MediQA-RQE\footnote{\url{https://sites.google.com/view/mediqa2019}}~\citep{mediqa-rqe} is a comprehensive compilation of biomedical NLP tasks designed to facilitate the recognition of question entailment. It consists of 8,588 pairs of medical questions, with the primary objective being the identification of entailment between two questions in the context of question answering.

\paragraph{PubHealth.}
PubHealth\footnote{\url{https://github.com/neemakot/Health-Fact-Checking}}~\citep{PUBHEALTH} is a comprehensive dataset designed for automated fact-checking of public health claims. Each instance in the PUBHEALTH dataset is assigned a veracity label, indicating whether it is \emph{true}, \emph{false}, \emph{unproven}, or a \emph{mixture}. It comprises 11.8K distinct claims related to public health and health policy, obtained from multiple health information websites or news journals.

\section{Implementation Details}
\label{app:prompt}
\subsection{Additional implementation details}
\paragraph{Black-Box LLM Adaptation.}
For black-box LLM adaptation, \texttt{gpt-3.5-turbo} (version 1106) serves as the main backbone LLM.
We also adapt \texttt{gpt-4} (version 1106) for a comprehensive evaluation. 
During the evaluation of Azure-SFT, certain questions and answers may be filtered by the Azure content filter to ensure the safety of the content generated. In order to avoid any potential bias caused by these filtered questions, we exclude them from the evaluation process to maintain the integrity of our assessments.

\paragraph{White-Box LLM Adaptation.}
For white-box LLM adaptation, we leverage \texttt{LLaMA-2-7B} as the backbone LLM.
During the fine-tuning phase, learning rates are set to $2e-5$ for \method and $2e-4$ for supervised fine-tuning and LoRA~\citep{hu2021lora}, respectively. The global batch size is maintained at 8 for all white-box adaptation experiments.
To maintain the same model size in the case of LoRA, we train the base LLM with $r=1024$, $\alpha=2048$, bfloat16 quantization, and DeepSpeed gradient checkpointing~\citep{rasley2020deepspeed}.

\paragraph{Hardware Details.}
All experiments are conducted on four NVIDIA A100 GPUs, accommodating a maximum sequence length of $512$ tokens.

\subsection{Prompt Templates}
\subsubsection{MedMCQA}
The prompting format for MedMCQA dataset is listed as follows:
\vspace{1ex}
\begin{Verbatim}
Use the step-by-step method as shown in the 
    example to answer the question. You should 
    give the explanation steps and final answer 
    based on the provided context.

Example:
Q: What is the most probable poal of entry of 
    Aspergillus? (A) Puncture wound, (B) Blood, 
    (C) Lungs, (D) Gastrointestinal tract 
A: Aspergillus species are widely distributed 
    on decaying plants, producing chains of 
    conidia. 
Aspergillus species unlike Candida species do 
    not form the pa of normal flora of humans. 
They are ubiquitous in the environment; hence 
    transmission of infection is mostly 
    exogenous. 
Aspergillus transmission occurs by inhalation 
    of airborne conidia. 
Risk Factors for invasive aspergillosis are: 
    Glucocoicoid use (the most impoant risk 
    factor) Profound neutropenia or Neutrophil 
    dysfunction Underlying pneumonia or COPD, 
    tuberculosis or sarcoidosis Antitumor 
    necrosis factor therapy.
#### C.

Here is your question. Please respond to this 
    question based on the context and by 
    adhering to the given format: provide 
    step-by-step reasoning (one sentence per 
    line), then give the final answer (A/B/C/D)
    after '####'.
\end{Verbatim}

\subsubsection{MedQA}
The prompting format for MedQA dataset is listed as follows:
\vspace{1ex}
\begin{Verbatim}
Use the step-by-step method as shown in the 
    example to answer the question. You should 
    give the reasoning steps and final answer 
    based on the provided context.

Example:
Q: A 21-year-old sexually active male complains 
    of fever, pain during urination, and 
    inflammation and pain in the right knee. A 
    culture of the joint fluid shows a bacteria 
    that does not ferment maltose and has no 
    polysaccharide capsule. The physician 
    orders antibiotic therapy for the patient. 
    The mechanism of action of action of the 
    medication given blocks cell wall synthesis, 
    which of the following was given? (A) 
    Gentamicin, (B) Ciprofloxacin, (C) 
    Ceftriaxone, (D) Trimethoprim.
A: The symptoms and culture results suggest a 
    bacterial infection that affects both the 
    urinary tract and joints, indicating a 
    systemic infection.
Bacteria that do not ferment maltose and lack a 
    polysaccharide capsule could indicate a 
    variety of bacteria, but the treatment 
    approach focuses on the mechanism of action 
    of the antibiotic rather than the specific 
    bacteria.
Antibiotics that block cell wall synthesis are 
    typically beta-lactams, which include 
    penicillins and cephalosporins.
Gentamicin is an aminoglycoside antibiotic, 
    which works by inhibiting protein 
    synthesis.
Ciprofloxacin is a fluoroquinolone, which works 
    by inhibiting bacterial DNA gyrase and 
    topoisomerase IV, affecting DNA replication.
Ceftriaxone is a third-generation cephalosporin, 
    which works by inhibiting cell wall 
    synthesis.
Trimethoprim is an antibiotic that inhibits 
    bacterial dihydrofolate reductase, 
    affecting folic acid synthesis.
#### C.

Here is your question. Please respond to this 
    question based on the context and by 
    adhering to the given format: provide 
    step-by-step reasoning (one sentence per 
    line), then give the final answer (A/B/C/D)
    after '####'.
\end{Verbatim}

\subsubsection{MMLU-Med}
The prompting format for MMLU-Med dataset is listed as follows:
\vspace{1ex}
\begin{Verbatim}
Use the step-by-step method as shown in the 
    example to answer the question. You should 
    give the reasoning steps and final answer 
    based on the provided context.

Example:
Q: What size of cannula would you use in a 
    patient who needed a rapid blood 
    transfusion (as of 2020 medical knowledge)?
    (A) 18 gauge, (B) 20 gauge, (C) 22 gauge, 
    (D) 24 gauge. 
A: The gauge of a cannula indicates its 
    diameter: the smaller the number, the 
    larger the diameter of the cannula.
A larger diameter cannula allows for the rapid 
    administration of fluids, including blood.
In emergency situations requiring rapid 
    transfusion, a larger cannula is preferred
    to ensure quick delivery of blood to the 
    patient.
An 18 gauge cannula is larger than the 20, 22,
    and 24 gauge options and is commonly used 
    for rapid transfusions.
#### A.

Here is your question. Please respond to this 
    question based on the context and by 
    adhering to the given format: provide 
    step-by-step reasoning (one sentence per 
    line), then give the final answer (A/B/C/D)
    after '####'.
\end{Verbatim}

\subsubsection{PubMedQA}
The prompting format for PubMedQA dataset is listed as follows:
\vspace{1ex}
\begin{Verbatim}
Use the step-by-step method as shown in the 
    example to answer the question. You should 
    give the reasoning steps and final answer
    based on the provided context.

Example:
Q: Do familiar teammates request and accept 
    more backup?
A: Transactive memory theory extends to 
    high-stress environments in which members' 
    expertise is highly overlapping.
Teammates' shared mental models about one 
    another increase the likelihood that they 
    will request and accept backup.
#### Yes.

Here is your question. Please respond to this 
    question based on the context and by 
    adhering to the given format: provide 
    step-by-step reasoning (one sentence per 
    line), then give the final answer 
    (Yes/No/Maybe) after '####'.
\end{Verbatim}

\subsubsection{BioASQ}
The prompting format for BioASQ dataset is listed as follows:
\vspace{1ex}
\begin{Verbatim}
Use the step-by-step method as shown in the 
    example to answer the question. You should 
    give the reasoning steps and final answer 
    based on the provided context.

Example:
Q: Can losartan reduce brain atrophy in 
    Alzheimer's disease?
A: Losartan is primarily used for hypertension 
    and may indirectly affect factors 
    associated with Alzheimer's disease 
    progression. 
Despite potential neuroprotective effects, such 
    as reducing inflammation and oxidative 
    stress, there is limited direct evidence 
    linking losartan to reduced brain atrophy 
    in Alzheimer's disease. 
Clinical trials specifically targeting this 
    outcome are necessary to establish a 
    definitive effect.
#### no

Here is your question. Please respond to this 
    question based on the context and by 
    adhering to the given format: provide 
    step-by-step reasoning (one sentence per 
    line), then give the final answer (yes/no)
    after '####'.
\end{Verbatim}

\subsubsection{MedNLI}
The prompting format for MedNLI dataset is listed as follows:
\vspace{1ex}
\begin{Verbatim}
Use the step-by-step method as shown in the 
    example to deduce the relationship between 
    the given two sentences. You should give 
    the reasoning steps and final answer based 
    on the provided context.

Example:
Sentence A: Labs were notable for Cr 1.7 
    (baseline 0.5 per old records) and lactate 
    2.4.
Sentence B: Patient has elevated Cr
Answer: Sentence A states that the patient's 
    Cr (creatinine) level is 1.7, which is 
    higher than the baseline of 0.5 according 
    to old records. 
Sentence B simply states that the patient has 
    elevated Cr. 
The information in Sentence A supports the 
    claim in Sentence B, making the relationship 
    entailment.
#### entailment

Here are the given two sentences. What is the 
    relationship between the given two sentences? 
    Please answer from [entailment, neutral, 
    contradiction]. Please give the answer after 
    '####'.
\end{Verbatim}

\subsubsection{MediQA-RQE}
The prompting format for MediQA-RQE dataset is listed as follows:
\vspace{1ex}
\begin{Verbatim}
Does the provided solution correctly answer the
    question? Please answer from [true, false]. 
    Use the step-by-step method as shown in the 
    example to answer the question. You should 
    give the reasoning steps and final answer 
    based on the provided context.

Example:
Question: What is High Blood Pressure?
Solution: High Blood Pressure. I know you may 
    not answer this but my blood pressure comes 
    up at night when I am asleep. I take four 
    medicines. I have asked doctors why this 
    happens and no one knows. This morning at 
    four A.M. It was 164 and I took a clonidine 
    to help get it done. It worries me so.
Judge: The provided solution does not correctly 
    answer the question "What is High Blood 
    Pressure?" 
The solution discusses a personal experience 
    with high blood pressure and medication but 
    does not define or explain what high blood 
    pressure is. 
A correct answer would define high blood 
    pressure  as a condition in which the force 
    of the blood against the artery walls is 
    too high, typically considered to be 140/90 
    mmHg or higher.
#### false

Here is the question and answer. Please then 
    give the final judge (true/false) after 
    '####'.
\end{Verbatim}

\subsubsection{PubHealth}
The prompting format for PubHealth dataset is listed as follows:
\vspace{1ex}
\begin{Verbatim}
Use the step-by-step method as shown in the 
    example to answer the question. You 
    should give the thought steps and final 
    answer based on the provided context. 
    Please judge whether the claim is true 
    or false. 

Example:
Claim: Annual Mammograms May Have More 
    False-Positives	October 18, 2011
Judge: This article reports on the results 
    of a study of nearly 170,000 women who 
    had screening mammograms beginning 
    between age 40-59. 
The study found that over ten years of 
    screening mammograms, over half of the 
    women will experience a false-positive 
    recall for additional mammography. 
In addition, 7%-9% of the women will have a 
    biopsy for a suspicious lump which is 
    not cancerous. 
Both of those percentages decrease if the
    woman is screened every other year 
    rather than every year. 
Even with biennial mammography, 41% of 
    women will experience a recall over 10 
    years of mammography. 
The study’s Principal Investigator 
    emphasized that “in most cases, a 
    recall doesn’t mean you have cancer.”  
She hoped this knowledge would reduce the 
    anxiety of women who are recalled. 
The story never explained the size of the 
    decrease in the number of false positives 
    between annual (61.3%) and biennial 
    screening (41.6%). 
Our first two reviewers were a researcher 
    who specializes in health decisions and 
    a breast cancer survivor trained in 
    evidence by the Natiional Breast Cancer 
    Coalition’s Project LEAD. 
This study is valuable because it helps to 
    quantify and compare the harms of annual 
    and biennial screening, specifically 
    the number of false positives and the 
    number of unnecessary biopsies. 
Prior to this study, estimates of false
    positive screening mammography rates 
    varied widely. 
The critical question is whether you can do 
    less frequent screening, subject women 
    to fewer harms and get similar results 
    in terms of detection of “early stage” 
    cancer. 
This study’s data seems to suggest that 
    answer is yes.
#### mixture

Here is the claim. Please then give the final 
    judge (true/false/mixture/unproven) after 
    '####'.
\end{Verbatim}


\section{HIPAA Compliance with API Service}
Black-box LLMs have set new standards for SOTA performance on biomedical NLP tasks with their inherent capabilities~\citep{nori2023capabilities}. Despite these advancements, there remains potential for improvement in domain-specific applications through domain specialization~\citep{chen2023large}.
However, the OpenAI fine-tuning API is not compliant with HIPAA regulations and cannot be used directly for clinical data that contains patient information. While the Microsoft Azure OpenAI fine-tuning API service is HIPAA-compliant, it still poses significant risks when it comes to data sharing through external APIs~\citep{shi2023detecting} and entails substantial costs for model fine-tuning and deployment.
\method offers an alternative approach for adapting black-box LLMs without the use of APIs, thereby greatly enhancing data privacy during training and substantially reducing associated API costs.

\section{Parameter Studies of Azure-SFT}
\label{app:parameter}
\begin{figure*}[ht]
	\centering
	\subfigure[MedMCQA (LRM=0.1, Epoch=3)]{
		\includegraphics[width=0.31\linewidth]{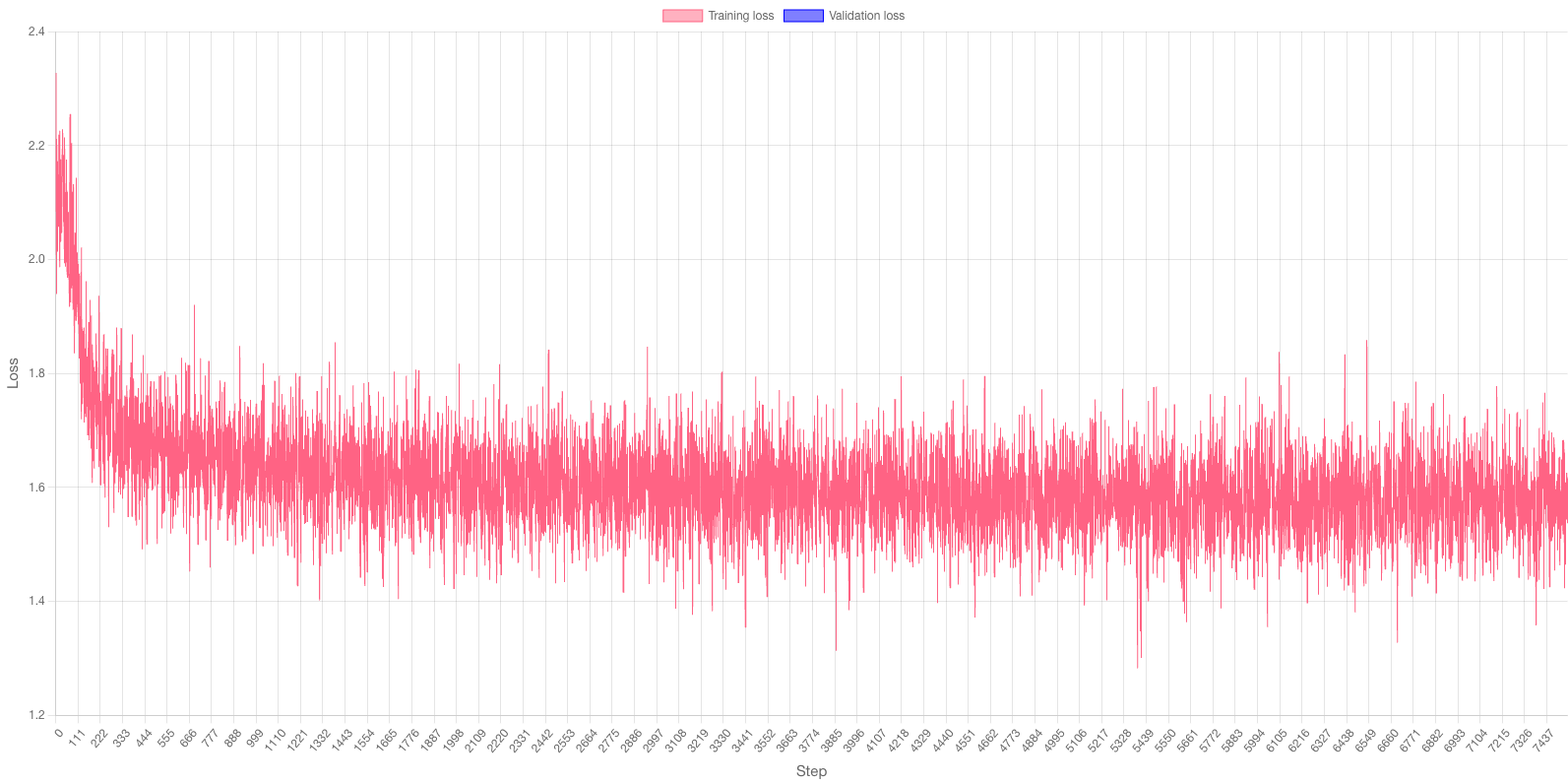}
		\label{fig:medmcqa-1}
	} 
     \subfigure[MedMCQA (LRM=1, Epoch=3)]{
		\includegraphics[width=0.31\linewidth]{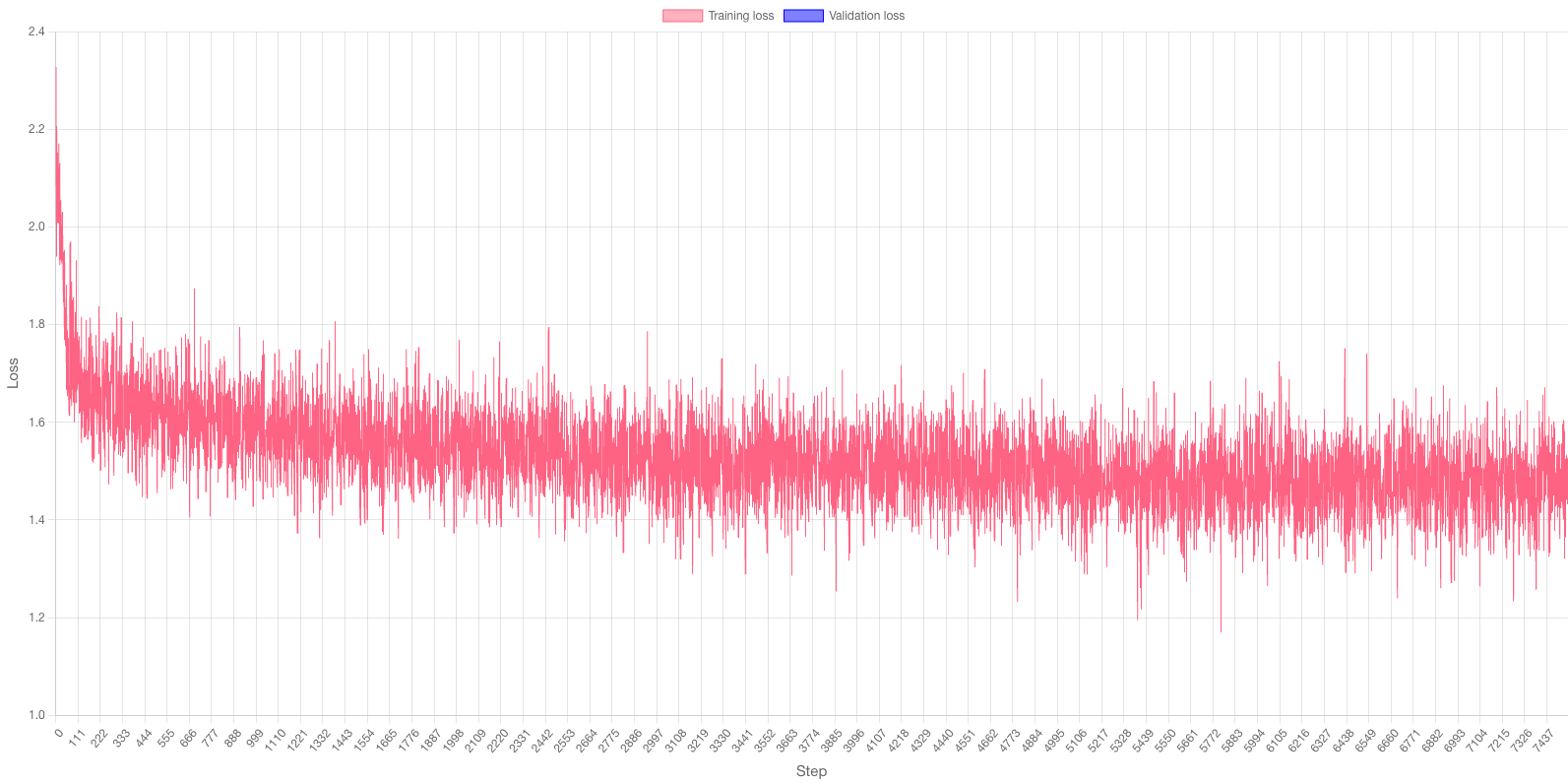}
		\label{fig:medmcqa-2}
	} 
     \subfigure[MedMCQA (LRM=0.1, Epoch=5)]{
		\includegraphics[width=0.31\linewidth]{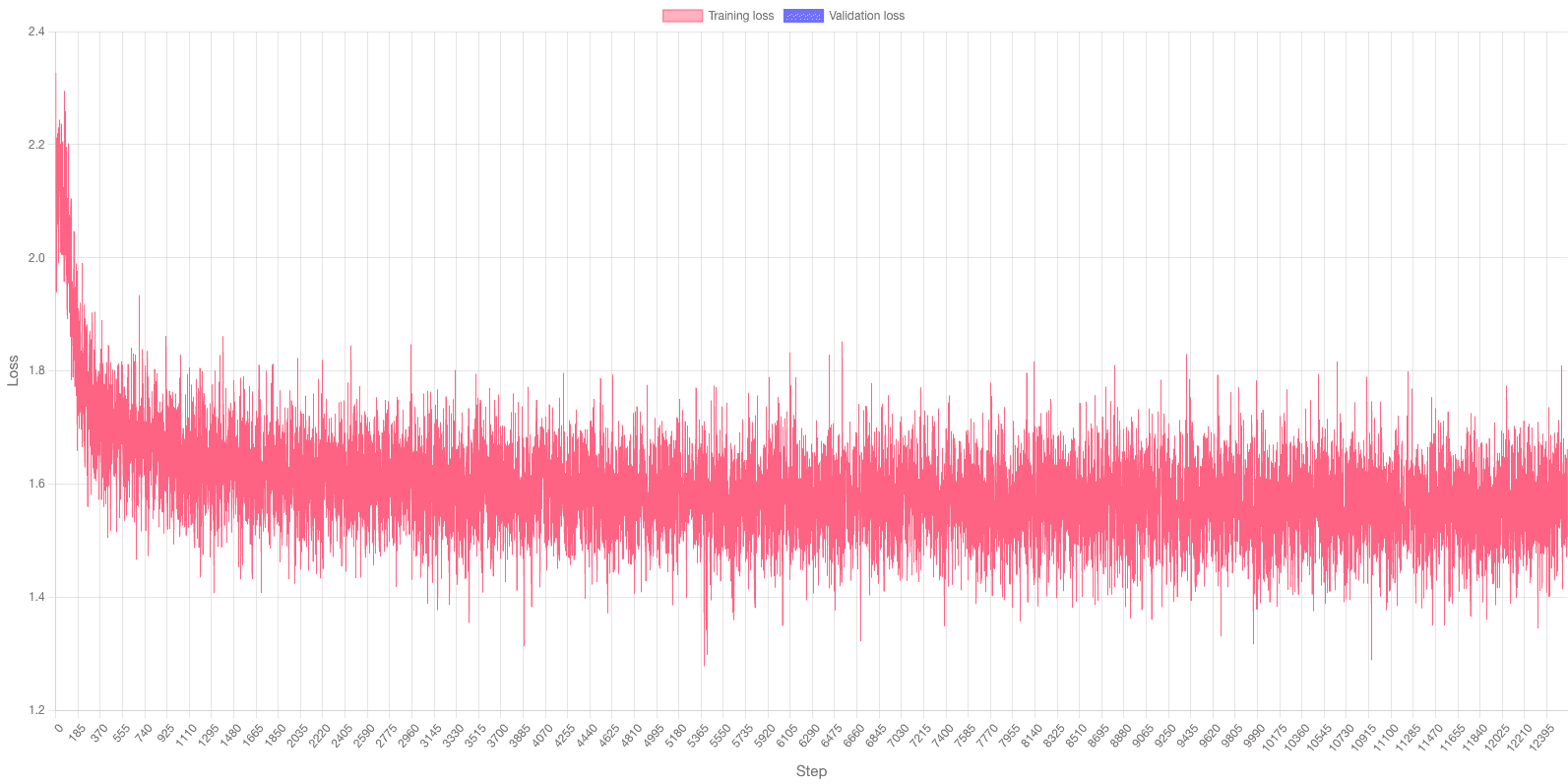}
		\label{fig:medmcqa-3}
	}

    \subfigure[MedQA (LRM=0.1, Epoch=3)]{
		\includegraphics[width=0.31\linewidth]{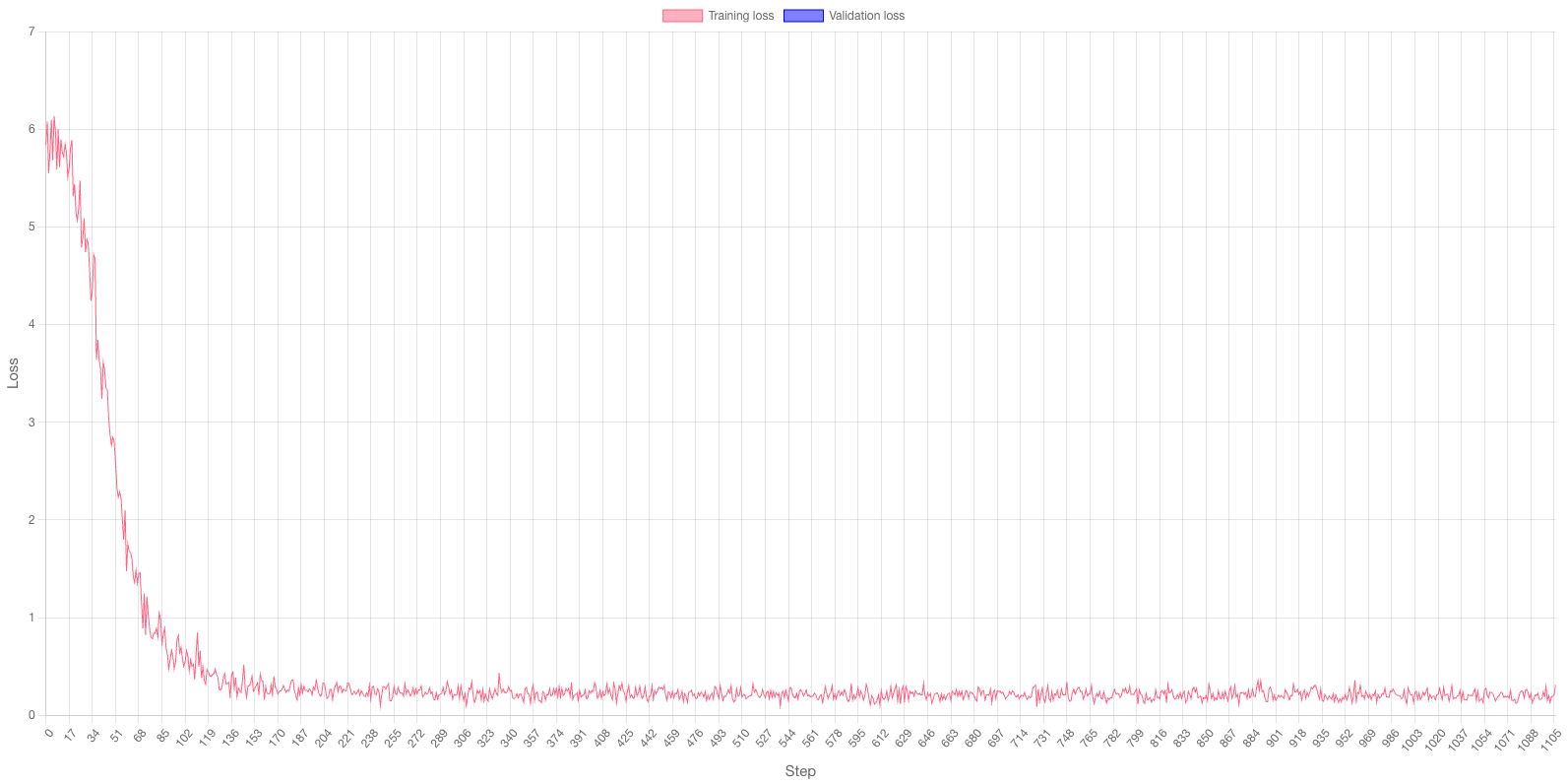}
		\label{fig:medqa-1}
	} 
     \subfigure[MedQA (LRM=1, Epoch=3)]{
		\includegraphics[width=0.31\linewidth]{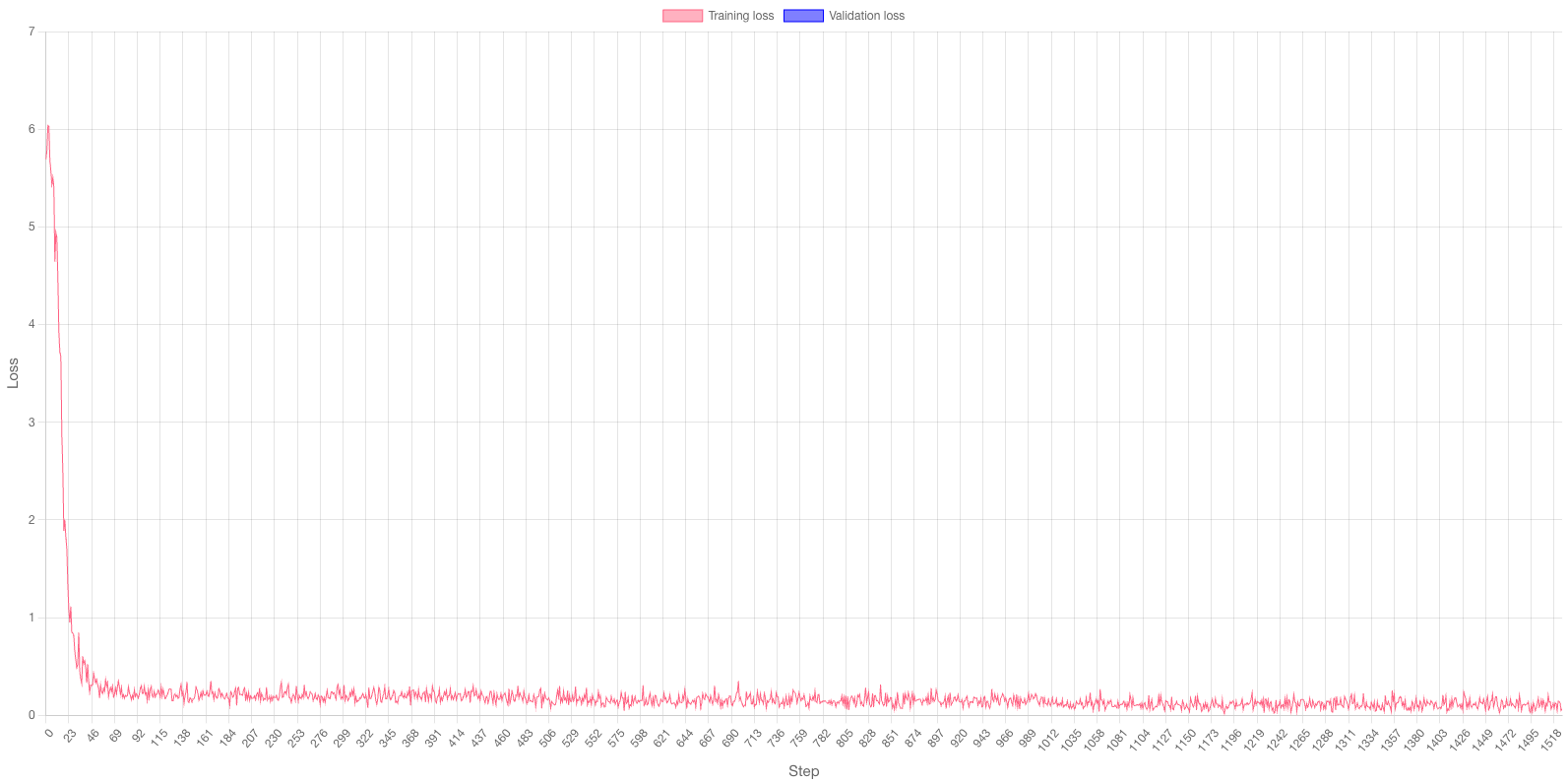}
		\label{fig:medqa-2}
	} 
     \subfigure[MedQA (LRM=0.1, Epoch=5)]{
		\includegraphics[width=0.31\linewidth]{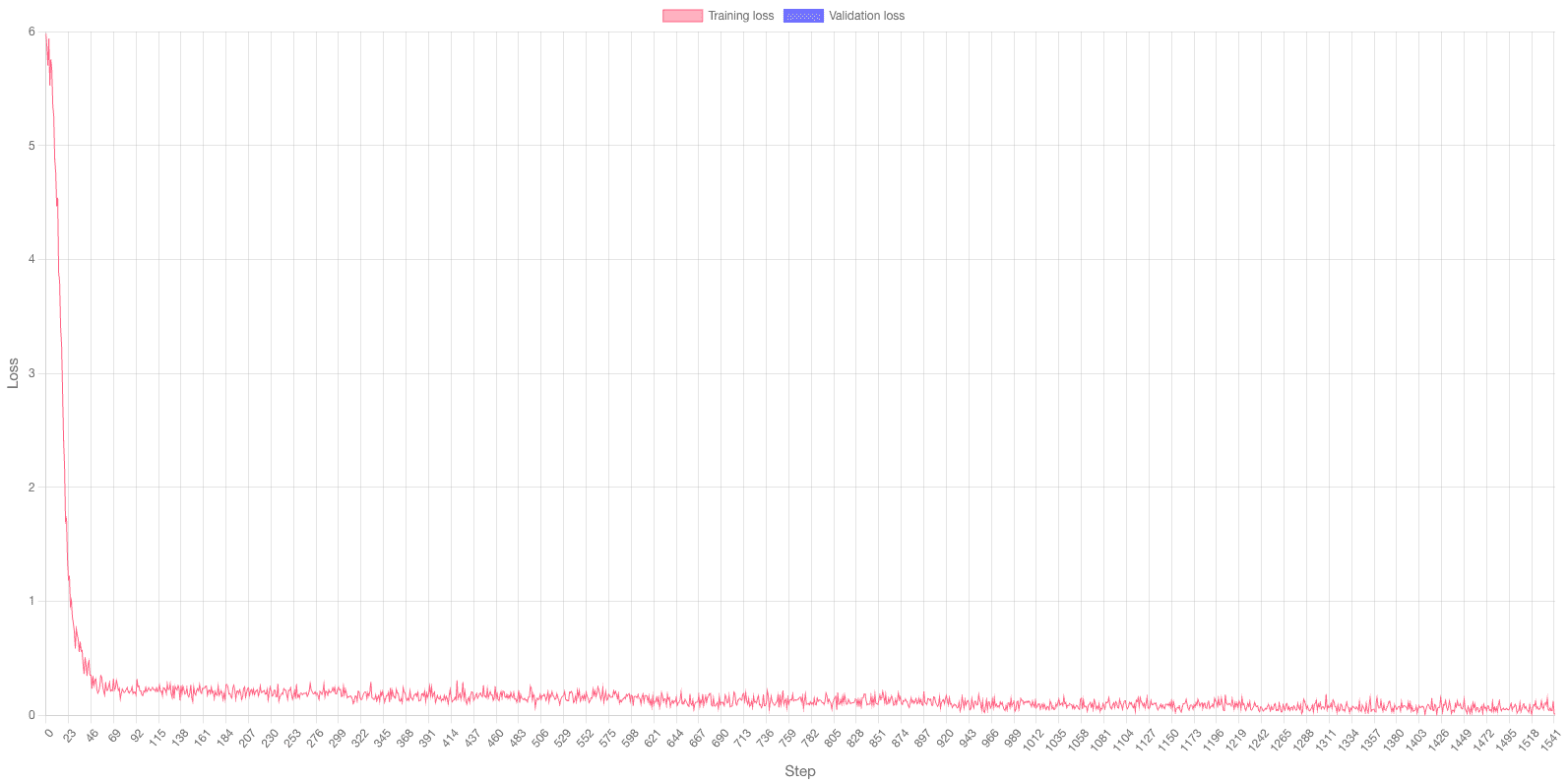}
		\label{fig:medqa-3}
	}

 \subfigure[MMLU (LRM=0.1, Epoch=3)]{
		\includegraphics[width=0.31\linewidth]{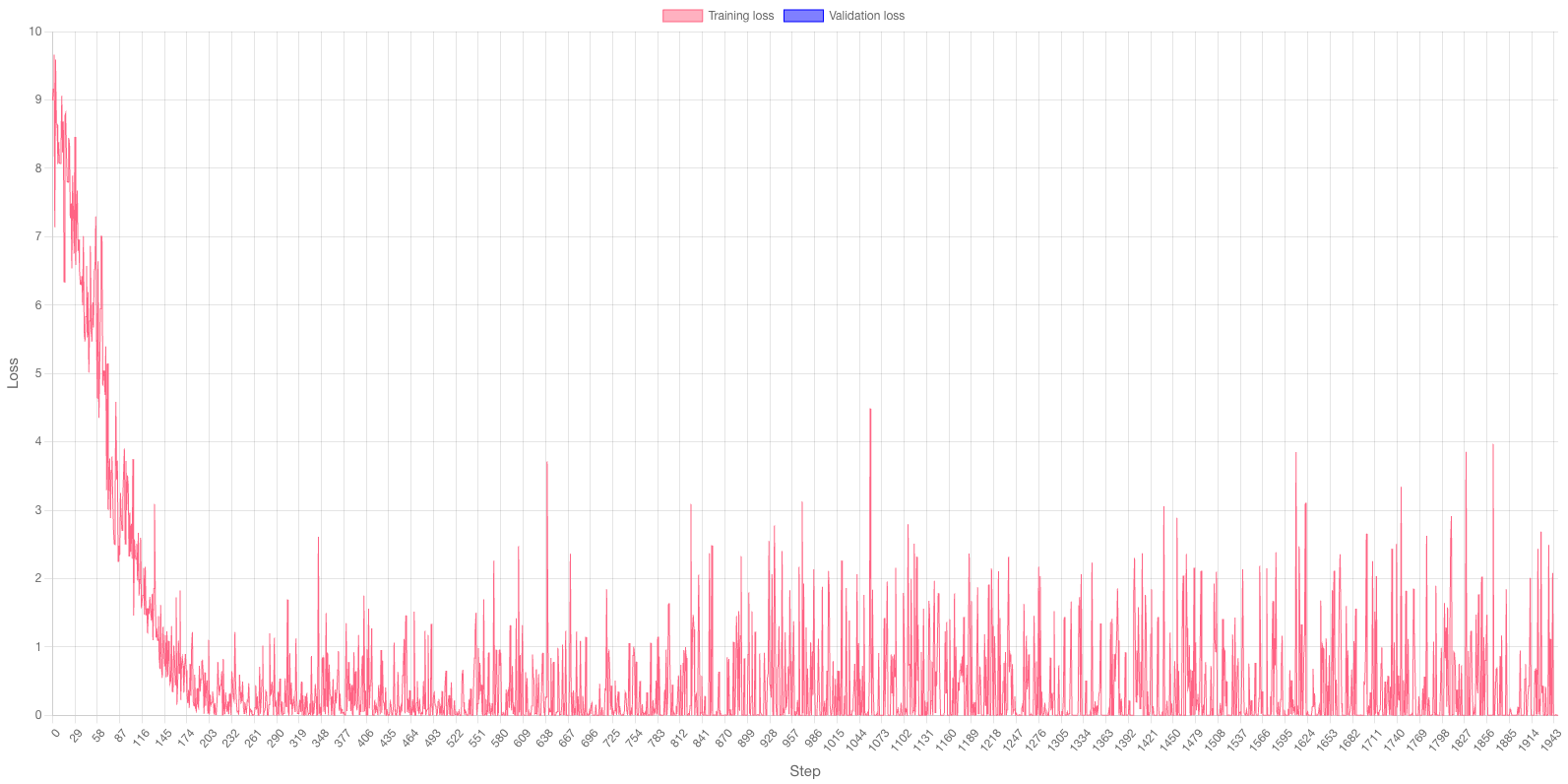}
		\label{fig:mmlu-1}
	} 
     \subfigure[MMLU (LRM=1, Epoch=3)]{
		\includegraphics[width=0.31\linewidth]{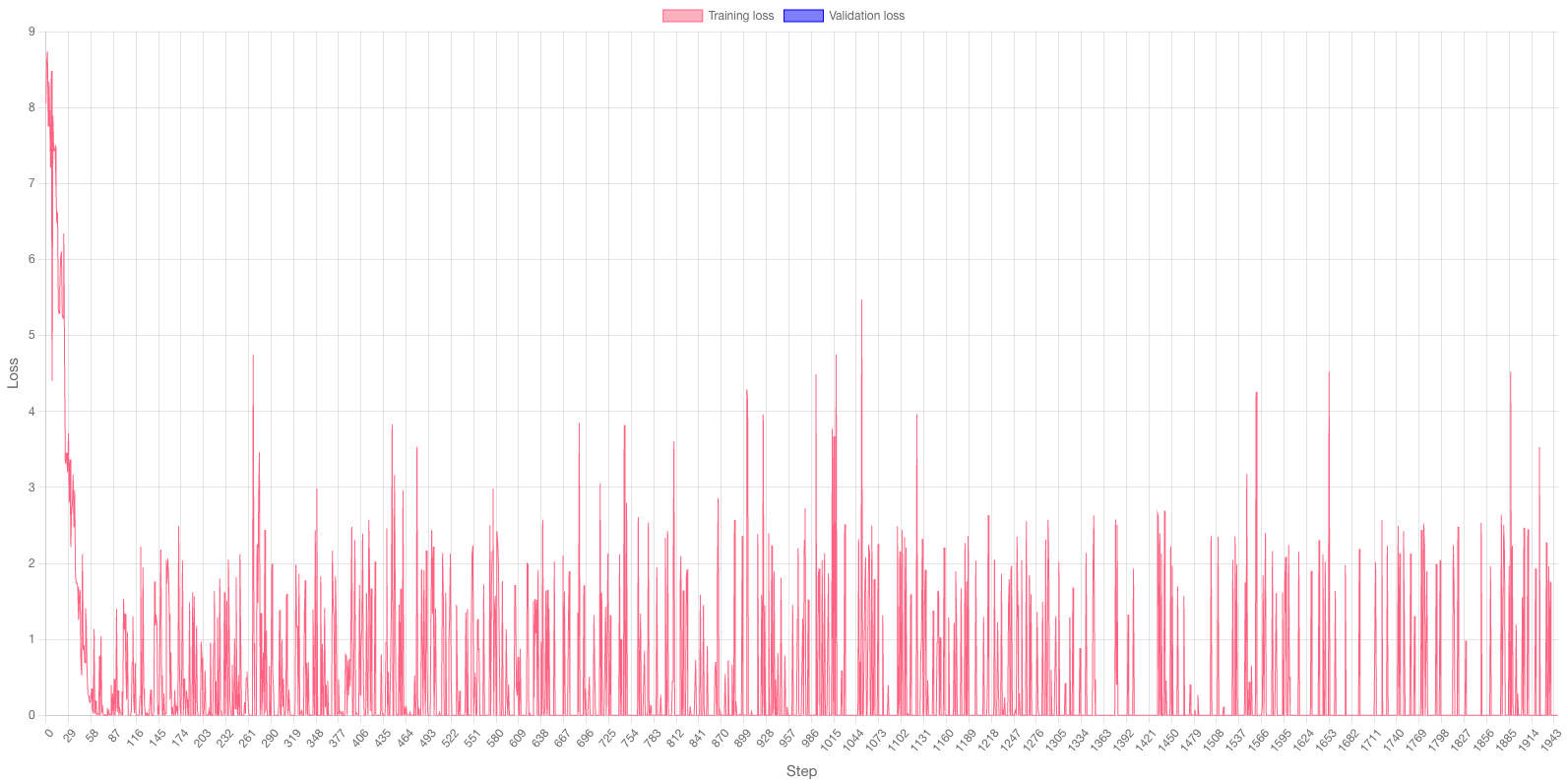}
		\label{fig:mmlu-2}
	} 
     \subfigure[MMLU (LRM=0.1, Epoch=5)]{
		\includegraphics[width=0.31\linewidth]{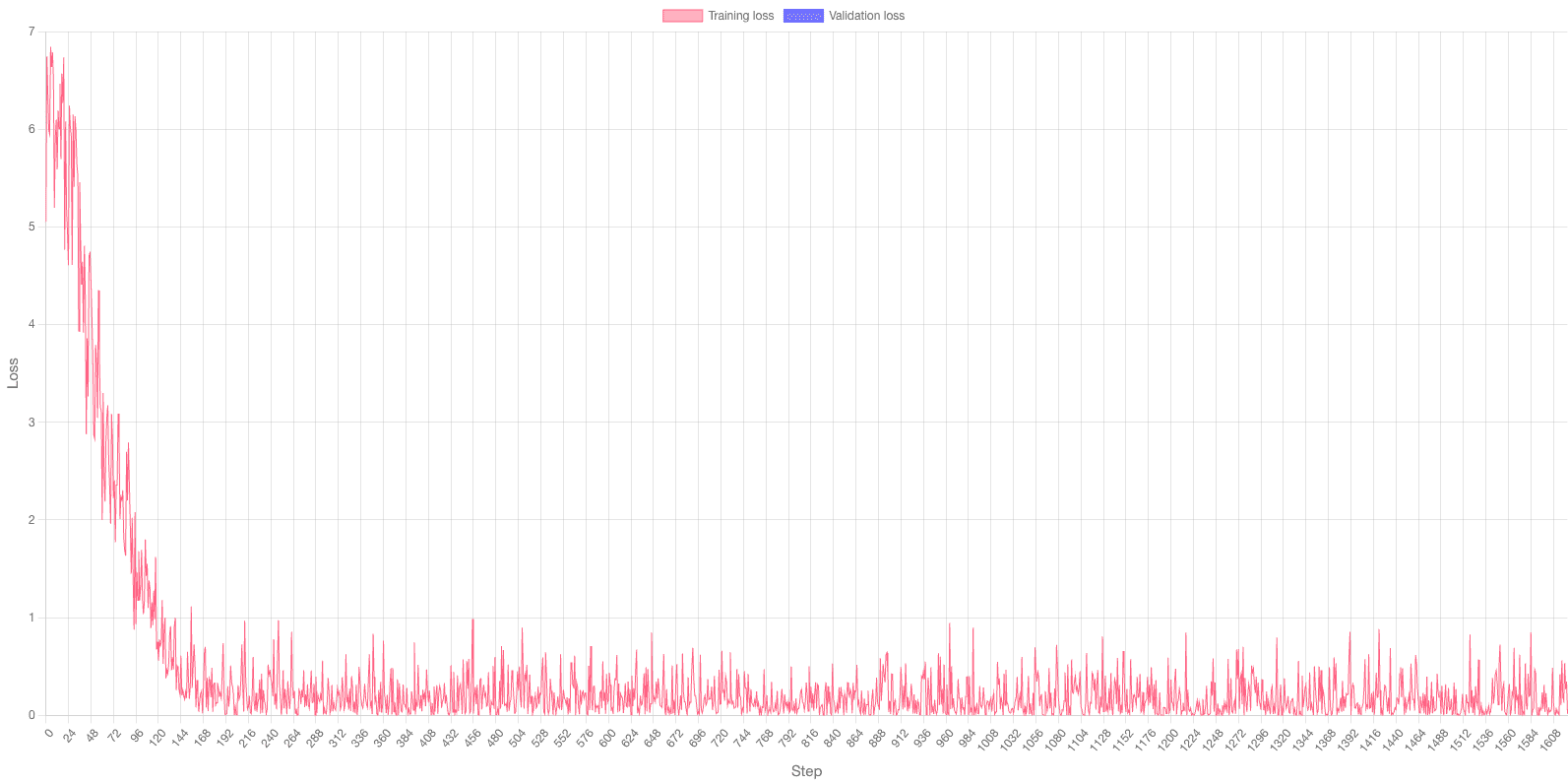}
		\label{fig:mmlu-3}
	}

 \subfigure[PubMedQA (LRM=0.1, Epoch=3)]{
		\includegraphics[width=0.31\linewidth]{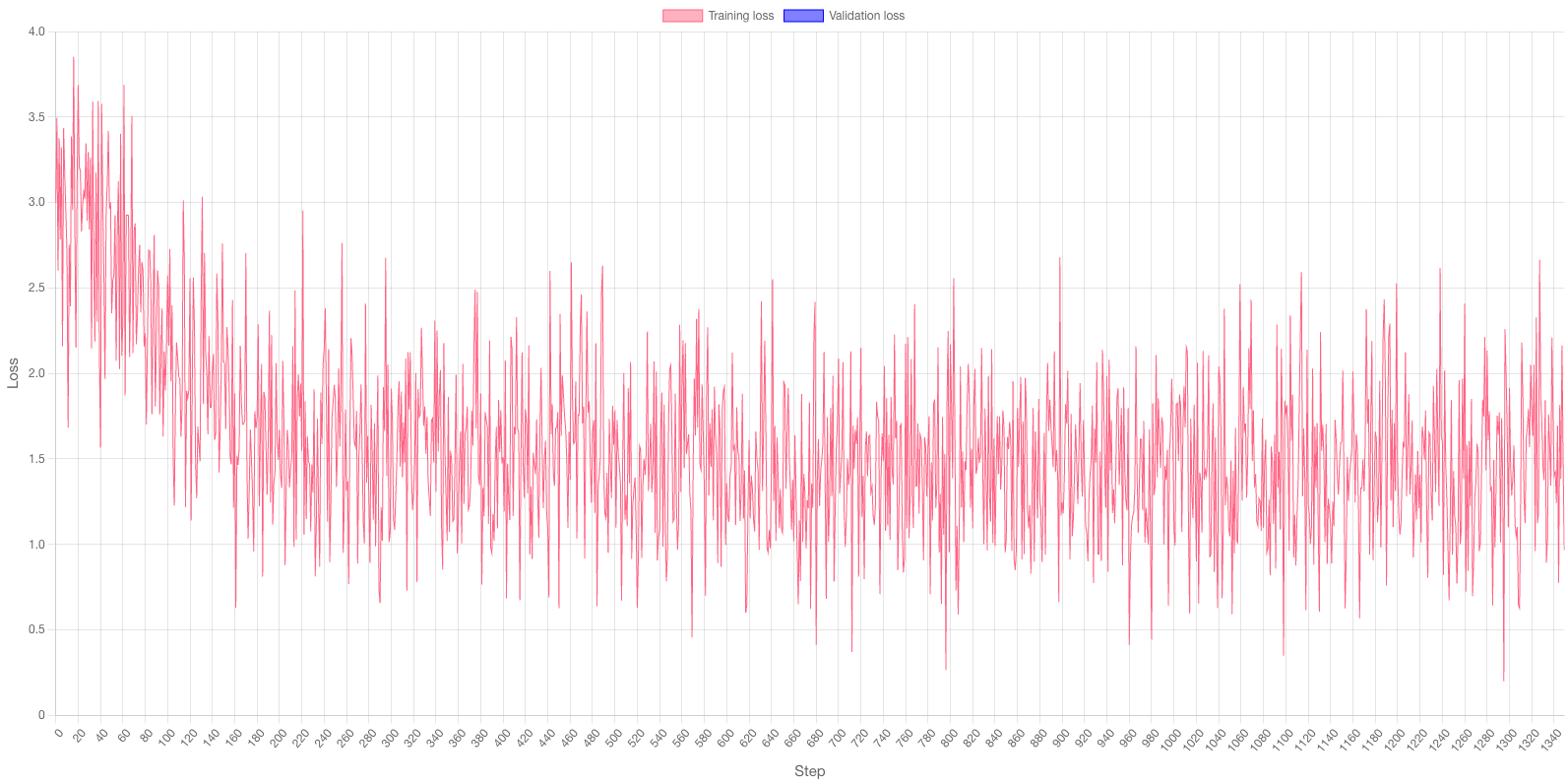}
		\label{fig:pubmedqa-1}
	} 
     \subfigure[PubMedQA (LRM=1, Epoch=3)]{
		\includegraphics[width=0.31\linewidth]{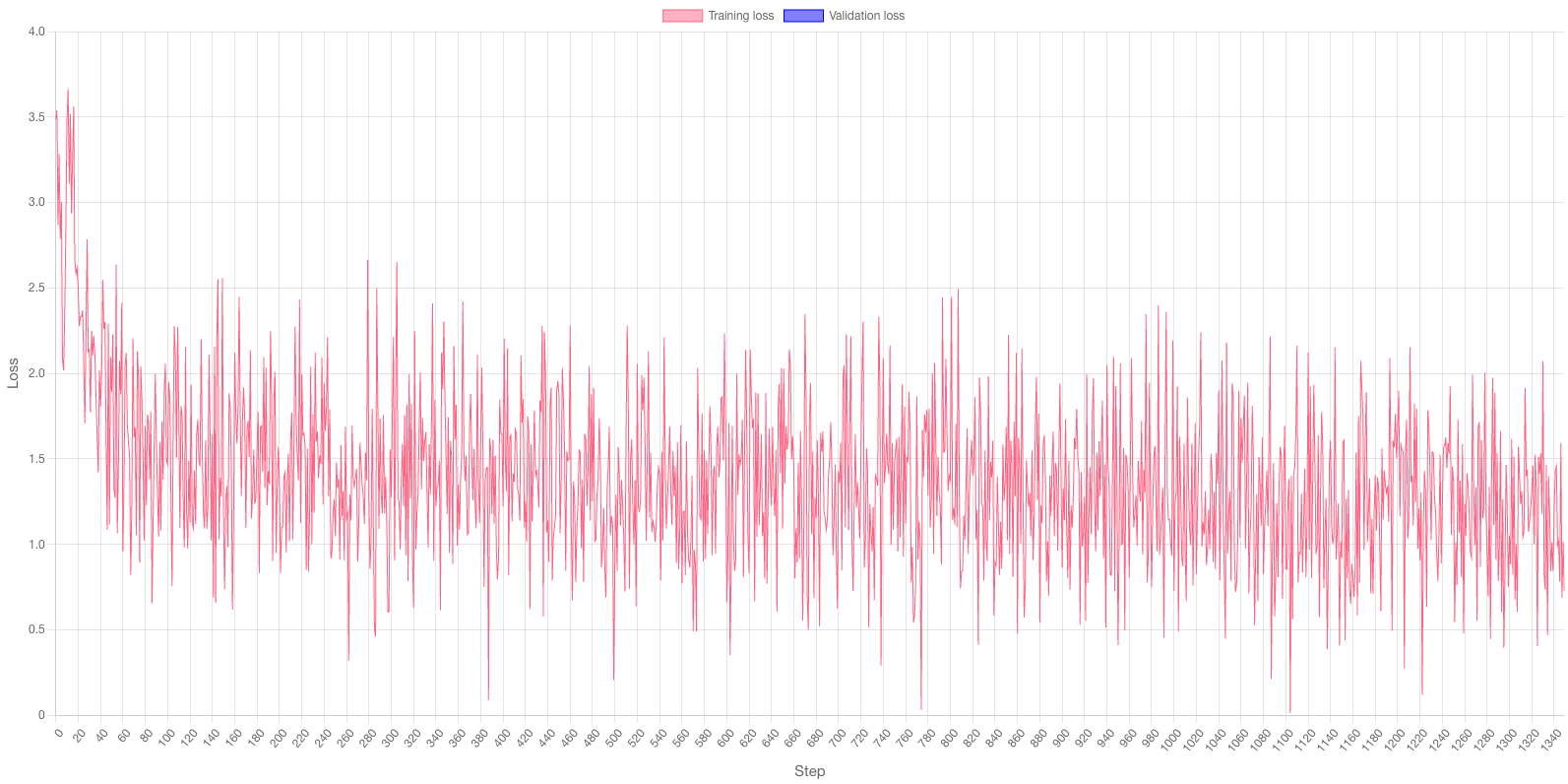}
		\label{fig:pubmedqa-2}
	} 
     \subfigure[PubMedQA (LRM=0.1, Epoch=5)]{
		\includegraphics[width=0.31\linewidth]{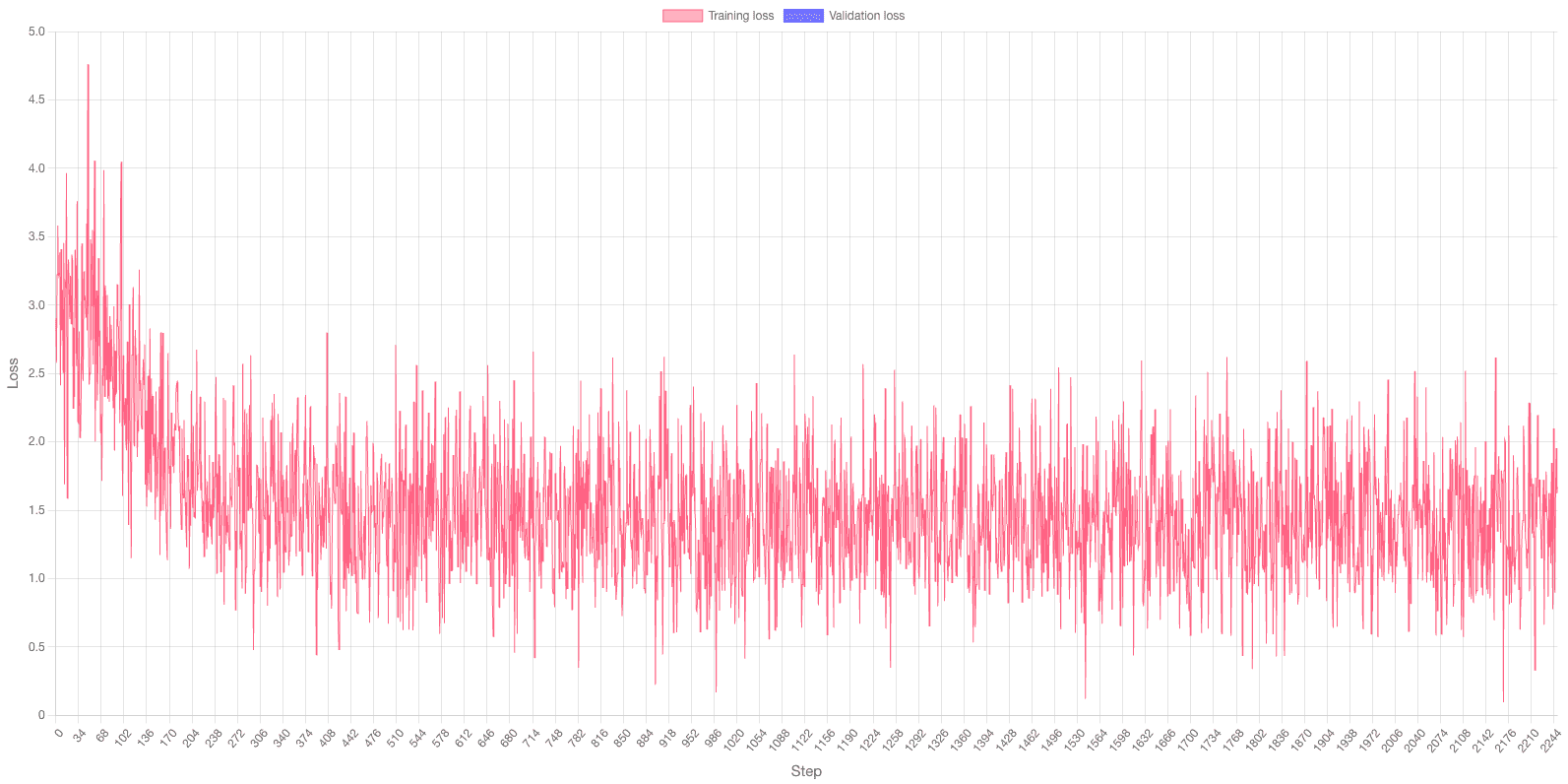}
		\label{fig:pubmedqa-3}
	}

 \subfigure[BioASQ (LRM=0.1, Epoch=3)]{
		\includegraphics[width=0.31\linewidth]{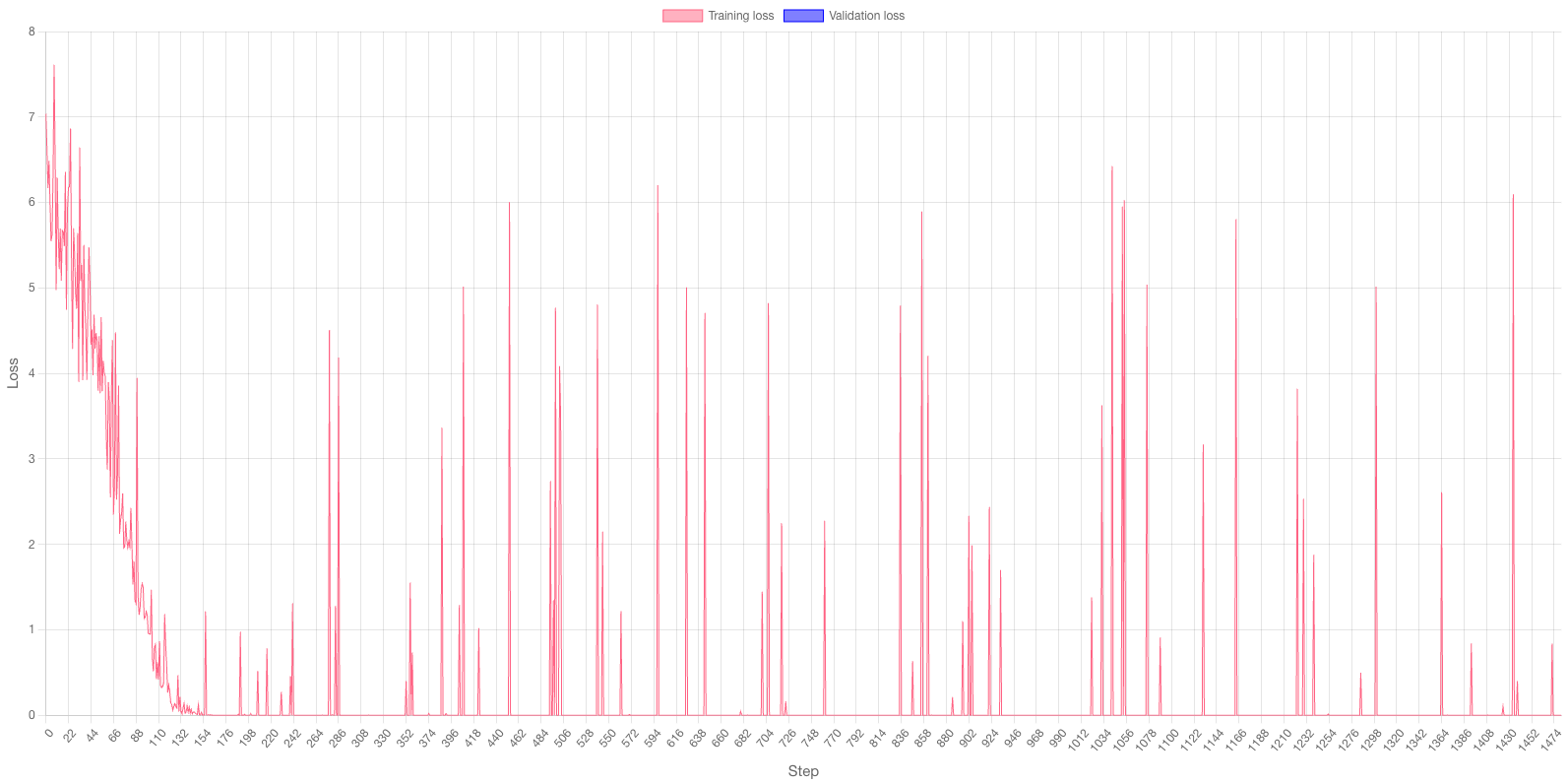}
		\label{fig:bioasq-1}
	} 
     \subfigure[BioASQ (LRM=1, Epoch=3)]{
		\includegraphics[width=0.31\linewidth]{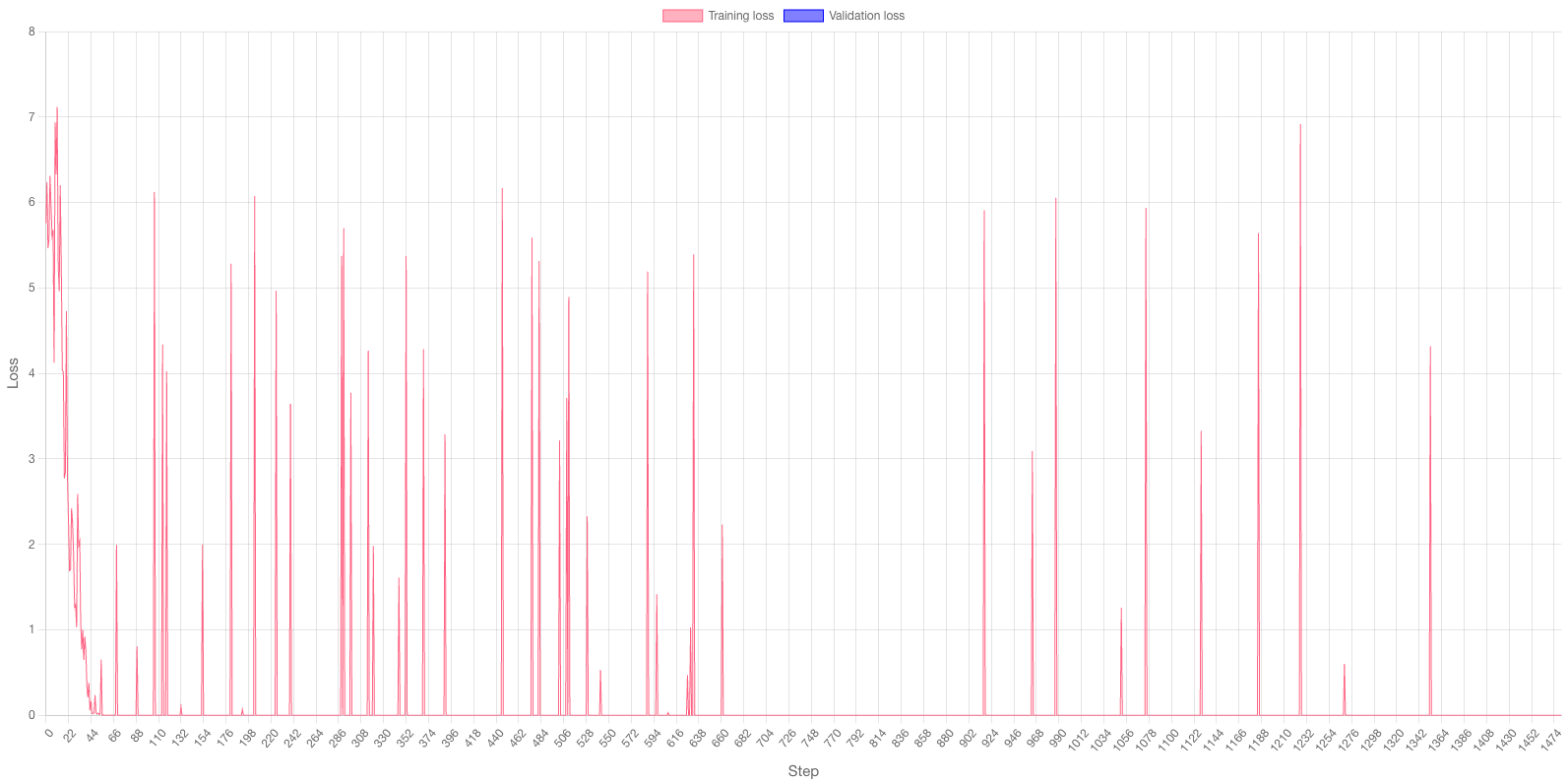}
		\label{fig:bioasq-2}
	} 
     \subfigure[BioASQ (LRM=0.1, Epoch=5)]{
		\includegraphics[width=0.31\linewidth]{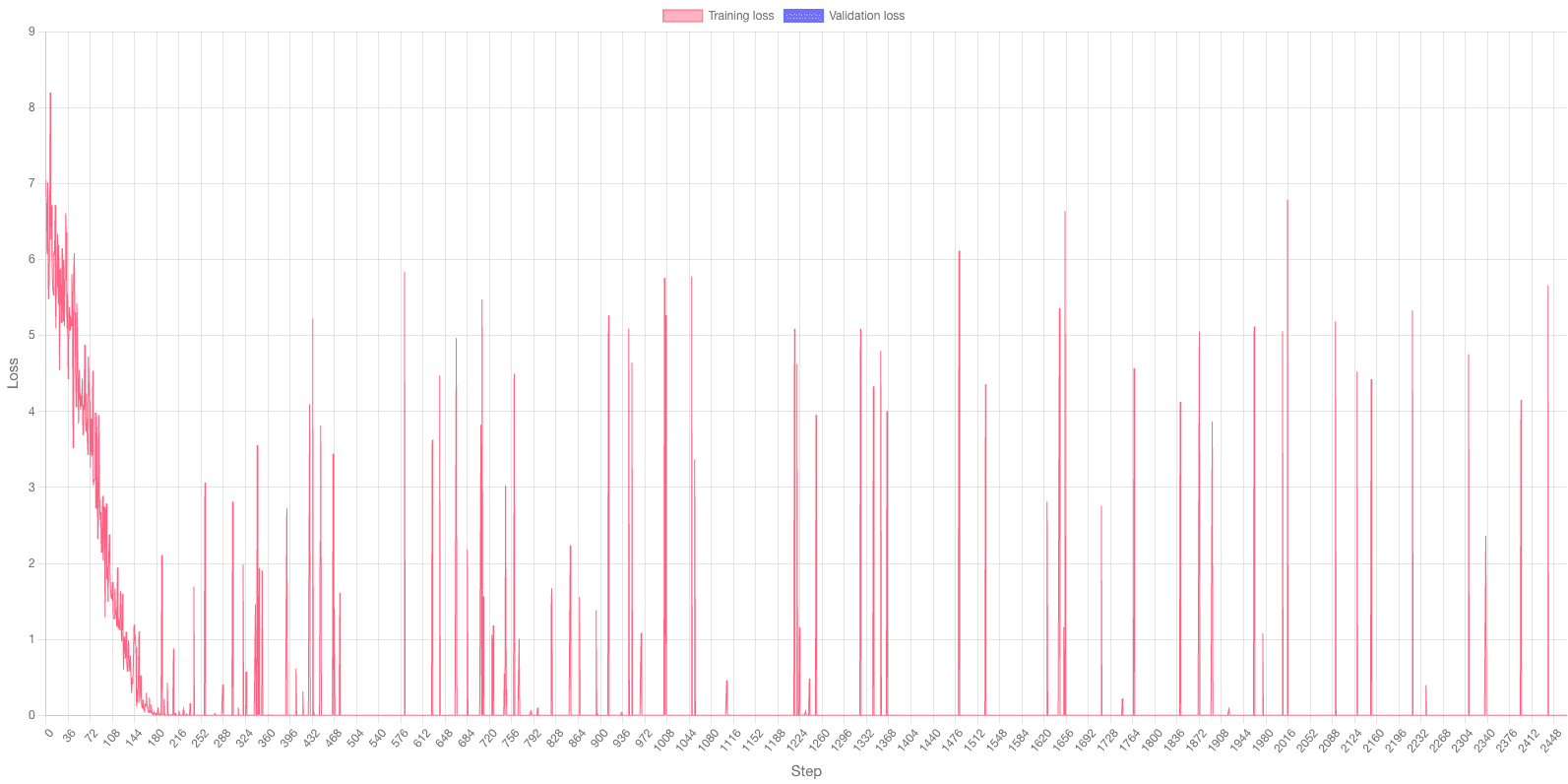}
		\label{fig:bioasq-3}
	}
	\caption{Loss function curve of fine-tuning GPT-3.5-Turbo for biomedical QA tasks through Microsoft Azure fine-tuning API service.  }
\label{fig:azure-loss}
\end{figure*}

\begin{figure*}[ht]
	\centering
	\subfigure[MedNLI (LRM=0.1, Epoch=3)]{
		\includegraphics[width=0.31\linewidth]{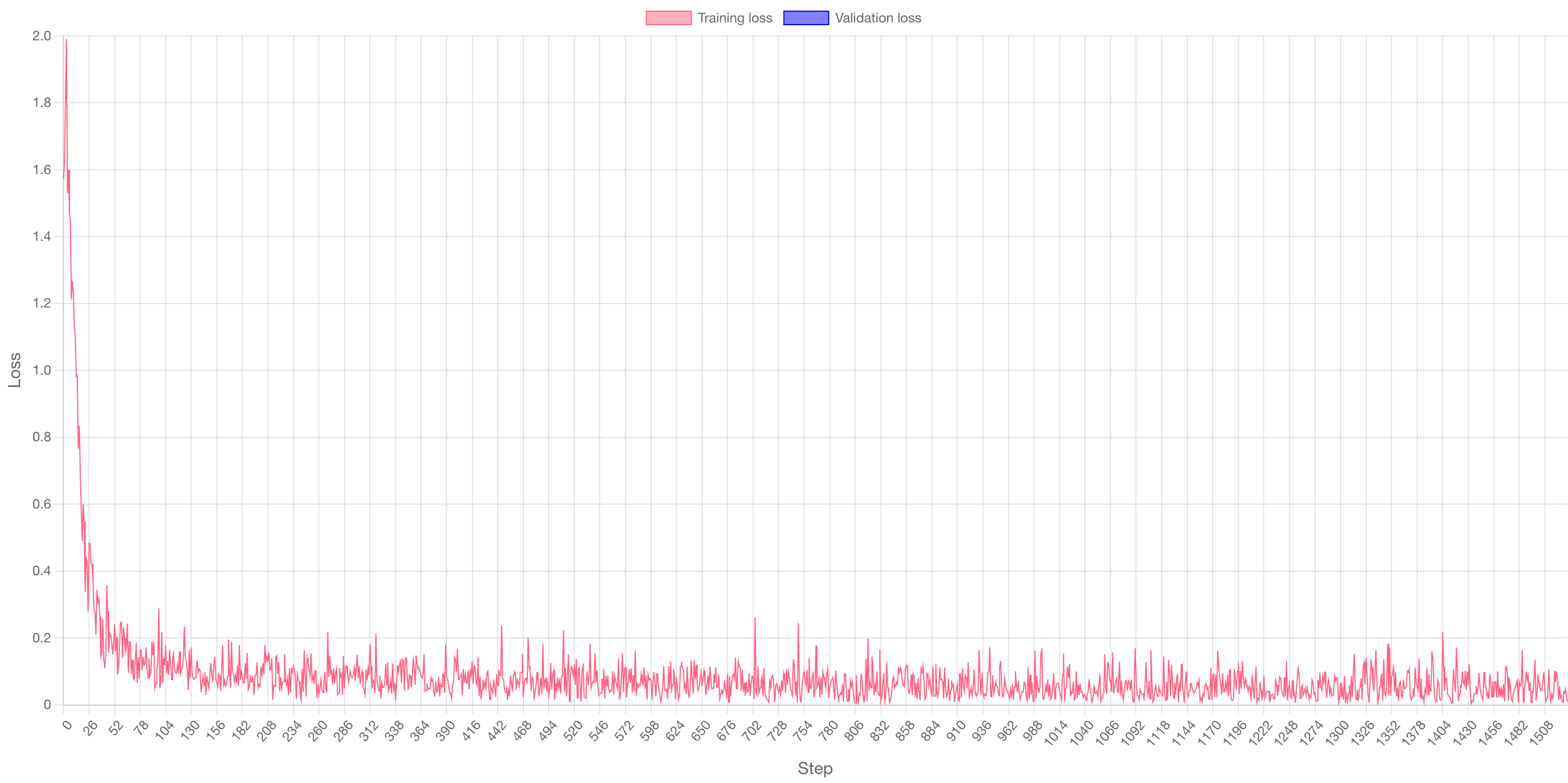}
		\label{fig:MedNLI-1}
	} 
     \subfigure[MedNLI (LRM=1, Epoch=3)]{
		\includegraphics[width=0.31\linewidth]{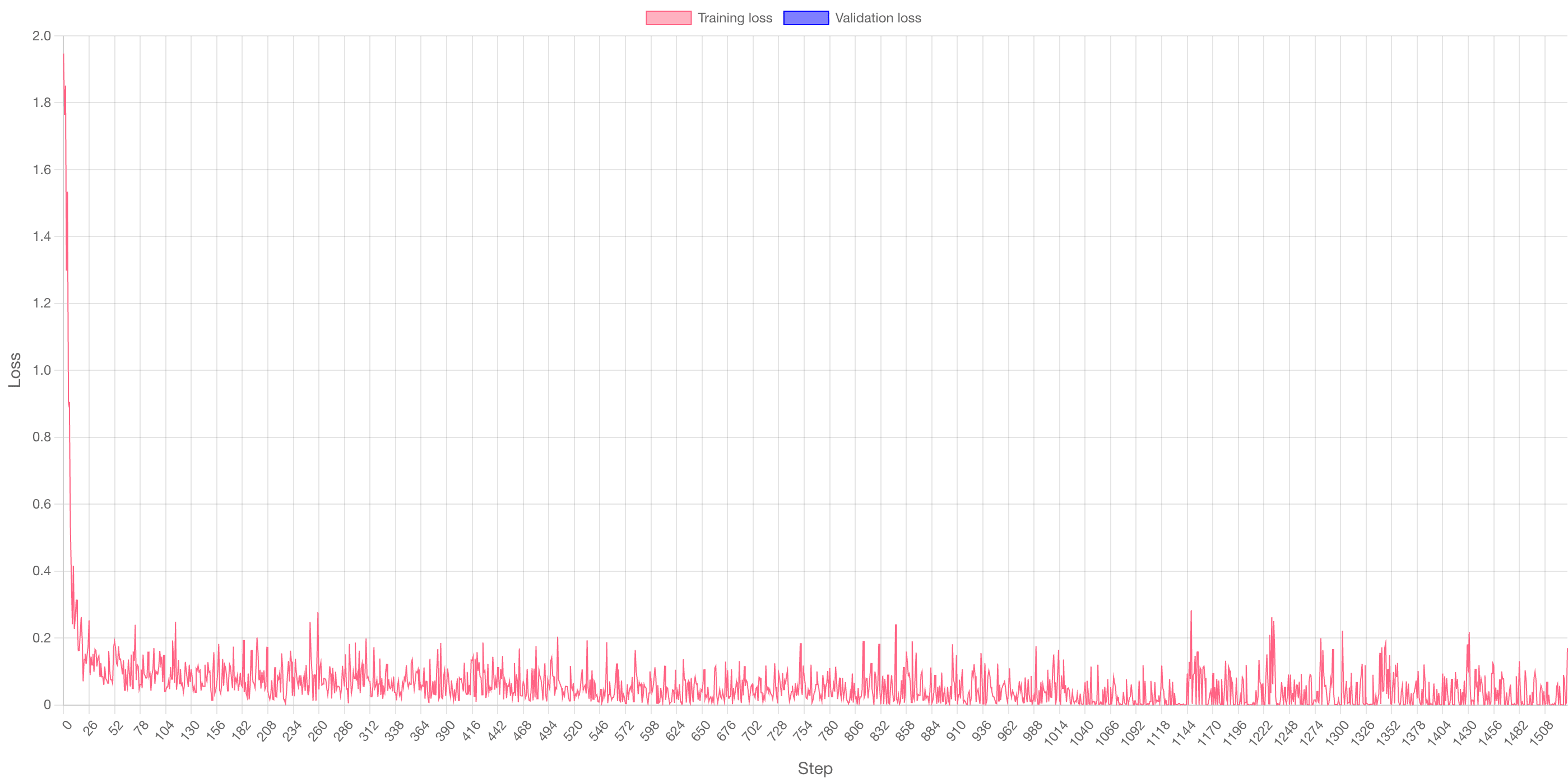}
		\label{fig:MedNLI-2}
	} 
     \subfigure[MedNLI (LRM=0.1, Epoch=5)]{
		\includegraphics[width=0.31\linewidth]{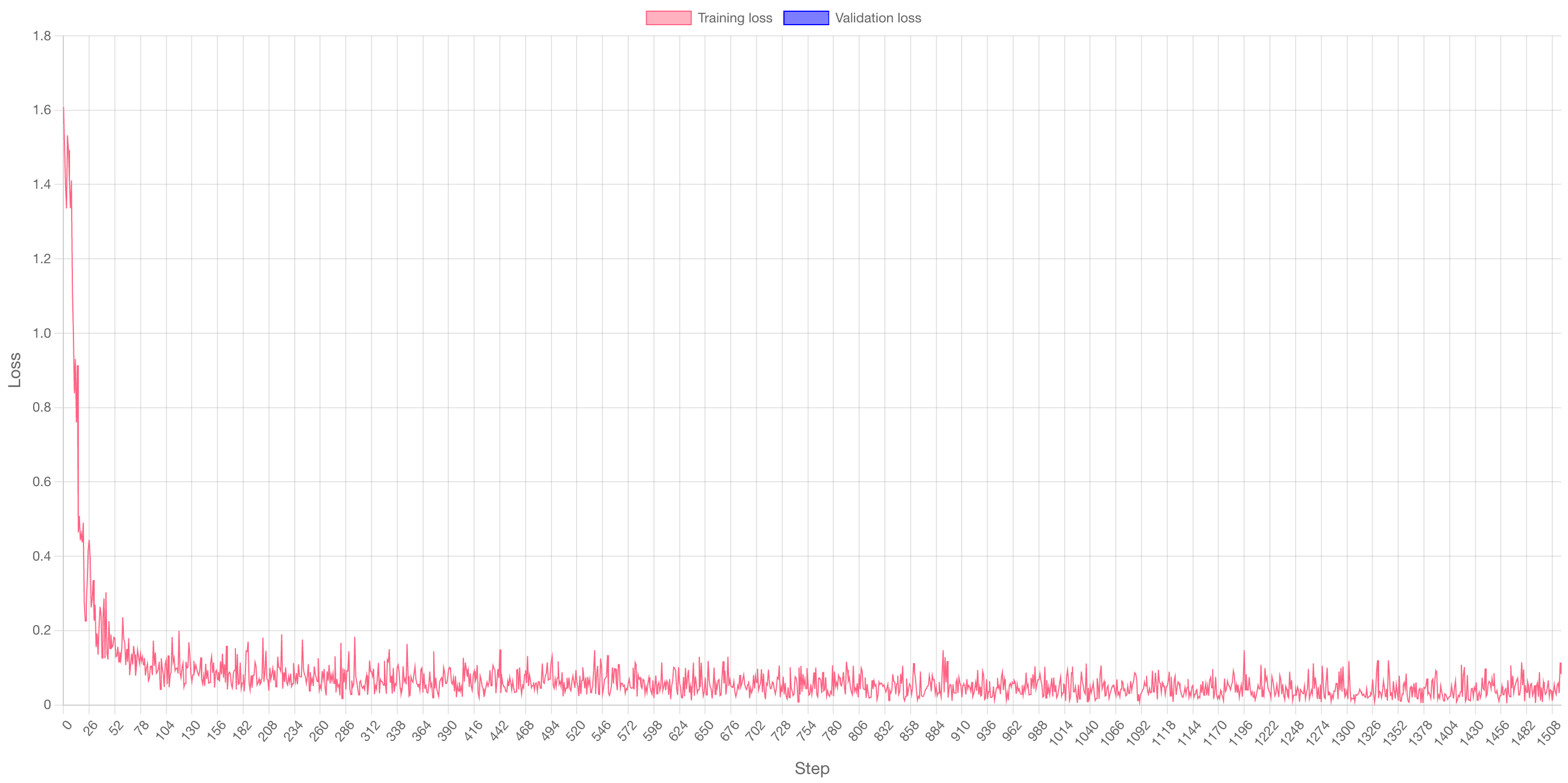}
		\label{fig:MedNLI-3}
	}

    \subfigure[MediQA-RQE (LRM=0.1, Epoch=3)]{
		\includegraphics[width=0.31\linewidth]{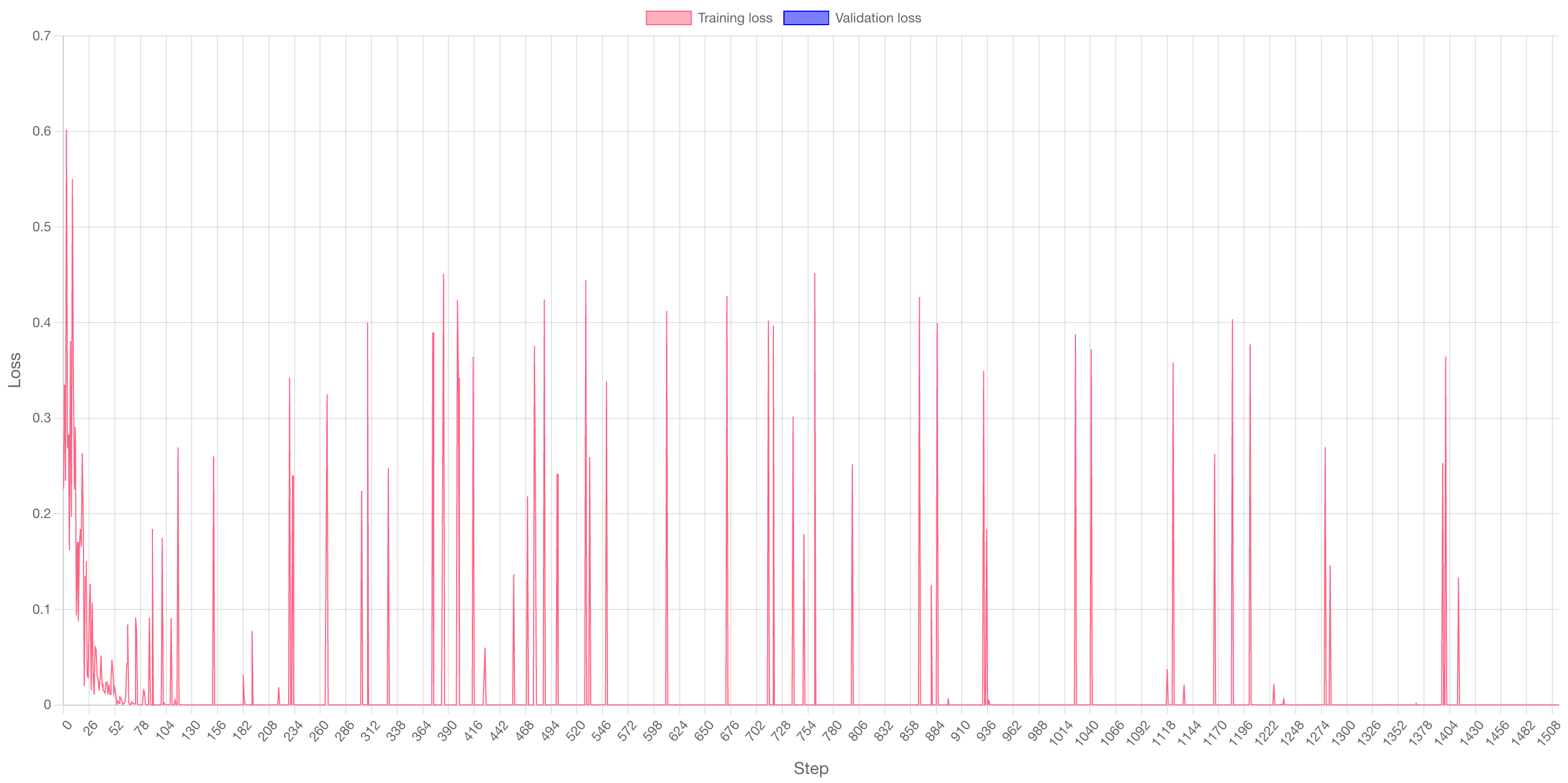}
		\label{fig:MediQA-RQE-1}
	} 
     \subfigure[MediQA-RQE (LRM=1, Epoch=3)]{
		\includegraphics[width=0.31\linewidth]{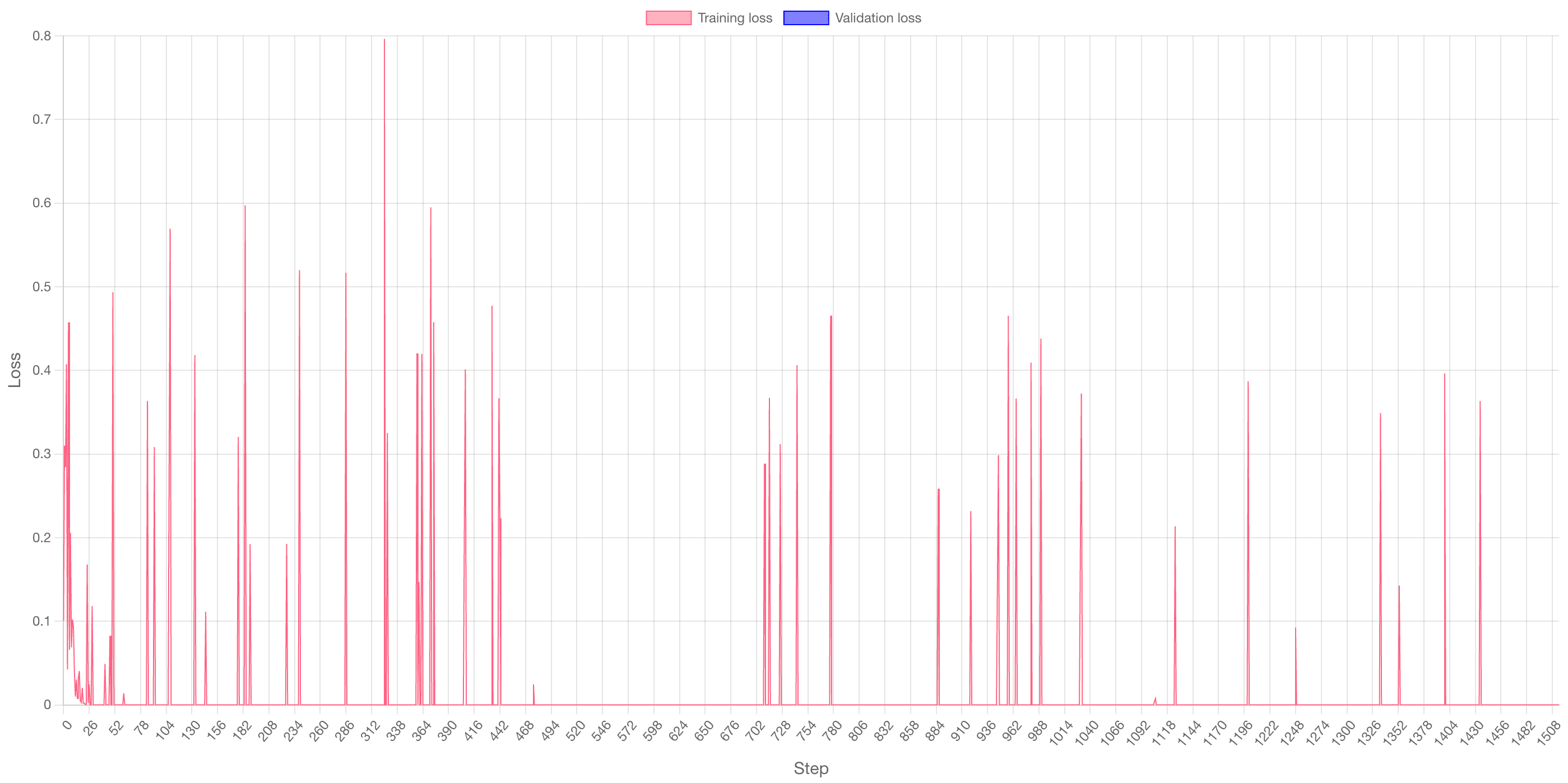}
		\label{fig:MediQA-RQE-2}
	} 
     \subfigure[MediQA-RQE (LRM=0.1, Epoch=5)]{
		\includegraphics[width=0.31\linewidth]{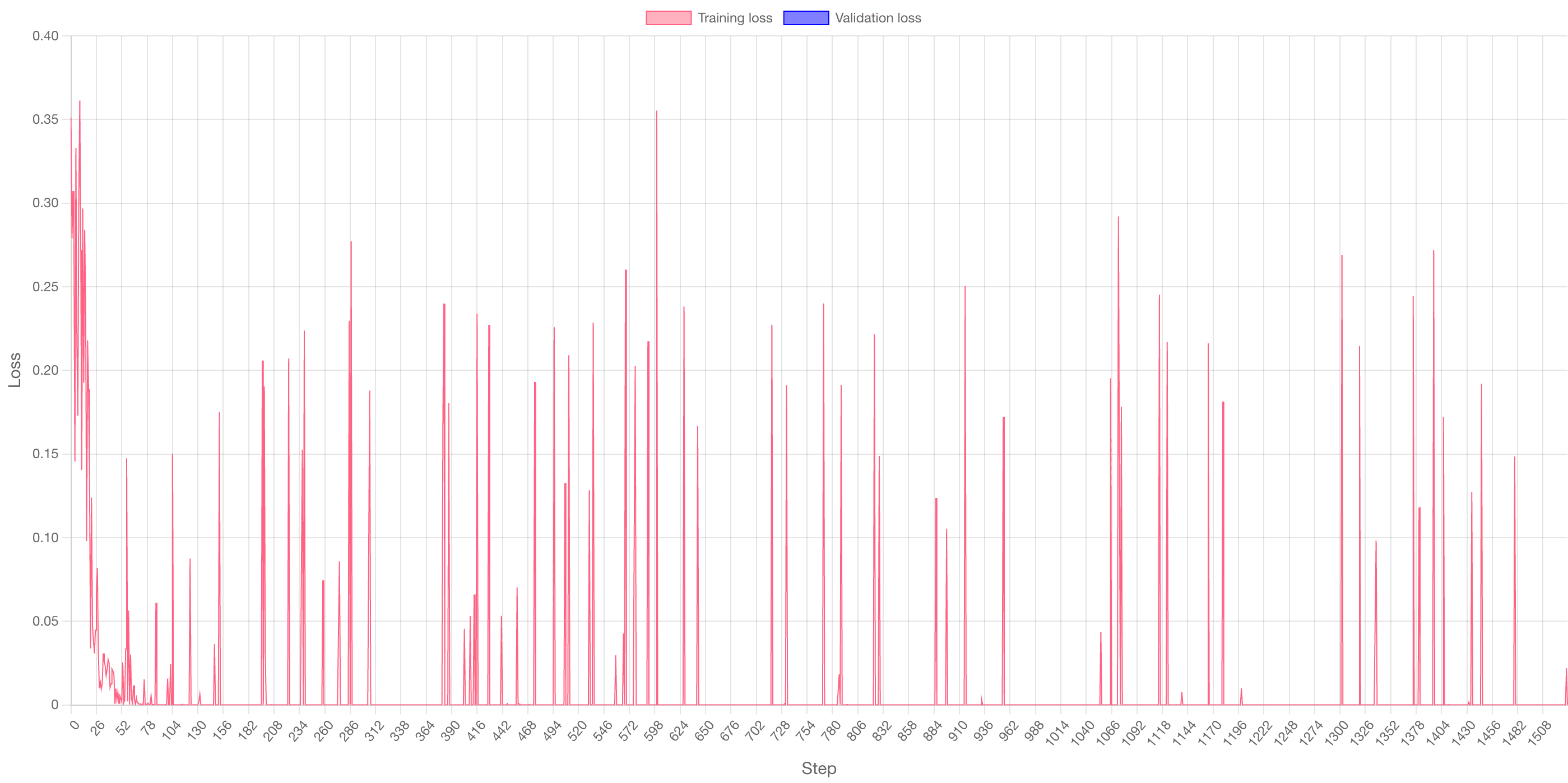}
		\label{fig:MediQA-RQE-3}
	}

 \subfigure[PubHealth (LRM=0.1, Epoch=3)]{
		\includegraphics[width=0.31\linewidth]{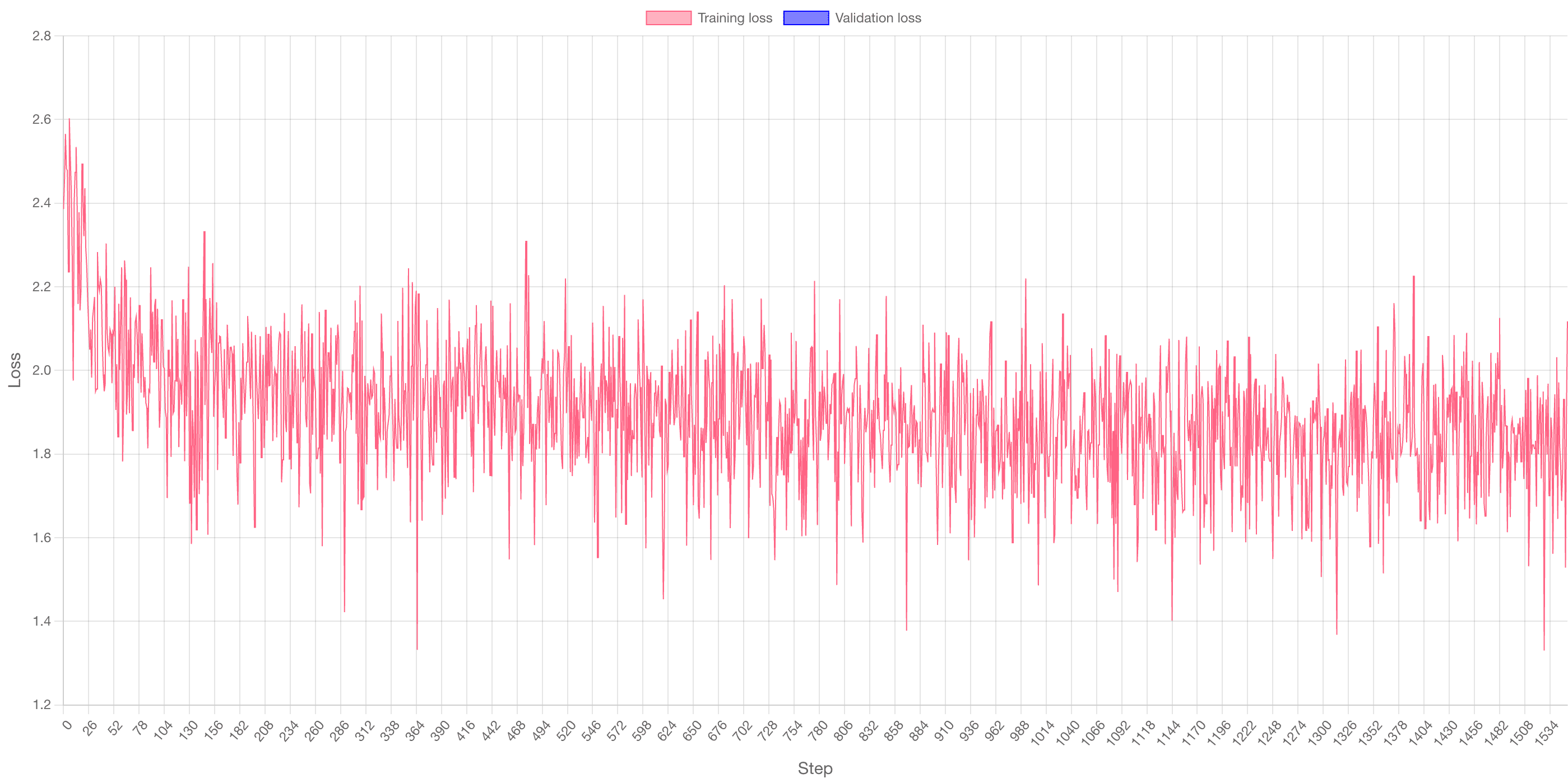}
		\label{fig:PubHealth-1}
	} 
     \subfigure[PubHealth (LRM=1, Epoch=3)]{
		\includegraphics[width=0.31\linewidth]{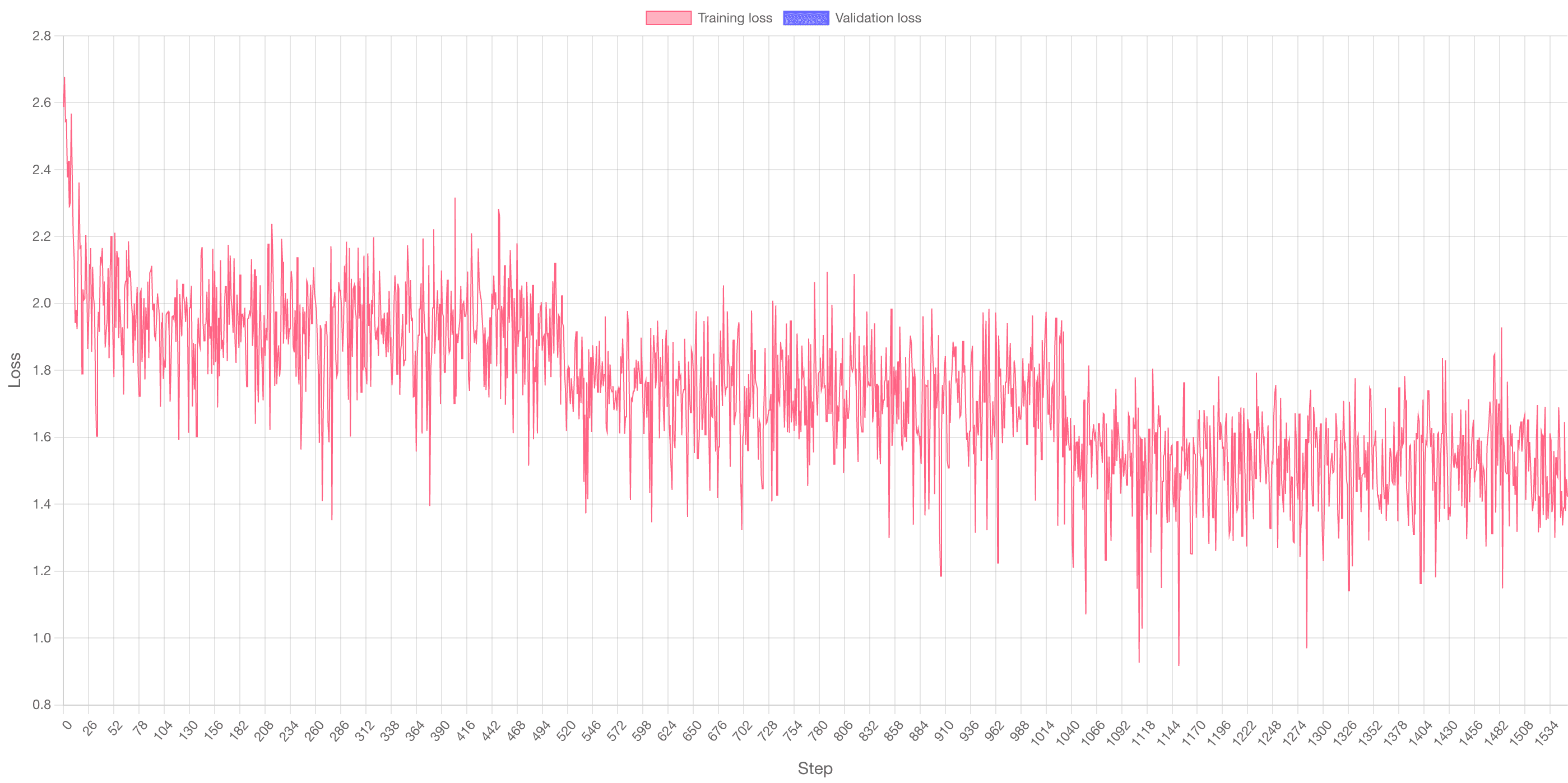}
		\label{fig:PubHealth-2}
	} 
     \subfigure[PubHealth (LRM=0.1, Epoch=5)]{
		\includegraphics[width=0.31\linewidth]{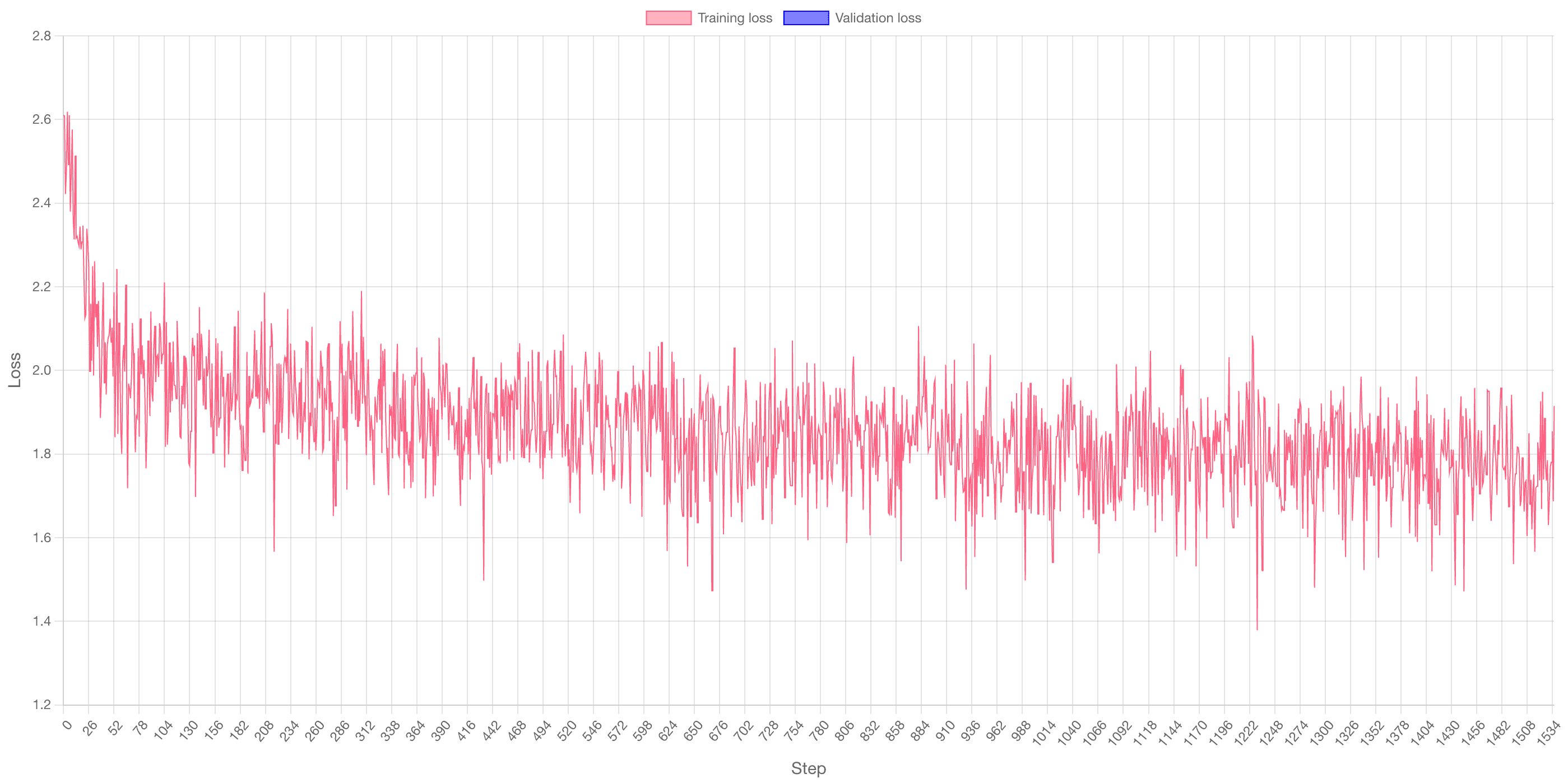}
		\label{fig:PubHealth-3}
	}
	\caption{Loss function curve of fine-tuning GPT-3.5-Turbo for other biomedical tasks through Microsoft Azure fine-tuning API service.  }
\label{fig:azure-loss2}
\end{figure*}

We conduct parameter studies on fine-tuning GPT-3.5-Turbo using the Microsoft Azure fine-tuning API service, as detailed in Table~\ref{tab:azure-sft-parastudy}. The training loss curves of the main biomedical QA and additional biomedical tasks are depicted in Figures~\ref{fig:azure-loss} and~\ref{fig:azure-loss2}, respectively. The Azure-SFT service offers only a very limited number of adjustable hyperparameters, such as the learning rate multiplier (LRM) and the number of epochs, which leads to a lack of transparency and results in suboptimal fine-tuning performance (Table~\ref{tab:main-white-box}).

\begin{table*}[ht]
\centering
\fontsize{8}{10}\selectfont\setlength{\tabcolsep}{0.2em}
\begin{tabular}{@{}cc|cccccccc@{}}
\toprule
\textbf{LRM} & \textbf{Epoch} & \textbf{MedMCQA} & \textbf{MedQA} & \textbf{MMLU} & \textbf{PubMedQA} & \textbf{BioASQ} & \textbf{MedNLI} & \textbf{MediQA-RQE} & \textbf{PubHealth} \\\midrule
0.1 & 3 & 57.87 & \textbf{63.32} & 64.78 & \textbf{68.80} & 95.16 & 87.06 & 55.65 & \textbf{36.56}\\
1 & 3 & 59.69 & 62.92 & \textbf{70.55} & 68.60 & \textbf{95.97} & \textbf{91.27} & 53.27 & 35.17\\ 
0.1 & 5 & \textbf{61.82} & 60.75 & 67.48 & 71.40 & 91.94 & 88.11 & \textbf{58.08} & 34.17\\\bottomrule
\end{tabular}
\caption{Grid search of fine-tuning GPT-3.5-Turbo through Microsoft Azure fine-tuning API service. 
\textbf{Bold} denotes the optimal results chosen as a reference for Azure-SFT.
}\label{tab:azure-sft-parastudy}
\end{table*}

\section{Base Language Model Details for Scale-up Analysis}
\label{app:scale}

Table~\ref{tab:scale-up-model} describes details of the base model of \method in scale-up analysis, ranging from 110M to 2.7B parameters. 

\begin{table}[ht]
\centering
\fontsize{8}{10.5}\selectfont\setlength{\tabcolsep}{0.5em}
\begin{tabular}{@{}lccc@{}}
\toprule
\textbf{Type}        & \textbf{Size}     & \textbf{Model}           \\ \midrule
General LM          & 110M          &   LongFormer-Base~\citep{beltagy2020longformer}            \\ 
General LM          & 330M             &   LongFormer-Large~\citep{beltagy2020longformer}                \\ 
General LM          & 1.3B           & Phi-1.5~\citep{li2023textbooks}   \\
General LM          & 2.7B        &    Phi-2~\citep{li2023textbooks}                             \\ \midrule
Biomedical LM          & 110M              & Clinical-LongFormer~\citep{li2022clinical}   \\ 
Biomedical  LM          & 2.7B              & BioMedLM~\citep{bolton2024biomedlm}   \\ 
\bottomrule
\end{tabular}
\caption{Details of base language models for scale-up analysis.}
\label{tab:scale-up-model}
\end{table}

\section{Learning Objectives Details}
\label{app:loss}

\noindent \textbf{Pairwise Loss.}
Similar to a reward model, our proposed \method also assigns a scalar reward value to each response.
We can then combine the pairwise loss used in reward models to differentiate between positive and negative samples.
We reconstruct the original dataset to be comprised of paired comparisons between two responses generated for the same input or prompt.
With the data generated in Section~\ref{subsec:data-gen}, given a problem description $\mathbf{x}_i$, we leverage the corresponding ground-truth answer and the generations with the correct answers as positive samples $\mathbf{h}^{+}=\mathbf{h}\cap\{\hat{\mathbf{h}}_{i,j}\cdot\mathbbm{1}_{(\hat{\mathbf{h}}_{i,j}=\mathbf{h}_i)}\}$, and those generated solutions with incorrect answers as negative samples $\mathbf{h}^{-}=\{\hat{\mathbf{h}}_{i,j}\cdot\mathbbm{1}_{(\hat{\mathbf{h}}_{i,j}\neq\mathbf{h}_i)}\}$.
We sample at most $k$ positive-negative pairs for each question.
The pairwise learning objective is defined as follows:
\begin{equation}
    \begin{aligned}
        &\mathcal{L}_{\text{pair}}(\mathbf{x}_i,\mathbf{h}^{+}_i,\mathbf{h}^{-}_i;\theta)\\
        &\ =\log \sigma (r_\theta(\mathbf{h}^{+}_i)-r_\theta(\mathbf{h}^{-}_i)),\\
        &\ =\log \sigma (r_\theta([\mathbf{x}_i||\hat{\mathbf{s}}^{+}_{i}||\hat{\mathbf{y}}^{+}_{i}])-r_\theta([\mathbf{x}_i||\hat{\mathbf{s}}^{-}_i||\hat{\mathbf{y}}^{-}_i])).
        \end{aligned}
\end{equation}

\noindent \textbf{InfoNCE Loss.}
InfoNCE Loss extends the original positive-negative pair into the comparison between one positive sample and $k$ negative samples.
To optimize towards the ground-truth answers, we set the corresponding ground-truth solution and answer as the positive sample $\mathbf{h}_i^{+}=\mathbf{h}_i$ for the given question $\mathbf{x}_i$.
Regarding the negative samples, we select all the generated samples from the LLM itself, denoted as $\mathbf{h}i^{-}={\hat{\mathbf{h}}{i,j}}$.
Thus, we can define the InfoNCE loss function as follows:
\begin{equation}
    \begin{aligned}
        \mathcal{L}_{\text{InfoNCE}}&=-\mathbb{E}[\log\frac{r_\theta(\mathbf{h}^{+})}{\sum_{\hat{\mathbf{h}}_{i,j}\in\mathbf{h}^{-}}r_\theta(\hat{\mathbf{h}}_{i.j})}].
    \end{aligned}
\end{equation}


\section{Human Evaluation Guidelines}
\label{app:human}

\subsection{Biomedical QA Task}
The human guideline for biomedical QA tasks is listed as follows:
\vspace{1ex}
\begin{Verbatim}
The goal of this evaluation task is to assess 
the given task input, ground-truth answer, and
a pair of reasoning solutions from the LLM. 
Your objective is to determine which reasoning 
the solution will ultimately yield the correct 
ground-truth answer for that input. For all 
biomedical QA datasets, your responsibility is 
to provide a response to each question using 
either 'True' or 'False'.
\end{Verbatim}

\subsection{Biomedical NLI Task}
The human guideline for biomedical NLI tasks is listed as follows:
\vspace{1ex}
\begin{Verbatim}
The goal of this evaluation task is to assess 
the given task input, ground-truth answer, and
a pair of reasoning solutions from the LLM. 
Your objective is to determine which reasoning 
the solution will ultimately yield the correct 
ground-truth answer for that input. For the 
biomedical NLI datasets, your responsibility 
is to provide a response to predict if the 
hypothesis is entailed/neutral/contradicts 
the premise. 
\end{Verbatim}

\subsection{``Win-Tie-Lose'' Judge}
For each instance, we randomly sample two generated solutions, $e_1, e_2$, from eight candidates, with one from the top four (positive) and the other one from the bottom four scores (negative). We then compare \method with human raters by asking four humans to determine which candidate reasoning solution is better, using $c_i$ $(i=1,2)$ to denote the number of raters that select $e_i$. We denote the adaptation scores based on \method as $(s_{e_1}, s_{e_2})$. 
The final "Win-Tie-Lose" judgment is determined as follows:
(1) \textbf{Win}: if $(c_1 > c_2 \text{ and } s_{e_1} > s_{e_2})$ or $(c_1 < c_2 \text{ and } s_{e_1} < s_{e_2})$; (2) \textbf{Tie}: if $c_1 = c_2$; and (3) \textbf{Lose}: if $(c_1 < c_2 \text{ and } s_{e_1} > s_{e_2})$ or $(c_1 > c_2 \text{ and } s_{e_1} < s_{e_2})$. 
A higher win rate indicates a greater level of alignment with human preference.

\section{Case Study of Adaptation Scores}
\label{app:case}

\begin{table*}[t]
\centering
\caption{Case study for \method on MMLU Dataset.}
\vspace{-1ex}
\resizebox{0.98\textwidth}{!}{%
\begin{tabular}{p{16.8cm}|l}
\toprule
 \multicolumn{2}{p{20cm}}{\textbf{Question}: Which of the following is true in a sensory assessment of the arm? (A) It is acceptable to test pain with a venepuncture needle, (B) Proprioception is appreciated via the same tracts as light touch, (C) It is best to test vibration sense with a 256 Hz tuning fork, (D) Impulses for pain travel principally up the dorsal columns}\\  \multicolumn{2}{l}{\textbf{Ground-Truth Answer}: \textcolor{blue}{B}; \quad  \quad \textbf{Self-Consistency}: \textcolor{red}{D}.} \\
 \hline 
\textbf{Generation 1}: The sensory assessment of the arm includes testing for pain, proprioception, light touch, and vibration sense. To test pain, a venepuncture needle is not acceptable as it can cause injury and is not a standardized method for pain assessment. Proprioception is appreciated via the dorsal columns, distinct from the tracts for light touch. Vibration sense is best tested with a 128 Hz tuning fork, not a 256 Hz tuning fork. Impulses for pain travel principally up the anterolateral system, not the dorsal columns. \#\#\#\# A. \\ 
\textbf{Prediction 1}: \textcolor{red}{A} & Score 1: 0.318\\ \hline 
\textbf{Generation 2}:  Sensory assessment of the arm involves testing different types of sensation, including pain, proprioception, light touch, and vibration. It is not acceptable to test pain with a venepuncture needle as it can cause tissue damage and is not a standardized method of assessment. Proprioception is appreciated through the dorsal columns, while light touch is appreciated through the spinothalamic tracts, so they are not via the same tracts. Vibration sense is best tested with a 128 Hz tuning fork, not a 256 Hz tuning fork.Impulses for pain travel principally up the spinothalamic tracts, not the dorsal columns. Therefore, the correct answer is (D) Impulses for pain travel principally up the dorsal columns. \#\#\#\# D. \\ 
\textbf{Prediction 2}: \textcolor{red}{D} & Score 2: 0.767 \\ \hline 
\textbf{Generation 3}:  Sensory assessment of the arm involves testing different modalities such as pain, proprioception, light touch, and vibration. Pain testing should be performed using a sharp object or pinprick, not a venepuncture needle. Proprioception is appreciated via the dorsal column-medial lemniscal pathway, which is different from the pathway for light touch. Vibration sense is best tested with a 128 Hz tuning fork, not a 256 Hz tuning fork. Impulses for pain travel primarily up the anterolateral system, not the dorsal columns. (D) Impulses for pain travel principally up the dorsal columns.  \\ 
\textbf{Prediction 3}: \textcolor{red}{D} & Score 3: 0.366 \\ \hline 
\textbf{Solution 4}:  In sensory assessment of the arm, pain is typically tested using a sharp object or pinprick to assess the integrity of small, myelinated (A-delta) and unmyelinated (C) fibers. Proprioception, or the awareness of joint position, is appreciated via the dorsal columns of the spinal cord, which are different from the tracts for light touch. Vibration sense is best tested with a 256 Hz tuning fork, as this frequency is most commonly used for assessing vibration perception. Impulses for pain travel principally up the anterolateral (spinothalamic) tracts, not the dorsal columns. \#\#\#\# A.  \\ 
\textbf{Prediction 4}: \textcolor{red}{A} & Score 4: 0.143 \\ \hline 
\textbf{Explanation 5}:  Sensory assessment of the arm involves evaluating various sensations such as pain, proprioception, touch, and vibration. It is not acceptable to test pain with a venepuncture needle as it may cause unnecessary discomfort and potential harm to the patient. Proprioception, the sense of the relative position of one's own parts of the body and strength of effort being employed in movement, is appreciated via the dorsal columns of the spinal cord.  Light touch and proprioception are appreciated via the same tracts, specifically the dorsal column-medial lemniscal pathway. It is best to test vibration sense with a 256 Hz tuning fork as it is known to be the most effective frequency for testing vibration sense. Impulses for pain travel principally up the anterolateral pathways, specifically the spinothalamic tract. \#\#\#\# A.  \\ 
\textbf{Prediction 5}: \textcolor{red}{A} & Score 5: 0.301 \\ \hline 
\textbf{Explanation 6}: In a sensory assessment of the arm, pain should not be tested with a venepuncture needle as it may cause unnecessary harm. Proprioception is appreciated via the dorsal column-medial lemniscal pathway, different from the pathway for light touch. Vibration sense is best tested with a 128 Hz tuning fork, not a 256 Hz tuning fork. Impulses for pain travel principally up the anterolateral system, not the dorsal columns. \#\#\#\# D. \\ 
\textbf{Prediction 6}: \textcolor{red}{D} & Score 6: 0.777 \\ \hline 
\textbf{Explanation 7}:  The sensory assessment of the arm involves testing for different modalities of sensation, such as pain, proprioception, light touch, and vibration. It is not acceptable to test pain with a venepuncture needle as it could cause unnecessary discomfort and potential harm to the patient. Proprioception is appreciated via the dorsal column-medial lemniscal pathway, which is different from the pathway for light touch. Vibration sense is best tested with a 128 Hz tuning fork, not 256 Hz tuning fork, as the former is the standard frequency for this assessment. Impulses for pain travel principally up the anterolateral system, not the dorsal columns. Therefore, the correct statement is (D) Impulses for pain travel principally up the dorsal columns.  \#\#\#\# D. \\ 
\textbf{Prediction 7}: \textcolor{red}{D} & Score 7: 0.754 \\ \hline 
\textbf{Explanation 8}: Sensory assessment of the arm involves testing for pain, temperature, light touch, proprioception, and vibration. To test pain, a blunt object or pinprick is used instead of a venepuncture needle to prevent injury to the patient. Proprioception is appreciated via the dorsal columns, while light touch is appreciated via the spinothalamic tracts. Vibration sense is best tested with a 128 Hz tuning fork, not 256 Hz. Impulses for pain travel principally up the spinothalamic tracts, not the dorsal columns. \#\#\#\# B. (B) Proprioception is appreciated via the same tracts as light touch.  \\ 
\textbf{Prediction 8}: \textcolor{blue}{B} & \textbf{Score 8: 0.896} \\  
\bottomrule
\end{tabular}%
}
\label{tab:case_mmlu}
\end{table*}
Table~\ref{tab:case_mmlu} gives an example of \method on MMLU dataset.
Given the question displayed in the figure, the original self-consistency method selects the most commonly-seen answer ``D'' as the final answer.
Via going through all the training data, the adapter is able to select the most adapted answer from all the candidates and avoid factual errors.
For example, generation 4 makes an error regarding the frequency for testing vibration sense and the low score (0.143) is reflective of this mistake.
For generations 2, 6, and 7, the solutions provide accurate information but arrive at the wrong conclusion. The high scores (0.767, 0.777, 0.754) reflect the correctness of the reasoning but not the final answer.
With the guidance of the \method, we finally select ``B'', which is accurate and concludes with the correct answer.



\end{document}